\documentclass{article}

\usepackage{microtype}
\usepackage{graphicx}
\usepackage{subfigure}
\usepackage{booktabs} 

\usepackage{hyperref}

\usepackage[accepted]{icml2024}

\usepackage{amsmath}
\usepackage{amssymb}
\usepackage{mathtools}
\usepackage{amsthm}

\usepackage[capitalize,noabbrev]{cleveref}

\theoremstyle{plain}
\newtheorem{theorem}{Theorem}[section]

\newtheorem{lemma}[theorem]{Lemma}

\theoremstyle{definition}
\newtheorem{definition}[theorem]{Definition}
\newtheorem{assumption}[theorem]{Assumption}
\theoremstyle{remark}

\usepackage[textsize=tiny]{todonotes}

\icmltitlerunning{$\mathtt{VITS}$ :
      Variational Inference Thompson Sampling for contextual bandits}

\usepackage{bbm}
\usepackage{accents}
\usepackage{bm}
\usepackage{enumitem}
\usepackage{wrapfig}
\usepackage{algorithm}
\usepackage{algorithmic}
\usepackage{mathtools}
\usepackage{epstopdf}
\usepackage{graphicx}

\DeclareMathOperator*{\argmin}{arg\,min} 
\DeclareMathOperator*{\argmax}{arg\,max}

\newcommand{\pierre}[1]{{\textcolor{blue}{[Pierre: #1]}}}

\newcommand{\varfam}{\mathcal{G}}
\newcommand{\varpostexact}{\mathrm{q}  }

\newcommand{\A}{\mathcal{A}}
\newcommand{\filt}{\mathcal{F}}
\newcommand{\rset}{\mathbb{R}}

\newcommand{\X}{\mathsf{X}}
\newcommand{\nsets}{\mathbb{N}^{\star}}
\newcommand{\data}{\mathrm{D}}
\newcommand{\Rrm}{\mathcal{R}}
\newcommand{\Hyp}{(\mathrm{H})}

\newcommand{\grad}{\mathfrak{g}}

\newcommand{\rmR}{\mathcal{R}}
\newcommand{\R}{\mathbb{R}}
\newcommand{\hist}{\mathcal{H}}
\newcommand{\overbar}[1]{\mkern 1.5mu\overline{\mkern-1.5mu#1\mkern-1.5mu}\mkern 1.5mu}
\newcommand{\Ber}{\mathrm{Ber}}

\newcommand{\Etrue}{\mathrm{E}^{\textrm{true}}}
\newcommand{\Etruebar}[1]{\overbar{\mathrm{E}}_{#1}^{\textrm{true}}}

\newcommand{\Evar}{\mathrm{E}^{\textrm{var}}}
\newcommand{\Evarbar}[1]{\overbar{\mathrm{E}}_{#1}^{\textrm{var}}}

\newcommand{\post}{\hat{\mathrm{p}}}
\newcommand{\varpost}{\tilde{\mathrm{q}}}
\newcommand{\thetastar}{\theta^{\star}}
\newcommand{\meanpost}{\hat{\mu}}

\newcommand{\thetavar}{\tilde{\theta}}
\newcommand{\meanvar}{\tilde{\mu}}
\newcommand{\stdvar}{\tilde{\Sigma}}

\newcommand{\precivar}{\tilde{\Sigma}^{-1}}

\newcommand{\stdpost}{\hat{\Sigma}}

\newcommand{\prior}{\rmp_0}

\newcommand{\meanmeanvar}{\tilde{m}}
\newcommand{\stdstdvar}{\tilde{W}}

\newcommand{\ba}{\bar{a}}

\newcommand{\phistar}{\accentset{\ast}{\phi}}

\newcommand{\phibar}{\overline{\phi}}

\newcommand{\Vinv}{V^{-1}}
\newcommand{\1}{\mathbbm{1}}

\newcommand{\prob}{\mathbb{P}}
\newcommand{\E}{\mathbb{E}}
\newcommand{\D}{\mathrm{d}}

\newcommand{\KL}{\mathrm{KL}}
\newcommand{\norm}[1]{\lVert#1\rVert}
\newcommand{\normdeux}[1]{\lVert#1\rVert_{2}}

\newcommand{\lmax}{\lambda_{\text{max}}}
\newcommand{\lmin}{\lambda_{\text{min}}}
\newcommand{\id}{\mathrm{I}}
\newcommand{\CREG}[1]{\mathrm{CRegret}(#1)}

\newcommand{\ps}[2]{\left\langle#1,#2 \right\rangle}
\def\eqsp{\;}
\newcommand{\sdp}{\mathcal{S}_{+}}
\newcommand{\sdpp}{\mathcal{S}^{*}_+}
\newcommand{\pivar}{\pi^{(\mathrm{VITS})}}

\newcommand{\likelihood}{\mathrm{L}}
\newcommand{\nloglikelihood}{\ell}

\newcommand{\as}{a^{\star}}

\newcommand{\N}{\mathrm{N}}

\def\VITS{\texttt{VITS}}

\def\msa{\mathsf{A}}

\def\mse{\mathsf{E}}

\def\msr{\mathsf{R}}

\def\msx{\mathsf{X}}

\def\mca{\mathcal{A}}

\def\lambdaref{\lambda_{\mathrm{ref}}}
\def\Leb{\mathrm{Leb}}

\def\rmd{\mathrm{d}}

\def\rme{\mathrm{e}}

\def\rmp{\mathrm{p}}

\newcommand{\V}{\mathbb{V}}

\newcommand{\nofrac}[2]{#1/#2}

\newcommand{\defEns}[1]{\left\{ #1 \right \}}
\newcommand{\parenthese}[1]{\left( #1 \right)}
\def\VITSI{\textcolor{blue}{\VITS-\texttt{I}}}
\def\VITSII{\textcolor{orange}{\VITS-\texttt{II}}}
\def\VITSIIHF{\textcolor{purple}{\VITS-\texttt{II Hessian-free}}}

\begin{document}
\twocolumn[
\icmltitle{$\mathtt{VITS}$ :
      Variational Inference Thompson Sampling for contextual bandits}



\icmlsetsymbol{equal}{*}

\begin{icmlauthorlist}
\icmlauthor{Pierre Clavier}{equal,cmap,heka,inserm}
\icmlauthor{Tom Huix}{equal,cmap}
\icmlauthor{Alain Durmus}{cmap}
\end{icmlauthorlist}

\icmlaffiliation{cmap}{CMAP, CNRS, Ecole Polytechnique,
Institut Polytechnique de Paris,
91120 Palaiseau, France.}
\icmlaffiliation{heka}{Inria Paris, 75015 Paris, France}
\icmlaffiliation{inserm}{Centre de Recherche des Cordeliers, INSERM, Universite de Paris, Sorbonne Universite, 75006 Paris, France }
\icmlcorrespondingauthor{Tom Huix}{tom.huix@polytechnique.edu}
\icmlcorrespondingauthor{Pierre Clavier}{pierre.clavier@polytechnique.edu}

\icmlkeywords{Machine Learning, Bandit, Thompson Sampling}

\vskip 0.3in
]



\printAffiliationsAndNotice{\icmlEqualContribution} 

\begin{abstract}
  In this paper, we introduce and analyze a variant of the Thompson sampling (TS) algorithm for contextual bandits. At each round, traditional TS requires samples from the current posterior distribution, which is usually intractable. To circumvent this issue, approximate inference techniques can be used and provide samples with distribution close to  the posteriors. However, current approximate techniques yield to either poor estimation (Laplace approximation) or can be computationally expensive (MCMC methods, Ensemble sampling...). In this paper, we propose a new algorithm, Varational Inference TS ($\mathtt{VITS}$), based on Gaussian Variational Inference. This scheme provides powerful posterior approximations which are easy to sample from, and is computationally efficient, making it an ideal choice for TS. In addition, we show that $\mathtt{VITS}$~achieves a sub-linear regret bound of the same order in the dimension and number of round as traditional TS for linear contextual bandit. Finally, we demonstrate experimentally the effectiveness of $\mathtt{VITS}$ on both synthetic and real world datasets. 
\end{abstract}

\section{Introduction}

In traditional Multi-Armed Bandit (MAB) problems, an agent, has to sequentially choose between several actions (referred to as "arms"), from which he receives a reward from the environment. The arm selection process is induced by a sequence of policies, which is inferred and refined at each round from past observations. These policies are designed to optimize the cumulative rewards over the entire process. The main challenge in this task is to effectively manage a suitable exploitation and exploration trade-off  \citep{robbins1952some, Katehakis1987TheMB,berry1985bandit,auer2002finite,lattimore2020bandit,pmlr-v108-kveton20a}. Here, exploitation refers to selecting an arm that is currently believed to be the best based on past observations, while exploration refers to selecting arms that have not been selected frequently in the past in order to gather more information.  

Contextual bandit problems is a particular instance of MAB problem, which supposes, at each round, that the set of arms and the corresponding reward depend on a $d$-dimensional feature vector called a contextual vector or context. This scenario has been extensively studied over the past decades and learning algorithms have been developed to address this problem \citep{langford2007epoch,abbasi2011improved,agrawal2013thompson,pmlr-v108-kveton20a}, and they have been successfully applied in several real-world problem such as recommender systems, mobile health and finance \citep{li2010contextual,agarwal_bird,Tewari2017FromAT,bouneffouf2020survey}. The existing algorithms for addressing contextual bandit problems can be broadly categorized into two groups. The first category is based on maximum likelihood and the principle of optimism in the face of uncertainty (OFU) and has been studied in \citep{auer2002finite,chu_et_al:2011,abbasi:2011,liluzhou,menard2017minimax,zhou2020neural,foster2020beyond, zenati2022efficient}.

The second category consists in randomized probability matching algorithms, which is based on Bayesian belief and posterior sampling. Thompson Sampling (TS)  is one of the most famous algorithms that fall into this latter category. Since its introduction by \citet{thompson1933likelihood}, it has been widely studied, both theoretically and empirically \citep{agrawal2012analysis,kaufmann2012thompson,agrawal2013thompson,russo2014learning,russo2016information,lu2017ensemble,riquelme2018deep,jin2021mots}. Despite the fact that OFU algorithms offer better theoretical guarantees compared to classic TS-based algorithms, traditional TS methodologies still appeal to us due to their straightforward implementation and empirical advantages. In \cite{agrawal2012analysis}, the authors claimed that: "In applications like display advertising and news article recommendation, TS is competitive with or better than popular methods such as UCB". Similarly, \cite{chapelle2011empirical} has examined the empirical performances of TS on both simulated and real data. Their experiments demonstrate that TS outperforms OFU methods, leading them to conclude: "In any case, TS is very easy to implement and should thus be considered as a standard baseline". Taking all these factors into account, we have decided to focus on TS-based algorithms for addressing contextual bandit problems. 

Despite its relative simplicity, effectiveness and convergence guarantees, TS comes with a computational burden which is to sample, at each iteration $t \in \nsets$,  from an appropriate Bayesian posterior distribution $ \post_t$ defined from the previous observations. Indeed, these posteriors are usually intractable and
approximate inference methods have to be used to obtain samples with distributions ”close” to the posterior. The family of TS methods using approximate inference methods will be referred to as approximate inference TS in the sequel. Among the simplest approximate
inference methods, Laplace approximation has been proposed for TS in
\citet{chapelle2011empirical}. This method consists of approximating
the posterior distribution $\post_t$ by a Gaussian distribution with a
carefully chosen mean and covariance matrix. More precisely, the mean
is a mode of the target distribution which is typically found using an
optimization algorithm, while the covariance matrix is taken to be the
negative Hessian matrix of the log posterior at the considered
mode. Despite this method is easy to implement, it may lead to poor posterior representations. Indeed, while Laplace method achieves minimal optimality in terms of regret \citep{faury2022jointly}, it doesn't dictate the posterior convergence rate. More precisely, in \cite{katsevich2023approximation} it has been demonstrated that VI outperforms Laplace in terms of mean convergence by a factor of $1/n$ . It is worth noting that the covariance rates remain the same for both methods. This discrepancy can lead to inadequate approximations, especially in high-dimensional settings, as highlighted in section I.4 of \cite{katsevich2023approximation}.

Another class of popular approximate inference
methods are  Markov Chain Monte Carlo (MCMC) methods, such as Metropolis or Langevin Monte Carlo (LMC) algorithms. In the bandit literature, LMC has been proposed to get approximate samples from TS posteriors for solving traditional bandit problem in  \cite{mazumdar2020thompson} and for contextual
bandit problems in \citet{xu2022langevin, fghuix}. Also, \citet{lu2017ensemble} have proposed to adapt Ensemble Methods to the bandit setting. Roughly, the idea here is to maintain and
incrementally update an ensemble of statistically plausible models and
to draw a uniform sample from this family at each iteration.

Finally, Variational Inference (VI) \citep{blei2017variational} is another class of approximate method that could be used to get samples from the posterior distribution. The core concept behind VI is to find a distribution $\varpost$, referred to as the variational posterior, to closely match the true posterior $\post$ in terms of Kullback-Leibler divergence ($\KL$) within a predefined family of distributions known as the variational family $\varfam$. In general, the variational family is chosen to make the optimization of the $\KL$ tractable and to be easy to sample from. In their work \citet{urteaga2018variational} propose the mean-field mixture of Gaussian variational family for TS. This family of distributions is quite extensive and provides an accurate approximation for a wide range of posterior distributions. However, in our perspective, it might not be the most suitable choice for TS. Firstly, the optimization algorithm at each time step can be computationally expensive. Secondly, the mean-field assumption assumes that the parameters are independent, a premise that holds true in the regime of large, overparameterized models. In our perspective, this regime may not align with the Bandit problem, which often operates in a setting where the number of data points tends towards infinity in comparison to the model size.
Finally, \citet{yu2020graphical} also employs VI in more general graphical models but focuses on structured arms and rewards, where the rewards are correlated through latent variables.

In this paper, we develop an efficient VI method that makes use of the whole family of non-degenerate Gaussian distributions. This choice of VI family is supported by
the Bernstein-Von Mises theorem \citep{van2000asymptotic}
. This theorem, subject to specific regularity conditions, asserts that a properly scaled version of the posterior converges to a Gaussian as the sample size grows. When applied to contextual bandits, the data points progressively accumulate over time, leading to the gradual concentration of the posterior around a dominant mode. As a consequence, the Gaussian approximation becomes increasingly suitable for representing the posterior in this particular setting. Furthermore, the covariance of the rescaled posterior distribution tends to converge towards the inverse Fisher information matrix, which may not necessarily be diagonal, thus justifying the need for a non-mean-field hypothesis. Our main contributions can be summarized as follows: 

Our \textbf{first contribution} is methodological. We develop a novel variant of the TS algorithm, referred to as Variational Inference TS ($\mathtt{VITS}$). Our method addresses the main challenges encountered by the existing approximate TS algorithms and can be applied to a very large class of TS posteriors. Moreover, it enjoys a low computational cost both theoretically and empirically, since it boils down to adding a few optimization steps per round. We also propose two approximate versions of $\VITS$, called $\VITSII$ and $\VITSIIHF$, that scale with the problem dimension.


Our \textbf{second contribution} is theoretical. We establish that our proposed methodology achieves a sub-linear regret of order $\tilde{\mathcal{O}}(d^{3/2}\sqrt{T})$ (up to logarithmic term) in the linear contextual bandit framework, where $T$ is the number of rounds and $d$ is the dimension of the policy parameter. To the best of our knowledge, this is the first regret bound derived for VI in the context of sequential learning. 

Finally, our \textbf{last contribution} is to illustrate the empirical performances of our method on a synthetic and on the real world dataset MovieLens (\citet{movielens}). It has been shown that in many cases, $\VITS$ outperforms existing approximate TS algorithms such as LMC algorithm. 
\paragraph{Related work.}
    The theoretical foundations of TS for linear contextual bandits were initially explored by \citet{agrawal2013thompson}. In this paper, the authors establish a sub-linear cumulative regret bound $\tilde{\mathcal{O}}(d^{3/2}\sqrt{T})$ for Linear TS (Lin-TS). Compared to this study, our method achieves a similar regret bound in the linear framework. However, it should be noted that Lin-TS is a specialized algorithm that can be only used when the posterior is known and can be efficiently sampled from.
   
As mentioned previously, VI has  been suggested for TS in \citet{urteaga2018variational}. This paper introduces a TS algorithm called VTS that utilizes a mixture of mean-field Gaussian distributions to approximate the sequence of posteriors. In comparison to this work, the setting and the variational family we consider are richer than \cite{urteaga2018variational}. A more detailed comparison is postponed in Appendix \ref{VTS}. Moreover, the methodology developed in \cite{urteaga2018variational}  does not come with any convergence guarantees.
An empirical and theoretical study of using LMC as approximate inference method for TS for contextual bandit problems was carried out in \citet{xu2022langevin}.  This paper establishes that the resulting algorithm, called LMC-TS, achieves a state-of-the-art sub-linear cumulative regret for linear contextual bandits. Compared to this method, our approach yields a similar sub-linear regret in the same setting.    
Finally, \cite{zhang2020neural} suggests a TS method based on Neural Tangent Kernel. While this performs well on real datasets, their method is much more expensive than previously mentioned approaches, as it requires training a neural network.
\paragraph{Notation.}
For $n \geq1$, $[n]$ represents the set of integers between 1 and
$n$. 
$\N(\mu, \Sigma)$ denotes the d-multidimensional Gaussian
probability distribution with mean $\mu \in \R^d$ and covariance
matrix $\Sigma \in \R^{d \times d}$.
The transpose of a matrix $M$ is denoted by $M^{\top}$. For any symmetric-real matrix $A$, $\lmax(A)$ and $\lmin(A)$ represent the maximum and minimum  eigenvalues of $A$ respectively.
 The norm $\Vert \cdot \Vert_2$ will
refer to the 2-norm for vectors, and the operator norm for matrices. For any semi-definite positive matrix $A$, the norm $\Vert x \Vert_{A}$ denotes the Mahalanobis norm, i.e., $\Vert x \Vert_{A} = \sqrt{x A x^{\top}}$. For any event $\mse$ on a probability space, $\overbar{\mse}$ refers to the complementary of $\mse$. Finally, $\1$ is the indicator function and $\operatorname{tr}$ is the trace of a matrix. 
\section{Thompson sampling for contextual bandits}
\paragraph{Contextual bandit:} We now present in more details the
contextual bandit framework. Let $\X$ be a contextual space and
consider $\A: \X \rightarrow 2^{\msa}$ a set-valued action map, where
$2^{\msa}$ stands for the power set of the action space $\msa$. For
simplicity, we assume here that
$\sup_{x \in \X} \mathrm{Card}(\A(x)) < +\infty$. A (deterministic or
random) function $\pi : \X \to \msa$ is said to be a policy if for any
$x \in \X$, $\pi(x) \in \A(x)$.  Then, for a fixed horizon
$T \in \nsets$, a contextual bandit process can be defined as follows:
at each iteration $t \in [T]$ and given the past observations
$\data_{t-1} = \{(x_s,a_s,r_s)\}_{s < t}$:

\begin{itemize}
\itemsep0em 
\leftmargin=0.1cm
    \item The agent receives a contextual feature  $x_t \in \X$;
    \item The agent chooses an action $a_t = \pi_t(x_t)$ where $\pi_t$ is a policy sampled from  $\mathbb{Q}_t( \cdot | \data_{t-1})$;
    \item Finally, the agent receives a reward $r_t$ sampled from  $\Rrm(\cdot| x_t,a_t)$ given $\data_{t-1}$. Here, $\Rrm$ is a Markov kernel on $(\msa \times \X) \times \mathsf{R}$, where $\mathsf{R} \subset \R$ 
\end{itemize}
For a fixed family of conditional distributions $\mathbb{Q}_{1:T} = \{\mathbb{Q}_t\}_{t \leq T}$, this process defines a random sequence of policies, $\pi_{1:T} = \{\pi_t\}_{t \leq T}$ with distribution still denoted by $\mathbb{Q}_{1:T}$ by abuse of notation. Let's defined the optimal expected  reward for a contextual vector $x \in \msx$  and the expected reward given $x$ and any action $a \in \mca(x)$ as follow  
\begin{align}\label{eq:def_mean_reward_distribution}
f_{\star}(x) = \max_{a \in \A(x)} f(x, a)\eqsp,
 f(x, a) = \int r   \Rrm(\D r| x, a)\eqsp.
\end{align}
The main challenge of a contextual bandit problem is to find the distribution  $\mathbb{Q}_{1:T}$ that minimizes the cumulative regret defined as 
\begin{align}
    \label{eq:regret}
    &\textstyle \CREG{\mathbb{Q}_{1:T}} = \sum_{s \leq T} \mathrm{Regret}_s^{\pi_s} \eqsp \\
 & \nonumber \text{ with} \quad \textstyle \mathrm{Regret}_s^{{\pi_s}} = f_{\star}(x_s) - f(x_s, \pi_s(x_s)).
\end{align}
The main difficulty in the  contextual bandit problem, comes from the fact that the reward distribution $\Rrm$ is intractable and must be inferred to find the best policy to minimize the instantaneous regret $ \pi \mapsto f_{\star}(x) - f(x, \pi(x)) $ for a context $x\in\msx$. However, the estimation of $\Rrm$ may be in contradiction with the primary objective to minimize the cumulative regret \eqref{eq:regret}, since potential non-effective arms has to be chosen to obtain a complete description of $\Rrm$. Therefore, bandit learning algorithms have to achieve an appropriate trade-off between
exploitation of arms which have been confidently learned and exploration of misestimated arms. 
 \paragraph{Thompson sampling:} To achieve such a trade-off, we consider the popular Thompson Sampling (TS) algorithm.  Consider  a parametric model $\{\Rrm_{\theta} \, :\, \theta \in \R^d\}$ for the reward distribution, where for any $\theta$, $\Rrm_{\theta}$ is a Markov kernel on $(\msa \times \X) \times \mathsf{R}$ parameterized by $\theta \in \rset^d$.  We assume in this paper that $\Rrm_{\theta}$ admits a density with respect to some dominating measure $\lambdaref$. An important example are
generalized linear bandits \cite{filippi:2010, pmlr-v108-kveton20a}.
In particular, it assumes that $\{\Rrm_{\theta}(\cdot |x,a) \,: \, \theta \in\Theta\}$ is an exponential family with respect to $\lambdaref$, i.e., for  $x \in\msx$ and $a \in \msa$, 
\begin{equation}
\label{eq:gen_lin_bandit}
\frac{\rmd \rmR_{\theta}}{\rmd \lambdaref}(r|x,a) = \mathrm{h}(r) \exp(g(\theta,x,a) \mathrm{T}(r) - \mathrm{C}(\theta,x,a)), 
\end{equation}
for $h : \msr \to \rset_+$, natural parameter and log-partition function $g,\mathrm{C}: \rset^d \times \msx \times \msa \to \R$ and sufficient statistics $\mathrm{T} : \msr \to \rset$. The family is said to be in canonical form if  $g(\theta,x,a) = \ps{\phi(x,a)}{\theta}$ for some feature map  $\phi : \X\times \msa \rightarrow \mathbb{R}$ and
 $\mathrm{C}(\theta,x,a) = \sigma(\ps{\phi(x,a)}{\theta})$  for some link function $\sigma$. 
 Linear contextual bandits \cite{chu_et_al:2011, abbasi:2011} fall into this model taking $\lambdaref = \Leb$, $\mathrm{T}$ equals to the identity function, \begin{equation}
\label{eq:lin_bandit}
 \text{$\mathrm{h}(r) = \exp(-\eta r^2/2)$ and $g(\theta,x,a)=  \eta \ps{\phi(x,a)}{\theta}$} ,
 \end{equation}
 for some $\eta > 0$.
As a result, $ \rmR_{\theta}(\cdot| x,a)$ is simply the Gaussian distribution with mean  $\ps{\phi(x,a)}{\theta}$ and variance $1/\eta$. Finally \cite{riquelme2018deep, zhou2020neural, xu2020neural} introduced an extension of linear contextual bandits, referred to as linear  neural contextual bandits where $g$ is a neural network with weights $\theta$ and taking as input a pair $(x,a)$. With the introduced notations, the likelihood function associated to the observations $\data_{t}$ at step $t >1$ is given by 
\begin{equation}
    \label{eq:likelihood}
    \likelihood_t(\theta) \propto \exp \defEns{\sum_{s=1} ^{t-1} \nloglikelihood(\theta|x_s,a_s,r_s) } \eqsp,
\end{equation}
 where the log-likelihood is given by $\nloglikelihood(\theta|x_s,a_s,r_s) = \log (\rmd\rmR_{\theta}/\rmd\lambdaref )(r_s|x_s,a_s)\eqsp.$
 Choosing a prior  on $\theta$ with density $\prior$ with respect to $\Leb$, and applying Bayes formula, the posterior distribution at round $t\in[T]$ is given by 
\begin{equation}
    \label{eq:post}
    \post_t =  \nofrac{\likelihood_t(\theta ) \prior(\theta)}{\mathfrak{Z}_t} 
\end{equation}
where $\mathfrak{Z}_t = \int \likelihood_t(\theta)\prior(\theta) \rmd \theta$ denotes the normalizing constant  and we used the convention that $\post_1 = \prior$. Moreover we define the potential function $ U(\theta)\propto -\log \post_t(\theta)$. Then, at each iteration $t \in [T]$, TS consists in sampling a sample $\theta_t$ from the posterior $\post_t$ and from it, use as a policy, $\pi_t^{(\mathrm{TS})}(x)$ defined for any $x$ by
\begin{align}\label{eq:def_a_theta_max}
\small\pi_t^{(\mathrm{TS})}(x) = a^{\theta_t}(x)\eqsp, 
   a^{\theta}(x) = \argmax_{a}\int r \rmR_{\theta}(\rmd r |x,a)  
\end{align}
 Since  $\mathfrak{Z}_t$ is generally intractable, sampling from the posterior distribution is not in general an option. 
 \paragraph{Variational inference TS:} To address this challenge, practitioners often employ approximate inference methods to generate samples from a distribution that is expected to be "close" to the actual posterior distribution. In this context, we specifically concentrate on the application of VI. In this scenario, we consider a variational family $\varfam$ which is a set of probability densities with respect to the Lebesgue measure, from which it is typically easy to sample from. Then ideally,  at each round $t \in [T]$, the posterior distribution $\post_t$ is approximated by the variational posterior distribution $\varpost_t$ which is defined as: 

\begin{equation}
    \label{eq:varpost}
    \varpost_t = \argmin_{\mathrm{p} \in \varfam} \KL(\mathrm{p}| \post_t) \eqsp,
\end{equation}
where $\KL$ is the Kullback-Leibler divergence.  
However, we have to determine at each round a solution to the problem specified in  \eqref{eq:varpost}. 
In this paper, we consider as variational family the set of non-degenerate Gaussian distribution  $\varfam = \{\N(\mu, \Sigma) \, : \, \mu \in \R^d, 
 \, \Sigma \in \sdpp \}$ where $\N(\mu,\Sigma)$ is the Gaussian distribution with mean $\mu$ and covariance matrix $\Sigma$ and $\sdpp$  is the set of symmetric positive definite matrices. As explained in the introduction, this Gaussian variational family is particularly relevant in bandit framework according to Bernstein-Von Mises theorem.  
  \paragraph{Presentation of $\VITSI$:} As we will see, this choice of variational family will allow to derive an efficient method for solving \eqref{eq:varpost} using the Riemannian structure of $\varfam$. As noted in \cite{lambert2022variational}, $\varfam$ equipped with the Wasserstein distance of order 2 is a complete metric space as a closed subset of $\mathcal{P}_2(\R^d)$, the set of probability distributions with finite second moment. Recall that for two Gaussian distributions $p_0 = \N(\mu_0, \Sigma_0)$ and $p_1 = \N(\mu_1, \Sigma_1)$, their Wasserstein distance has a closed form: 
  \begin{align*} \small
     W_2^2\left(p_0, p_1\right) &=\left\|\mu_0-\mu_1\right\|^2 \\
     &+\operatorname{tr}(\Sigma_0
     +\Sigma_1-2(\Sigma_0^{1/2} \Sigma_1 \Sigma_0^{1/2})^{1/2})\eqsp.
 \end{align*}
 This Wasserstein distance on $\varfam$ allows to derive a Riemannian metric denoted $\grad$. The corresponding geodesic is given through the exponential map. More precisely, for a Gaussian distribution $p = \N( \mu_p, \Sigma_p)$, this map is defined as 
 \begin{align}\label{eq:expmap}
    \exp_p(\mu_v, \Sigma_v) =& (\mu_p + \mu_v +(\Sigma_v+\id) (\cdot  - \mu_p))_{\#} p \\
    =\nonumber& \N(\mu_p + \mu_v,\eqsp (\Sigma_v+\id)\eqsp \Sigma_p \eqsp(\Sigma_v+\id))\eqsp.
\end{align} 
With all these preliminaries, we can now present and motivate the algorithm developed in 
\cite{lambert2022variational} to efficiently solve \eqref{eq:varpost}. This method can be formalized as a Riemannian gradient descent scheme on $\varfam$. Firstly, we define the loss function $\mathcal{F}_t : p \rightarrow \KL(p| \post_t)$. Then, following \cite{lambert2022variational}, we derive the gradient operator of $\mathcal{F}_t$  on $\varfam$ equipped with $\grad$ as
\begin{equation}
\small
  \label{eq:def_grad}
    \nabla_{\grad} \mathcal{F}_t(p) = (\int  \nabla U_t(\theta) \D p(\theta),\eqsp \int \nabla^2 U_t(\theta) \D p(\theta) - \Sigma_{p}^{-1}) 
\end{equation}
where $\Sigma_p$ is the covariance matrix of $p$. From this expression, the corresponding Riemannian gradient descent \cite{bonnabel2013stochastic} using a step size $h_t >0$  defines the  sequence of iterates $\{\varpostexact_{t,k}\}_{k=1}^{K_t}$  recursively as: 
\begin{equation*}
\varpostexact_{t, k+1} = \exp_{\varpostexact_{t, k}}(- h_t \nabla_{\grad} \mathcal{F}_t(\varpostexact_{t, k}))\eqsp.
\end{equation*}
At each time step $t$, this sequence is initialized with variational posterior at the previous step, ie, $\varpostexact_{t, 0} = \varpostexact_{t-1, K_{t-1}}$. Please note that this warm initialization of the posterior results in an efficient algorithm and has been directly used in our main theoretical result (see \eqref{eq:phi_upp}). Combining \eqref{eq:expmap} and \eqref{eq:def_grad}, this recursion amounts  defining a  sequence of means $\{\mu_{t,k}\}_{k=1}^{K_t}  $ and covariance matrices $\{\Sigma_{t, k}\}_{k=1}^{K_t}$ by the recursions 
\begin{align*}
 &\mu_{t, k+1} = \mu_{t, k} - h_t \int \nabla U_t(\theta) \D \varpostexact_{t, k}(\theta),  \\
    &\Sigma_{t, k+1} = A_{t,k} \Sigma_{t, k}  A_{t,k},\\
    &\varpostexact_{t, k+1} = \N(\mu_{t,k+1},\Sigma_{t,k+1}) \\
    &\text{where} \quad A_{t,k}=\id - h_t (\int \nabla^2 U_t(\theta) \D \varpostexact_{t, k}(\theta) - \Sigma_{t, k}^{-1})
\end{align*}
 The main computational challenge in this recursion stems is that the integrals involved are typically intractable. To overcome this issue, we employ a Monte Carlo procedure to approximate these integrals. Subsequently, we consider a sequence of mean values denoted as $\{\meanvar_{t, k}\}_{k=1}^{K_t}$ and covariance matrices $\{\stdvar_{t, k}\}_{k=1}^{K_1}$ such that:
\begin{align*}
    &\meanvar_{t, k+1} = \meanvar_{t, k} - h_t \nabla U_t(\thetavar_{t, k}), \quad\stdvar_{t, k+1} =  \tilde{A}_{t,k}  \stdvar_{t, k} \eqsp  \tilde{A}_{t,k}   \\
    &\text{with} \quad \tilde{A}_{t,k}= \id - h_t (\nabla^2 U_t(\thetavar_{t, k})  - \stdvar_{t, k}^{-1})\eqsp, 
\end{align*}
where $\thetavar_{t, k} \sim \N(\meanvar_{t, k}, \stdvar_{t, k})$. Consequently, following \cite{lambert2022variational} we obtain an algorithm capable of addressing the problem defined in \eqref{eq:varpost}. However, this algorithm exhibits computational inefficiency, particularly in high-dimensional scenarios. This inefficiency arises from the necessity to sample from a Gaussian distribution with a non-diagonal covariance matrix during each updating step $k \in [K_t]$. As a result, it becomes impractical for use in a contextual bandit problem, where, at each time step $t$, we must solve the problem described in \eqref{eq:varpost}. This paper introduces an improved version of the earlier algorithm, designed to efficiently address the problem presented in \eqref{eq:varpost}. To achieve this, we begin by examining a sequence of matrices denoted as $B_{t, k}$, defined by the following 
\begin{equation}\label{eq:updates}
B_{t, k+1} =  \big\{\id - h_t   \nabla^2 U_t(\thetavar_{t, k})      \big\} B_{t, k} + h_t {(B_{t, k}^{-1})^{\top} }\eqsp.
\end{equation}
It is important to note that $B_{t, k}$ is a square-root matrix of the covariance of the variational distribution $\stdvar_{t, k}$, ie, $B_{t, k} B_{t, k}^{\top} = \stdvar_{t, k}$. Then we can sample efficiently from the variational distribution using $B_{t, k}$ with
   $ \tilde{\theta}_{t, k} = \meanvar_{t, k} + B_{t, k} \epsilon_{t, k} \eqsp, \quad \epsilon \sim \N(0, \id).$
   As a result, note that our method does not require any Cholesky decomposition, which has a complexity of $\mathcal{O}(d^3)$, contrary to the algorithm derived in \cite{lambert2022recursive} and also in LinTS. The updating strategies for the sequence of $\meanvar_{t, k}$ and $B_{t,k}$ are given by 
\begin{align*}
    &\meanvar_{t, k+1} = \meanvar_{t, k} - h_t \nabla U_t(\thetavar_{t, k}) \\
    &B_{t, k+1} =  \big\{\id - h_t   \nabla^2 U_t(\thetavar_{t, k})      \big\} B_{t, k} + h_t {(B_{t, k}^{-1})^{\top} }\\
    &\thetavar_{t, k} \sim \N(\meanvar_{t, k}, B_{t, k}^{\top} B_{t, k}) \eqsp.
\end{align*}
From this methodology, we can now complete the description of our first algorithm, referred to as \VITSI. At each step $t$, we consider the variational distribution $\varpost_{t} = \varpost_{t, K_t} = \N(\meanvar_{t, K_t}, B_{t, k}^{\top} B_{t, k})$ which approximates the solution of \eqref{eq:varpost}. Then, at round $t+1$, \VITSI~consists in sampling $\thetavar_{t+1}$ according to $\varpost_t$  and choosing  
\begin{equation}
    \label{eq:varpolicy}
    \pi^{\VITSI}_{t+1}(x) = \argmax_{a \in \A(x)} a^{\thetavar_{t+1}}(x) \eqsp.
\end{equation}
\vspace{-0.0cm}
As in TS, the likelihood function and the posterior distribution $\post_{t+1}$ are updated following equations \eqref{eq:likelihood} and \eqref{eq:post} using the new observed reward $r_{t+1}$ distributed according to $\Rrm(\cdot | x_{t+1},  a_{t+1})$ with $a_{t+1} =  \pi^{\VITSI}_{t+1}(x)$. The round $t+1$ is then concluded by solving $\varpost_{t+1} = \varpost_{t+1, K_{t+1}}$. The pseudo-code associated with this algorithm is given in Algorithm~\ref{alg:varts} and Algorithm~\ref{alg:update}.
\begin{algorithm}[H]
\begin{algorithmic}
\label{alg:VITS}
\caption{\VITS~algorithm}
\label{alg:varts}
\STATE $B_{1, 1} = \id / \sqrt{\lambda \eta}$ , $\stdstdvar_{1,1}=\id/(\eta \lambda)$ ,$\meanvar_{1, 1} \sim \N(0, \stdstdvar_{1,1})$
\FOR{$t = 1, \dots, T$}
\STATE receive $x_t \in \X$
\STATE sample $\thetavar_{t}$  from $\varpost_{t, K_t} = \N(\meanvar_{t, K_t}, B_{t, K_t}^{\top} B_{t, k})$
\STATE choose $a_t = \pivar(x_t)$ presented in \eqref{eq:varpolicy}
\STATE receive $r_t \sim \Rrm(\cdot|x_t, a_t)$
\STATE update $\varpost_{t+1, K_{t+1}}$ using Alg. \ref{alg:update} or \ref{alg:update_approx}. 
\ENDFOR
\end{algorithmic}
\end{algorithm}

        \begin{algorithm}[H]
\begin{algorithmic}
\caption{\VITSI}
\label{alg:update}
\STATE \textbf{Parameters}: step-size $h_t$, number of iterations $K_t$
\STATE $\meanvar_{t, 1} \leftarrow \meanvar_{t-1, K_{t-1}}$,$B_{t, 1} \leftarrow B_{t-1, K_{t-1}}$
\FOR{$k = 1, \dots, K_t$}
\STATE  draw  $\thetavar_{t, k} \sim \varpost_{t, k} = \N(\meanvar_{t, k}, B_{t, k}^{\top} B_{t, k})$
\STATE  $\meanvar_{t, k+1} \leftarrow \meanvar_{t, k} -h_t \nabla U_t(\thetavar_{t, k})$
\STATE
$ B_{t, k+1} \leftarrow
     \big\{\id-h_t \nabla^2(U_t(\thetavar_{t, k}))\big\} B_{t, k} 
     + h_t (B_{t, k}^{-1})^{\top} $
\ENDFOR
\end{algorithmic}
\end{algorithm}
\paragraph{Presentation of \VITSII:} In high dimension, the computational cost of the recursion of mean values and covariance matrices may be prohibitive since at each iteration $k \in [K_t]$, it requires inverting the matrix $B_{t, k}$. To tackle this computational issue, we propose a new version of $\VITS$. More precisely, the inverse of the square root covariance matrix $B_{t,k}^{-1}$ can be approximated using a first order Taylor expansion in $h_t$; see Appendix \ref{DLsequence} for more details. We denote by $C_{t, k}$ the approximation of $B_{t,k}^{-1}$, and we obtain recursions for the sequence of $\{C_{t, k}\}_{k \leq K_t}$ and $\{B_{t, k}\}_{k \leq K_t}$ such that: 
 \begin{align*}
     &C_{t,k+1} =  C_{t, k}  \{\id- h_t (C_{t, k}^{\top} C_{t, k} - \nabla^2 U_t(\thetavar_{t, k}))\}\eqsp,\\
     & B_{t, k+1} = (\id - h_t \nabla^2 U_t(\thetavar_{t, k})) B_{t, k} + h_t {C_{t, k}^{\top}}\eqsp.
 \end{align*}
 This trick reduces the complexity from $\mathcal{O}(d^3)$ to $\mathcal{O}(d^2)$ for the computation of the inverse. This version of $\VITS$ is referred to as $\VITSII$ and is given in Algorithm~\ref{alg:varts} and ~\ref{alg:update_approx}. 

\paragraph{Presentation of $\VITSIIHF$:}
The most computationally intensive step in $\VITSII$ remains the computation of the Hessian of $U_t$. In scenarios with a large number of data points and high dimensions, this step can become highly demanding. To avoid computing the Hessian of $U_t$, we suggest to use the following property of Gaussian distribution which is the result of a simple integration by part: 
 \begin{equation}\label{eq:gaussianipp}\small
   \int  \nabla^2 U_t  \rmd \N (\mu,\Sigma)  =
\int    
   \Sigma^{-1} (\id- \mu) \nabla U_t^{\top}   \, \rmd \N (\mu,\Sigma) \eqsp.
 \end{equation}
After approximating this right side integral using Monte Carlo, we derive a new sequence of square-root covariance matrix $\{B_{t, k}\}_{k \leq K_t}$ and inverse square-root covariance matrix $\{C_{t, k}\}_{k \leq K_t}$, defined recursively by:
 \begin{align*}
 \small
     &C_{t,k+1} =  C_{t, k}  \{\id- h_t (C_{t, k}^{\top} C_{t, k} -  A_{t, k})\}\eqsp,\\
     & B_{t, k+1} = (\id - h_t A_{t, k}) B_{t, k} + h_t {C_{t, k}^{\top}}\eqsp,
 \end{align*}
 where $A_{t, k} = C_{t, k}^{\top} C_{t, k} (\thetavar_{t, k}- \meanvar_{t, k}) \nabla U_t^{\top}(\thetavar_{t, k})$ and $\thetavar_{t, k} \sim \N(\meanvar_{t, k}, B_{t, k}^{\top} B_{t, k})$. This last version of $\VITS$ is referred to as $\VITSIIHF$ and its pseudo-code is given in Algorithm~\ref{alg:varts} and  Algorithm~\ref{alg:update_approx}.

\begin{algorithm}[]
\begin{algorithmic}
\caption{$\VITSII$ / $\VITSIIHF$}
\label{alg:update_approx}
\STATE \textbf{Parameters}: step-size $h_t$, number of iterations $K_t$
\STATE $\meanvar_{t, 1} \leftarrow \meanvar_{t-1, K_{t-1}}$, $B_{t, 1} \leftarrow B_{t-1, K_{t-1}}$
\FOR{$k = 1, \dots, K_t$}
\STATE draw $\thetavar_{t, k} \sim \varpost_{t, k} = \N(\meanvar_{t, k}, B_{t, k}^{\top} B_{t, k})$
\STATE$\meanvar_{t, k+1} \leftarrow \meanvar_{t, k} -h_t \nabla U_t(\thetavar_{t, k})$
\STATE  $A_{t, k} = \left\{
    \begin{array}{ll}
        \nabla^2(U_t(\thetavar_{t, k})) \hspace{-0.3cm} & \texttt{\textcolor{orange}{(H)}}\\
         C_{t, k}^2 (\thetavar_{t, k} - \meanvar_{t, k}) (\nabla U_t(\thetavar_{t, k}))^{\top}  \hspace{-0.3cm} & \textcolor{purple}{\texttt{(H free)}}
    \end{array}
\right. $
\STATE $B_{t, k+1} \leftarrow  \big\{\id - h_t A_{t, k}\big\} B_{t, k} + h_t {C_{t, k}^{\top}}$
\STATE  $C_{t, k+1} \leftarrow C_{t, k}  (\id- h_t (C_{t, k}^{\top} C_{t, k} - A_{t, k}))$
\ENDFOR
\end{algorithmic}
\end{algorithm}
where \textcolor{orange}{\texttt{(H)}} and \textcolor{purple}{\texttt{(H free)}} are for respectively Hessian and Hessian Free version. The computational complexity of all methods has been experimentally studied in a simple case, as discussed in Section \ref{sec:compt}.

\section{Main results}
\subsection{Linear Bandit}
In this section, we are interested in convergence guarantees for $\VITSI$ applied to the linear contextual bandit framework. 
This framework consists in assuming that $\rmR_{\theta}$ has form \eqref{eq:gen_lin_bandit} with $\lambdaref= \Leb$, $\mathrm{T}$ is the identity function and $\mathrm{h}$ and $g$ are specified by \eqref{eq:lin_bandit}: 
\begin{equation}
\label{eq:def_r_lin}
   \frac{\rmd R_{\theta}}{\rmd \Leb}(r |x,a) \propto \exp[\eta(r-\ps{\phi(x,a)}{\theta})^2/2] \eqsp. 
\end{equation} 
 Assumption on the reward kernel $\rmR$ is the following:
\begin{assumption}(Sub-Gaussian Reward Distribution)\label{as:subgaussian_rew}
There exists $R > 1$ such that for any $x \in \X$, $a \in \A(x)$, $\rho >0$,  $\log \int \exp\{\rho(r-  f(x, a))\} \Rrm(\D r |x,a) \leq  R \rho^2 \eqsp,$ where $f$ is defined in \ref{eq:def_mean_reward_distribution}
\end{assumption} 
We could only assume that $R >0$ in Assumption~\ref{as:subgaussian_rew} since if a distribution is $R$-sub-Gaussian, it is also $R'$-sub-Gaussian for any $R' \geq R$, however, we choose to set $R \geq 1$ to ease the presentation of our main results. We also assume that the model is well-specified. 
\begin{assumption}\label{ass:thetastar}
There exists $\thetastar$ such that $\rmR = \rmR_{\thetastar}$ and satisfying $\normdeux{\thetastar} \leq 1$.  Feature map $\phi$ satisfies  the boundedness condition.
\end{assumption} 
\begin{assumption}\label{ass:lowernorm}
For any contextual vector $x \in \R^d$ and action $a \in \A(x)$,  it holds that $\normdeux{\phi(x, a)} \leq 1$. 
\end{assumption}
Uniform boundedness condition on the feature map is relatively common for obtaining regret bounds for linear bandit problems \citep{agrawal2013thompson,xu2022langevin,pmlr-v108-kveton20a,abbasi:2011}. Note that Assumption \eqref{ass:lowernorm} is equivalent to $\sup_{x\in \msx\, , \, a \in \mca(a)} \normdeux{\phi(x, a)} \leq M_{\phi}$ for some arbitrary but fixed constant $M_{\phi} >0$, changing the feature map $\phi$ by $\phi/M_{\phi}$. Finally, we specify the prior distribution. 
\begin{assumption}
\label{ass:prior}
     The prior distribution is assumed to be zero-mean Gaussian distribution with variance $1/(\lambda\eta)$, where $\eta$ also appears in the definition $\rmR_{\theta}$ in \eqref{eq:def_r_lin}, 
\end{assumption}
While our theoretical results can readily be extended to accommodate a non-zero mean Gaussian prior, for the sake of simplicity, we have chosen to center the prior. Under Assumption~\ref{ass:prior}, combining \eqref{eq:post} and \eqref{eq:def_r_lin}, 
the negative log posterior $-\log \post_t$ denoted by $U_t$ is given by 
\begin{align} \label{def:Vt}
\textstyle    U_t(\theta) &= \frac{\eta}{2} \parenthese{\sum_{s=1}^{t-1} (\phi(a_s, x_s)^{\top} \theta - r_s)^2 + \lambda \Vert\theta\Vert_2^2}\\
    &=\frac{\eta}{2} (\theta^{\top} V_t \theta - 2\theta^{\top} b_t + \sum_{s=1}^{t-1} r_s^2)\eqsp,\\
    V_t = &\lambda \id_d + \sum_{s=1}^{t-1} \phi_s \phi_s^{\top} \in \R^{d \times d},  \eqsp b_t = \sum_{s=1}^{t-1} r_s \phi_s \in \R^{d \times 1}\nonumber.
\end{align}
Therefore, it follows that the gradient of $U_t$ is  given by $\nabla U_t(\theta) = \eta (V_t \theta - b_t)$ and its hessian matrix is equal to $\nabla^2 U_t(\theta) = \eta V_t$. Consequently, we recover the well-known fact that the posterior is a Gaussian distribution with mean $\meanpost_t = V_t^{-1} b_t$ and covariance matrix $\stdpost_t = (\eta V_t)^{-1}$.
 Denote by $\mathbb{\tilde{Q}}_{1:T}$ the distribution on the sequence of policies induced by the sequence of variational posterior $\{\varpost_t = \N(\meanvar_{t, K_t}, B_{t, K_t}^{\top} B_{t, K_t})\}_{t \in [T]}$ obtained with $\VITSI$.
We now state our main result on the cumulative regret associated to $\VITSI$ for linear contextual bandit, where a the proof is provided in Appendix \ref{sec:theo}. 
\begin{theorem}\label{theo:main}
Assume Assumptions \ref{as:subgaussian_rew} to \ref{ass:prior} hold. For the choice of hyperparameters $\{K_t,h_t\}_{t \in[T]}$ and $\eta$ specified in Section \ref{def:hyp},  for any $\delta \in (0, 1)$, with probability at least $1-\delta$, the cumulative regret is bounded by  
\begin{align*}  
    \CREG{\mathbb{\tilde{Q}}_{1:T}} 
  \leq \frac{CR^2 d\sqrt{d T} \log(3T^3)}{\lambda^2}  \log\Big(\frac{(1 + T/\lambda d )
  }{\delta}\Big)
\end{align*}
\end{theorem}
where $C \geq 0$ is a constant independent of the problem. Our main result shows that the distribution of the sequence of policies generated by $\VITSI$ results in a cumulative regret of order $\tilde{\mathcal{O}}(d\sqrt{dT})$. It is in the same order as the state-of-the-art cumulative regret obtained in \cite{agrawal2013thompson} for LinTS. 
The number of optimization steps $K_t$ we found are of order $\kappa_t^2\log(dT \log(T))$ where  $\kappa_t = \lmax(V_t) / \lmin(V_t)$. Following \citep{hamidi2020worst,wu2020stochastic}, if the diverse context assumption holds,  the condition number is $\kappa_t = \mathcal{O}(1)$. Therefore, under this previous assumption, $\VITSI$ requires a number of optimization steps that scale as $\log(d T\log(T))$. Finally,  \cite{xu2022langevin} derived similar bounds for TS using LMC for linear contextual bandit problems.
Although our proof is based on the linear case, it could be extended to more general cases insofar as our updates remain Gaussian by definition of the variational family. This allows the use of Gaussian (anti) concentration bound in the theoretical analysis. This is in contrast to other approximation methods, which do not possess this advantage.

\paragraph{Comparison table.} In this paragraph we have added a comparison table between Linear TS (LinTS), Linear UCB (LinUCB), Feel-Good TS \cite{fghuix, zhang2022feel}, $\VITSI$, $\VITSII$ (VITS-I/II), $\VITSIIHF$ (VITS-II HF), Langevin Monte Carlo TS (LMCTS) and Variational TS (VTS). The column "Regret" corresponds to the theoretical regret bound obtained by the algorithm. "Complexity" is the computational complexity, more precisely the symbol $(++)$ corresponds to a regret $O(\sqrt{dT})$, $(+)$ to $O(d^{3/2}\sqrt{T})$ and $(-)$ to no existing regret bound.  " Linear" is set to Yes when the algorithm is designed only for the Linear Bandit setting and No for general setting including Linear. The "Conditioning" column describes the algorithm's robustness against the conditioning of the problem. 

\begin{table}[H]
\footnotesize
    \begin{tabular}{  c | c | c | c | c }
          & Regret & Complexity &  Linear & Conditioning \\
         \hline \hline
         LinTS & + & ++ & Yes & ++\\
         \hline
         LinUCB & ++ & ++ & Yes & ++\\
         \hline
         FG-TS & ++ & & No & \\
         \hline VITS-I/II & + & + & No & + \\
         \hline VITS-II HF & - & + + & No & + \\
         \hline LMC-TS & + & ++ & No &  - \\
         \hline VTS & - & - & No & \\
        \hline
    \end{tabular}
\end{table}

\section{Numerical experiments}\label{sec:exp}
\subsection{Linear and quadratic bandit}
Our initial investigation focused on a toy setting where contextual vectors are sampled from a Gaussian distribution. However, in this specific setting, the contextual vectors exhibit high diversity, resulting in a posterior covariance matrix with a condition number of $O(1)$. This condition makes the optimization problem overly simplistic, as a result, all approximation methods seem to perform identically in this simple well-conditioned problem. So we introduce a novel setting in which the diversity of arms is controlled by a parameter, denoted as $\zeta$. Firstly, we consider a fixed pool of arms denoted as $P = [\tilde{x}_1, \ldots, \tilde{x}_{n}]$ with $n = 50$, where each arm $\tilde{x}_i$ follows a normal distribution $\N(0_d, \id_d)$. This fixed pool is relevant in real-world scenarios, such as in a Recommender system, where this pool corresponds to the concept of a meta-user. Then, at each step $t \in [T]$, for every arm, we randomly sample a vector $\tilde{x}_i$ from the pool $P$, and the contextual vector associated with this arm is defined as $x = \tilde{x}_i + \zeta \epsilon$, where $\epsilon \sim \mathcal{N}(0_d, \id_d)$. When $\zeta$ has a high value, the corresponding user is far from the meta-user. Consequently, the diversity among arms is high, resulting in a well-conditioned problem. However, in cases where $\zeta$ is low, the problem is ill-conditioned and the optimization becomes challenging. 

 We consider the linear bandit and the quadratic bandit problems. In both settings, the bandit environment is simulated using a random vector $\theta^{\star}$ sampled from a normal distribution $\N(0_d, \sigma^{\star} \id_d)$. We opted for $\sigma^{\star} = 1/d$ to ensure that the variance of the scalar product $x^{\top} \theta^{\star}$ remains independent of the dimension $d$. The parameter dimension $d$ is set to $20$ and we consider a number of arms $K = 50$. In the linear bandit setting, the reward associated with the contextual vector $x$, is $r = x^{\top} \thetastar + \alpha \epsilon $ where $\epsilon \sim \N(0_d, \id_d)$. However, to maintain problem complexity independent of $\zeta$, we have set the signal-to-noise ratio to a fixed value of 1, meaning $\E[(x^{\top} \theta^{\star})^2] / \E[(\alpha \epsilon)^2] = 1$. This implies that  $\sqrt{1 + \zeta^2} = \alpha$. See Appendix \ref{sec:simu} for more details about the setting. In these experiments, we have chosen to compare $\VITSII$, $\VITSIIHF$, Linear TS (LinTS), and LMC-TS, with 10 and 50 iterations of Langevin diffusion at each step. For $\VITS$  based algorithm, we have only used 10 updating steps. We have omitted the performance of $\VITSI$ since it experimentally performs identically to $\VITSII$. For the algorithm $\VITSIIHF$, we approximate the integral presented in \eqref{eq:gaussianipp} using 20 Monte Carlo samples. This choice is made due to the observed instability caused by the Monte Carlo error when considering high values of $\eta$. However, in our setting, even with 20 Monte Carlo samples, $\VITSIIHF$ remains a faster method compared to $\VITSII$. We also attempted to assess the performance of VTS, but, in the ill-conditioned setting, it exhibited a linear and notably high cumulative regret. Consequently, we have opted to exclude it from the figure for the sake of clarity and visibility. The mean and standard error are reported for all experiments over 50 runs. The hyperparameter is provided in Appendix \ref{sec:hypergrid}. 

\begin{figure}[H]
    \centering
    \includegraphics[width=0.49\columnwidth,scale=1.3]{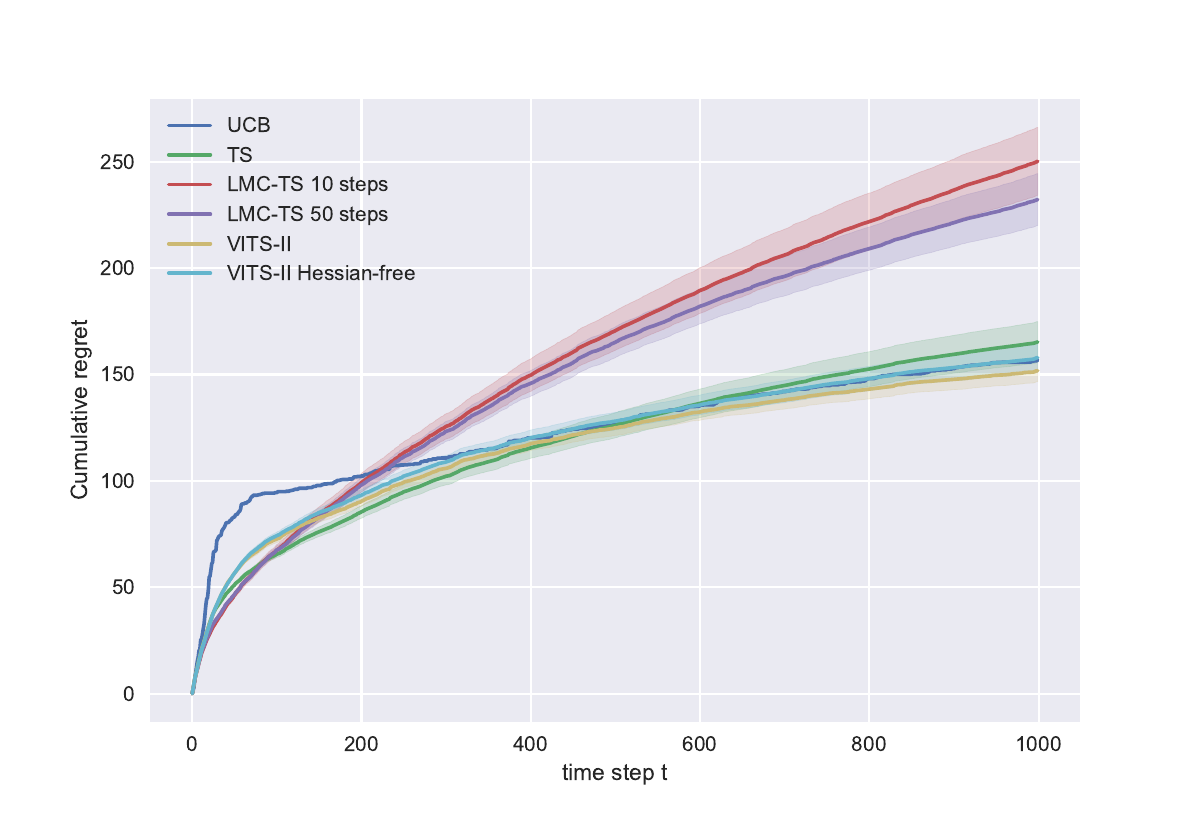}
    \includegraphics[width=0.49\columnwidth,scale=1.3]{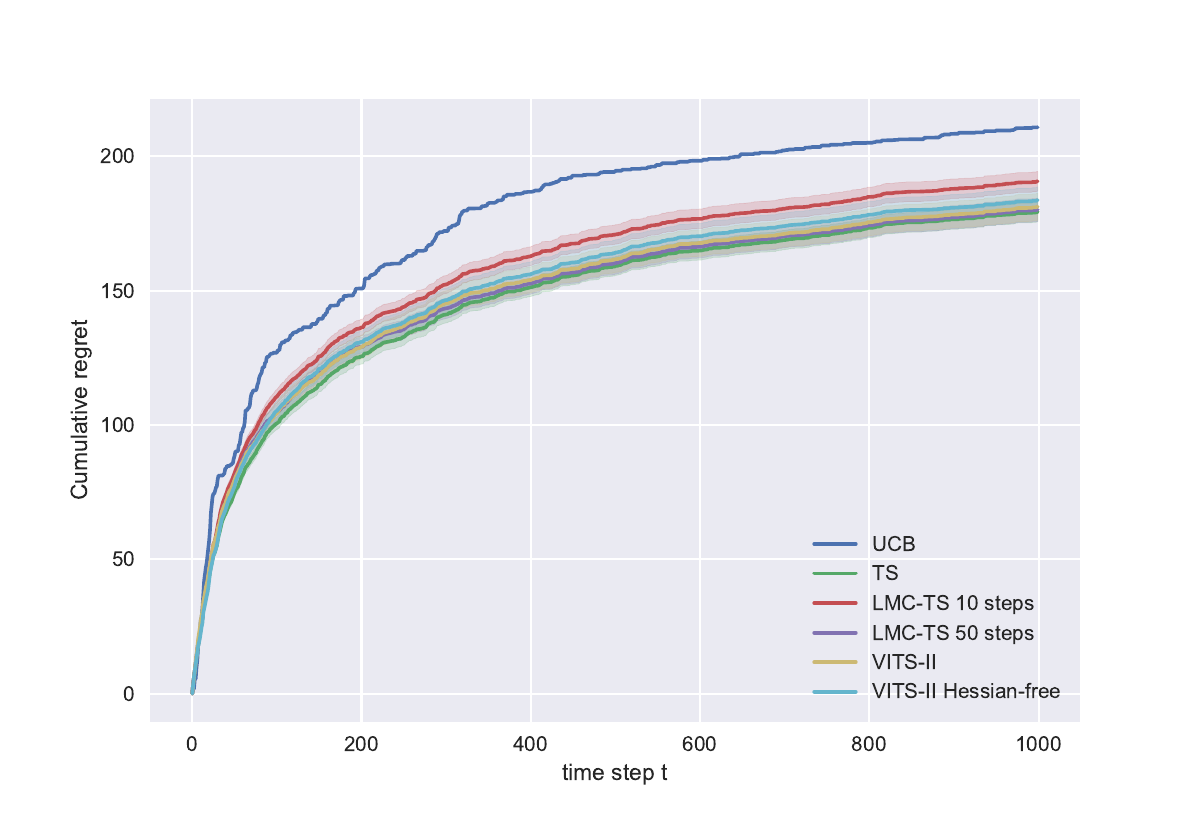}
    \caption{Linear bandits, $\zeta=0.1$ (left),  $\zeta=1$ (right).}
    \label{fig:linearbandit}
\end{figure}

Figure \ref{fig:linearbandit} illustrates the cumulative regret with respect to the time step $t$ for a well-conditioned problem ($\zeta = 1$) and a ill-conditioned problem ($\zeta = 0.1$). Firstly, for $\zeta=1$, it appears that all methods exhibit similar performance, with the exception of LMC-TS with 10 steps, which slightly underperforms. However, for $\zeta = 0.1$, the optimization problem becomes harder and LMC-TS underperforms even with $50$ Langevin steps. This behaviour was expected in our setting, because LMC requires a lot of iterations to converge to the posterior compared to VI.  A more complete explanation of this phenomenon can be found in Appendix \ref{sec:compar}. Finally, we can conclude that $\VITSII$ performs similarly to LinTS and that its \textcolor{purple}{\texttt{Hessian-free}} version slightly underperforms but is computationally more efficient. 

\begin{figure}[H]
    \centering
\includegraphics[width=0.49\columnwidth,scale=1.2]{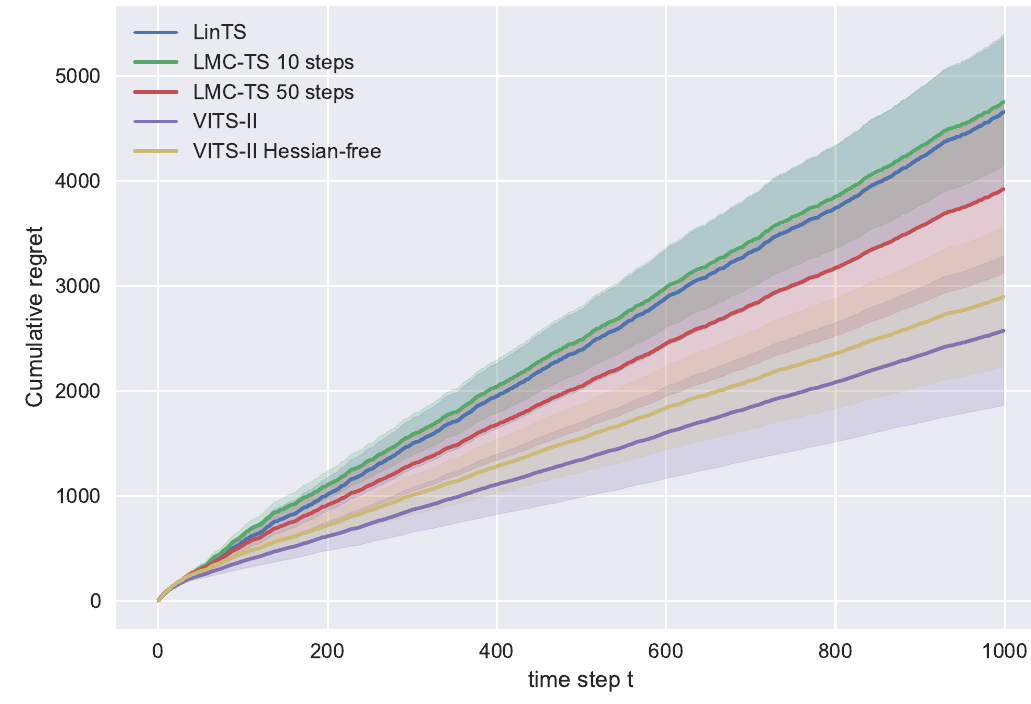}
\includegraphics[width=0.49\columnwidth,scale=1.2]{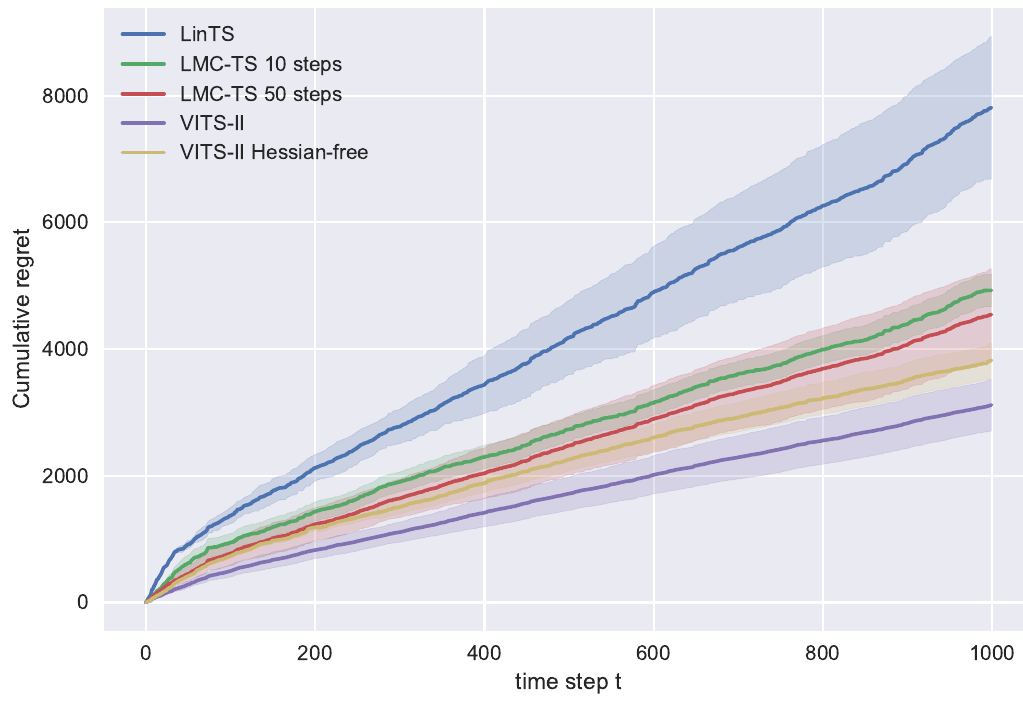}
    \caption{Quadratic bandit, $\zeta=0.1$(left), $\zeta=1$(right). }
    \label{fig:quadraticbandit}
\end{figure}

For Quadratic bandit in Fig \ref{fig:quadraticbandit}, the reward is $r = (x^{\top} \thetastar)^2 + \alpha \epsilon $. This setting is similar to the Linear setting, but we ensure the condition $\E[(x^{\top} \theta^{\star})^4] / \E[(\alpha \epsilon)^2] = 1$ to still get the signal-to-noise ratio equals to $1$. This implies a slight different condition $\alpha= (\zeta^2 +1) \sqrt{3 +6/d}$, see Appendix \ref{sec:simu}. Moreover, a simple MLP with two hidden layers of $20$ neurons is used for LMC, $\VITSII$, and its \textcolor{purple}{\texttt{Hessian-free}} version as neural network architecture. Performance in Fig \ref{fig:quadraticbandit} are similar to linear bandits where  $\VITSII$  slightly performs better than its \textcolor{purple}{\texttt{Hessian-free}} version but outperforms both LMC and LinTS algorithms as LinTS is not adapted for this setting. The gap between LMC and our algorithm is smaller in the well-conditioned setting than in the ill-conditioned, which was also expected. Finally, additional experience on non-contextual bandits can also be found in Appendix \ref{sec:add_simu}. 
\section{MovieLens Dataset}
In this section, we evaluate $\VITS$ on the MovieLens dataset, consisting of one million ratings by 6040 users for 3952 movies. We adopt the setup proposed in \cite{aouali2022generalizing}, involving a low-rank factorization of the rating matrix to yield 5-dimensional representations for users ($x_j \in \R^5$) and movies ($\theta_i \in \R^5$). Movies are treated as potential actions, and context $x_t$ is uniformly sampled from the pool of user vectors. We consider logistic rewards, sampled from $\text{Ber}(\mu(x_j^{\top} \theta_i))$, where $\mu$ is the sigmoid function. We conduct 50 simulations, each involving 100 randomly selected movies.Our prior distribution employs a Gaussian distribution with mean $\mu_0$ and covariance $\Sigma_0 = \text{diag}(\sigma_0)$. Here, $\mu_0$ and $\sigma_0$ represent the mean and variance of movie vectors across all dimensions. This setting deviates somewhat from our theoretical framework, where we consider a unified posterior distribution for all arms using a feature map function $\phi$ representing context-action pairs. In the MovieLens context, each arm possesses an individual posterior distribution. These two settings closely align when the feature map is the vector concatenation function. In practice, we can apply $\VITS$ or LMC at each arm to obtain posterior samples. In this experiment, we compare LinTS against LMC-TS, $\VITSII$, and the $\VITSIIHF$ variant. LMC-TS uses 10 Langevin updating steps. It's crucial to note that for each time step $t$ and each arm $a$, LMC-TS requires running Langevin diffusion to obtain a new parameter with low correlation to the previous one. This leads to a high computational complexity for LMC-TS. In contrast, $\VITS$ for each arm only involves sampling from a low-dimensional Gaussian distribution and updating the variational posterior corresponding to the chosen arm. This approach offers significant computational efficiency. 

\begin{figure}[H]
    \centering
    \includegraphics[scale=0.3]{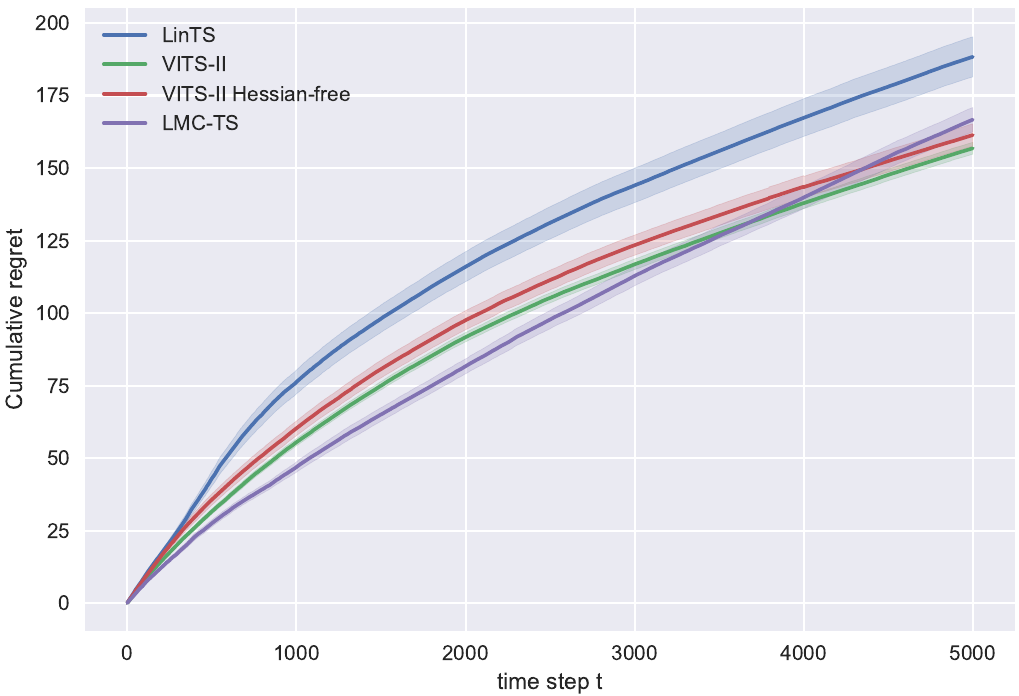}
    \caption{Cumulative regret for MovieLens dataset.}
    \label{fig:movielens}
\end{figure}
Figure \ref{fig:movielens} reveals that LinTS is ill-suited for this particular task, as it assumes rewards to be linear while the approximated algorithms outperform LinTS, as they specifically target the logistic posterior. Remarkably, $\VITS$ appears to slightly outperform LMC-TS, despite its computational efficiency advantages. 
\section{Conclusion and perspectives}

This paper presents a novel TS algorithms called $\VITSI$, that
use VI as an approximation method. This algorithms provide robust theoretical guarantees, in particular a
cumulative regret bound of $\tilde{\mathcal{O}}(d \sqrt{dT})$ in the linear setting. 
One limitation of our theoretical analysis is that the regret bound derived is limited to the linear setting while the interest of our algorithm relies on nonlinear tasks.  Additionally, we introduce two other algorithms, named $\VITSII$ and $\VITSIIHF$, which offer enhanced computational efficiency. The $\VITSII$ algorithm uses a Taylor expansion to remove the computation of inverse matrix and the $\VITSIIHF$ algorithm removes the computations of Hessian, resulting in faster execution. Finally, all algorithms have been extensively evaluated in both simulated and real problems.


\section*{Acknowledgements}
 Pierre Clavier has been supported by a grant from Région Île-de-France. Tom Huix acknowledge the support of ANR-CHIA-002, ”Statistics, computation and Artificial Intelligence”. Part of the work has been developed under the auspice of the Lagrange Center for Mathematics and Calculus. This work was granted access to the HPC resources of IDRIS under the allocation   AD011014442 and AD011013313R2 made by GENCI (Grand Equipement
National de Calcul Intensif) Alain Durmus is Funded by the European Union (ERC, Ocean, 101071601).
Views and opinions expressed are however those of the
author(s) only and do not necessarily reflect those of
the European Union or the European Research Council
Executive Agency. Neither the European Union nor the
granting authority can be held responsible for them.
\newpage

\section*{Impact Statement}  This paper presents a work whose goal is to advance the field of Machine Learning. There are many potential societal consequences of our work, none of which we feel must be specifically highlighted here.

\bibliography{example_paper}
\bibliographystyle{icml2024}

\newpage
\appendix
\onecolumn

\section{Proof of the regret bound}\label{sec:theo}
\subsection{Proof of Theorem \ref{theo:main}}

While \cite{lambert2022variational} establishes quantitative bounds on the bias introduced by Algorithm~\ref{alg:update} for the VI of the posterior. Combining this result with the one derived in  \cite{agrawal2013thompson} for TS leads to sub-optimal regret bounds. It is similar to LMC-TS \cite{xu2022langevin} which had to make a clever adaptation of \cite{agrawal2013thompson}. Similar to this work, we need here to revise the proof of \cite{agrawal2013thompson} to $\VITS$. We give in this section the main steps of our proofs.  Each step is based on Lemmas which are stated and proved in the next sections. First,  we define the filtration $(\filt_t)_{t \in \{0,\ldots,T-1\}}$ such that for any $t \in [T]$, $\filt_{t-1}$ is the $\sigma$-field generated by $\hist_{t-1}$ and $x_t$ where $\hist_{t-1} = \{(x_s, a_s, r_s)\}_{s\leq t-1}$  is the observations up to $t-1$ and $x_t$ is the contextual vector at step $t$. 
For some feature map  $\phi : \X\times \msa \rightarrow \mathbb{R}$ and 
for any $t \in [T]$, we denote by  
\begin{equation*}
\phistar_t = \phi(x_t, \as_t)  \eqsp, \, \phi_t = \phi(x_t, a_t) \eqsp,
\end{equation*}
the features vector of the best arm $\as_t$ and the features vector of the arm $a_t$ chosen by $\VITS$ at time $t$ respectively.
the difference between the best expected reward and the expected reward obtained by $\VITS$ is denoted by
\begin{equation*}
    \Delta_t = \phistar_t \thetastar - \phi_t^{\top}\thetastar\eqsp.
\end{equation*}
At each round $t \in [T]$, we consider the set of saturated arms $\mathcal{S}_t$ and unsaturated arms $\mathcal{U}_t$  defined by 
\begin{equation}\label{def:saturated}
    \mathcal{S}_t = \bigcap_{a \in \A(x_t)}\{ \Delta_t(a) > g(t) \eqsp \Vert\phi(x_t, a)\Vert_{\Vinv_t}\} \eqsp,
\end{equation}
and $\mathcal{U}_t=\A(x_t) \backslash\ \mathcal{S}_t$ where $\Vinv_t$ is defined in \eqref{def:Vt} and
\begin{equation}
    \label{eq:def_g}
    g(t) = C R^2 d \sqrt{\log(t) \log(T)} / \lambda^{3/2} \eqsp,
    \nonumber 
\end{equation}
for some constant $C \geq 0 $ independent of $d$, $t$ and $T$. In addition, consider the events $\Etrue_t$ and $\Evar_t$ such that
\begin{equation*}
  \bigcap_{a \in  \A(x_t)}\{ \vert\phi(x_t, a)^{\top}\meanpost_t - \phi(x_t, a)^{\top} \thetastar\vert 
\leq \eqsp g_1(t)\eqsp\eqsp \Vert\phi(x_t, a)\Vert_{\Vinv_t}\} \subset \Etrue_t
\end{equation*}
\begin{equation*}
    \Evar_t =  \bigcap_{  a  \subset \A(x_t)}\{ \vert\phi(x_t, a)^{\top}\thetavar_t - \phi(x_t, a)^{\top} \meanpost_t\vert 
    \leq \eqsp g_2(t)\eqsp \eqsp \Vert\phi(x_t, a)\Vert_{\Vinv_t} \} \eqsp,
\end{equation*}
where $\hat{\mu}_t$ is given by $\hat{\mu}_t = \Vinv b_t$ and $b_t$ is given in \eqref{def:Vt}.
The specific definitions of  $\Etrue_t$, $g, g_1$ and $g_2$ are given in Section \ref{sec:def} of the supplementary.  Nevertheless, by definition, it holds that $g_1(t)+g_2(t) \leq g(t)$.

\begin{enumerate}
\item For ease of notation, the conditional expectation $\E_{\pi_{1:T} \sim \mathbb{Q}_{1:T}}[\cdot]$ and probabilities $\prob_{\pi_{1:T} \sim \mathbb{Q}_{1:T}}(\cdot)$ with respect to the $\sigma$-field $\filt_{t-1}$ are denoted by $\E_t[\cdot]$ and $\prob_t(\cdot)$ respectively. 
Therefore, with these notations, we have by  definition of the cumulative regret:
$$\CREG{\mathbb{\tilde{Q}}_{1:T}} 
     = \sum_{t=1}^{T} \Delta_t \eqsp.$$ 
We now bound for any $t \in [T]$, with high probability, $\Delta_t(a_t)$.
To this end, in the next step of the proof, we show that the stochastic process $(X_t)_{t \in [T]}$ defined below is a $(\filt_t)_{t \in [T]}$ super-martingale.
$$X_t = \sum_{s=1}^t Y_s$$ with $$Y_s = \Delta_s - c g(s)\Vert\phi_s\Vert_{\Vinv_s}/p - 2/s^2\eqsp,$$
where $p \in (0,1)$ and $c$ is a sufficiently large real number, independent of $d$, $T$ and $s$.

\item \textbf{Showing that $(X_t)_{t \in [T]}$ is a super-martingale.}
  We consider the following decomposition
  \begin{align}
  \label{eq:sketch_proof_0}
    \E_t[\Delta_t(a_t)] &= \E_t[\Delta_t(a_t) \1_{\Etrue_t}]  +\eqsp\E_t[\Delta_t(a_t)| \Etruebar{t}]\prob_t(\Etruebar{t})
    \nonumber\\
    &\leq \E_t[\Delta_t(a_t) \1_{\Etrue_t}] + \prob_t(\Etruebar{t}) \eqsp,
\end{align}
where we used for the last inequality that $\normdeux{\thetastar} \leq 1$ and Assumption~\ref{ass:lowernorm}.
Then, since $\Etrue_t \in \filt_{t-1}$, we have,
\begin{align}
    \nonumber
    \E_t[\Delta_t(a_t) \1_{\Etrue_t}] &= \1_{\Etrue_t}\E_t[\Delta_t(a_t)| \Evar_t]\prob_t(\Evar_t) 
    + \1_{\Etrue_t}\E_t[\Delta_t(a_t)|\Evarbar{t}]\prob_t(\Evarbar{t})\\
    &\leq \1_{\Etrue_t}[\E_t[\Delta_t(a_t)|\Evar_t] + \prob_t(\Evarbar{t})]
    \label{eq:sketch_proof_1}
\end{align}
where in the last line we have used that $\Delta_t(a_t) \leq 1$ again.
Denote by $\ba_t = \argmin_{a \in \mathcal{U}_t} \Vert\phi(x_t, a)\Vert_{\Vinv_t}$ and  $\phibar_t =  \phi(x_t, \ba_t)$.  Then, given  $\Etrue_t$ and $\Evar_t$ we have
\begin{align}
    \Delta_t(a_t) &= \phistar_t^{\top}\thetastar - \phi_t^{\top}\thetastar\nonumber\\
    &= \phistar_t^{\top}\thetastar - \phibar_t^{\top}\thetastar +  \phibar_t^{\top}\thetastar - \phi_t^{\top}\thetastar \nonumber\\
    &\stackrel{(a)}{\leq} g(t) \Vert\phibar_t\Vert_{\Vinv_t} + \phibar_t^{\top}\thetastar - \phi_t^{\top}\thetastar \nonumber\\
    &\stackrel{(b)}{\leq} g(t) \Vert\phibar_t\Vert_{\Vinv_t} + (\phibar_t^{\top} \thetavar_t + g(t) \Vert\phibar_t\Vert_{\Vinv_t}) \nonumber
    - (\phi_t^{\top} \thetavar_t - g(t) \Vert\phi_t\Vert_{\Vinv_t}) \nonumber
    \label{eq:sketch_proof_2} \\
    &\stackrel{(c)}{\leq}  (2 \Vert\phibar_t\Vert_{\Vinv_t} + \Vert\phi_t\Vert_{\Vinv_t})g(t)
\end{align}
where inequality $(a)$ is due to $\ba_t \in \mathcal{U}_t$, and therefore $\Delta_t(\ba_t) \leq g(t) \Vert\phibar_t\Vert_{\Vinv_t}$, 
inequality $(b)$ uses that given $\Etrue_t$ and $\Evar_t$, for any $\phi \in \R^d$, $|\phi^{\top} \thetavar_t - \phi^{\top} \thetastar|\leq g(t) \Vert\phi\Vert_{\Vinv_t}$ since by definition $g_1(t)+g_2(t) \leq g(t)$; finally, the arm $a_t$ maximizes the quantity $\phi(x_t, a_t)^{\top} \thetavar_t$, $\phibar_t^{\top} \thetavar_t - \phi_t^{\top}\thetavar_t$ is obviously negative, which implies inequality $(c)$.

Moreover, given  $\Etrue_t$ and $\Evar_t$,
\begin{align}
    \E_t[\Vert\phi_t\Vert_{\Vinv_t}]  \nonumber\nonumber &=\E_t[\Vert\phi_t\Vert_{\Vinv_t}| a_t \in \mathcal{U}_t] \prob_t(a_t \in \mathcal{U}_t)\nonumber+\E_t[\Vert\phi_t\Vert_{\Vinv_t}|a_t \in \mathcal{S}_t] \prob_t(a_t \in \mathcal{S}_t)\\ \nonumber
    &\stackrel{(a)}{\geq} \Vert\phibar_t\Vert_{\Vinv_t}\prob_t(a_t \in \mathcal{U}_t) \nonumber \\
    &\stackrel{(b)}{\geq}  (p - 1/t^2)\Vert\phibar_t\Vert_{\Vinv_t}
    \nonumber
\end{align}
where $(a)$ is due to the definition of $\phibar_t$, i.e. for any $a \in \mathcal{U}_t, \Vert\phibar_t\Vert_{\Vinv_t} \leq \Vert\phi(x_t, a)\Vert_{\Vinv_t}$, and $(b)$ uses Lemma \ref{lemma:unsaturatedprob} with $p \in (0,1)$. Here is one of the main differences with the proof conducted by \cite{agrawal2013thompson}. Indeed, to obtain such a bound, we need to carefully dig into the convergence of the the sequence of means $\{\meanvar_{t, K_t}\}_{k \in [1, K_t]}$ and covariance matrices $\{ \stdvar_{t, K_t} \}_{k \in [1, K_t]}$ to obtain a fine-grained analysis of the distribution of $\varpost_t$.
Therefore, using equations \eqref{eq:sketch_proof_1} and \eqref{eq:sketch_proof_2}
\begin{align*}
 &\1_{\Etrue_{t}}   \E_t[\Delta_t(a_t)] 
    \leq \Big(\frac{2}{p - 1/t^2} + 1\Big) g(t)\E_t[\Vert\phi_t\Vert_{\Vinv_t}] + \frac{1}{t^2}
    \leq \frac{cg(t)}{p}\E_t[\Vert\phi_t\Vert_{\Vinv_t}] + \frac{1}{t^2}\eqsp,
\end{align*}
where $c$ is a sufficiently large real number independent of the problem. Plugging this bounds in \ref{eq:sketch_proof_0}, we obtain
 
\begin{align*}
    \E_t[\Delta_t(a_t)] &\leq \frac{c g(t)}{p} \E_t[\Vert \phi\Vert_{\Vinv_t}] + \frac{1}{t^2} + \prob_t(\Etruebar{t}) 
\end{align*}
Applying Lemma \ref{lemma:etrue} yields
\begin{equation*}
    \E_t[\Delta_t(a_t)] \leq \frac{c g(t)}{p} \E_t[\Vert \phi\Vert_{\Vinv_t}] + \frac{2}{t^2} \eqsp.
\end{equation*}
This is another important difference with the original proof of \cite{agrawal2012analysis} which uses our precise convergence study for $\{\meanvar_{t, K_t}\}_{k \in [1, K_t]}$.
Then, it follows that $(X_t)_{t\in[T]}$ is a $(\filt_t)_{t \in[T]}$-super martingale.

\item \textbf{Concentration for $(X_t)_{t \in [T]}$.}
Note that $(X_t)_{t \in [T]}$ is a super-martingale with bounded increments: for any $t \in [T]$
\begin{align*}
|X_{t+1} - X_t| &= |Y_{t+1}|\\
&= |\Delta_t(a_t) - \frac{c g(t)}{p} \Vert \phi_t \Vert_{\Vinv_t} - \frac{2}{t^2}| \\
&\stackrel{(a)}{\leq} |\Delta_t(a_t)(a_t) - \frac{c g(t)}{\sqrt{\lambda} p} - \frac{2}{t^2}| \\
&\leq \frac{3 c g(t)}{\sqrt{\lambda} p}\eqsp,
\end{align*}

where in $(a)$ we have used that $$\Vert\phi_t\Vert_{\Vinv_t} \leq \Vert\phi_t\Vert_{\Vinv_1} \leq 1/\sqrt{\lambda}$$ and inequality $(b)$ is due to $\Delta_t(a_t) \leq 1$, $2 / t^2 \leq 2$ and $3cg(t) / (p\sqrt{\lambda}) > 2$ for an appropriate choice of the numerical constant $c$. Therefore, applying Azuma-Hoeffding inequality (Lemma \eqref{lemma:azuma}), with probability $1 - \delta$ it holds that
\begin{align*}
X_T \leq \sqrt{2 \log(1/\delta) \sum_{s=1}^T \frac{9 c^2 g(s)^2}{p^2 \lambda}}
\leq \sqrt{18 \log(1/\delta) \frac{c^2 }{p^2 \lambda} g(T)^2 T}\eqsp,
\end{align*}
using that $g(T) \geq g(t)$.
\item \textbf{Conclusion.}
The super-martingale $(X_t)_{t=1}^T$ is directly linked to the cumulative regret by
\begin{align*}
    X_T &= \sum_{t=1}^{T} Y_t\\
    &= \sum_{s=1}^{T} \Delta_t - cg(t) \Vert \phi_t\Vert_{V_t^{-1}} / p - 2/t^2 \\
    &= \CREG{\mathbb{\tilde{Q}}_{1:T}}  - \sum_{t=1}^T cg(t) \Vert \phi_t\Vert_{V_t^{-1}} / p + 2/t^2
\end{align*}
then taking the expectation and using  the super-martingale previous argument of the  proof, we obtain the following upper bound for the cumulative regret :
\begin{align*}
\CREG{\mathbb{\tilde{Q}}_{1:T}} &\leq \sum_{t=1}^T \frac{c g(t)}{p} \Vert\phi_t\Vert_{\Vinv_t} 
+ \sqrt{18 \log(1/\delta) \frac{c^2}{p^2 \lambda} g(T)^2 T}+  \frac{\pi^2}{3}\eqsp.
\end{align*}

using that $\sum_{t=1}^{+\infty} 1/t^2 \leq \pi^2/6$. As a result, applying Lemma \ref{lemma:upperboundsum} yields
\begin{align*}
&\CREG{\mathbb{\tilde{Q}}_{1:T}} \leq \frac{c g(T)}{p}\sqrt{2 d T \log\Big(1 + T/(\lambda d})\Big)
+ \frac{c g(T)}{p \sqrt{\lambda}} \sqrt{18 \log(1/\delta) T}+  \frac{\pi^2}{3}\eqsp.
\end{align*}

Using the definition of $g(T)$ in \eqref{eq:def_g}, we get 
\begin{align*}
    \CREG{\mathbb{\tilde{Q}}_{1:T}} 
    \leq \frac{C R^2 d}{\lambda^2} \log(3T^3) \sqrt{dT \log\big(1 + T/(\lambda d)\big) \log\big(1/\delta\big)}\eqsp,
\end{align*}

where $C \geq 0$ is a constant, independent of the problem, which completes the proof.
\end{enumerate}

\subsection{Hyperparameters choice and values}\label{def:hyp}
In this section,  we define and discuss the values of the main hyperparameters.

\paragraph{\textbf{Parameter} $\bm{\eta:}$} is the inverse of the temperature. The lower is $\eta$, the better is the exploration. It is fixed to
\begin{equation}\label{def:eta}
      \eta = 4\lambda^2/(81R^2 d \log(3T^3)) \leq 1
\end{equation}

\paragraph{\textbf{Parameter} $\bm{\lambda:}$} is the inverse of the standard deviation of the prior distribution. It controls the regularization. The lower is $\lambda$, the better is the exploitation. This parameter is fixed but lower than 1.

\paragraph{\textbf{Parameter} $\bm{h_t:}$}  is the step size used in all Algorithms. It is fixed to
\begin{equation}\label{def:ht}
    h_t = \lmin(V_t)/(2\eta (\lmin(V_t)^2 + 2\lmax(V_t)^2))
\end{equation}
\paragraph{\textbf{Parameter} $\bm{K_t:}$} is the number of gradient descent steps performed. It is fixed to
 \begin{equation}\label{def:kt}
     K_t = 1 + 2 (1 + 2\kappa_t^2) \log\Big(2R\kappa_t d^2 T^2\log^2( 3T^3)\Big)  \eqsp.
 \end{equation}
Therefore the number of gradient descent steps is $K_t \leq \mathcal{O}(\kappa_t^2\log(dT\log(T)))$.

\subsection{Useful definitions}\label{sec:def}
\begin{definition}\label{def:paramalgo}\textbf{(Variational approximation)}
Recall that $\post_t(\theta) \propto \exp(-U_t(\theta))$ is the posterior distribution. And $\varpost_t$ is the variational posterior distribution in the sense that
\begin{equation*}
    \varpost_t = \argmin_{\mathrm{p} \in \varfam} \KL(\mathrm{p}|\post_t)\eqsp,
\end{equation*}
where $\varfam$ is a variational family. In this paper we focus on the Gaussian variational family and we denote by $\meanvar_t$ and $B_t$ respectively the mean and the square root covariance matrix of the variational distribution, ie, 
\begin{equation*}
    \varpost_t = \N(\meanvar_t, B_t B_t^{\top})\eqsp.
\end{equation*}
The values of $\meanvar_t$ and $B_t$ are obtained after running $K_t$ steps of algorithm \ref{alg:update} or \ref{alg:update_approx}. Note that the sequence of means $\{\meanvar_{t, k}\}_{k=1}^{K_t}$ is defined recursively by
\begin{align*}
\meanvar_{t, k+1} &= \meanvar_{t, k} - h_t \nabla U_t(\thetavar_{t, k})\\
&=\meanvar_{t, k} - h_t \eta V_t(\thetavar_{t, k} - \meanpost_t)
\end{align*}

where $\thetavar_{t, k} \sim \N(\meanvar_{t, k}, B_{t, k}^{\top} B_{t, k})$ and we have used that $\nabla U_t(\theta) = \eta (V_t \theta - b_t)$ (see equation \eqref{def:Vt}). Consequently, $\meanvar_{t, k}$ is also Gaussian and we denote by $\meanmeanvar_{t, k}$ and $\stdstdvar_{t, k}$ its mean and covariance matrix, ie, $\meanvar_{t, k} \sim \N(\meanmeanvar_{t,k}, \stdstdvar_{t, k})$. 
Furthermore, the sequence of square root covariance matrix $\{B_{t, k}\}_{k=1}^{K_t}$ is defined recursively in Algorithm \ref{alg:update} by
\begin{align*}
    B_{t, k+1} &= \left\{\id - h_t \nabla^2(U_t(\thetavar_{t, k}))\right\} B_{t, k} + (B_{t, k}^{\top})^{-1} \\
    &= \left\{\id - \eta h_t V_t\right\} B_{t, k} + h_t (B_{t, k}^{\top})^{-1}
\end{align*}
where we have used that $\nabla^2(U_t(\theta)) = \eta V_t$ for the linear bandit case (see \eqref{def:Vt}). Let denote by $\stdvar_{t, k} = B_{t, k} B_{t, k}^{\top}$ the covariance of the variational posterior $\varpost_{t, k}$. For ease of notation we denote by $A_t = \id - \eta h_t V_t$, it follows that
\begin{equation*}
    \stdvar_{t, k+1} = A_t \stdvar_{t, k} A_t + 2 h_t A_t + h_t^2 \precivar_{t, k}
\end{equation*}
If $\Lambda_{t, k} = \stdvar_{t, k} - 1/\eta \Vinv_t$ denotes the difference between the covariance matrix of the varational posterior and the true posterior, therefore it holds that
\begin{align*}
    \Lambda_{t, k+1} &= A_t \stdvar_{t, k} A_t + 2 h_t A_t + h_t^2 \precivar_{t, k} - 1/\eta \Vinv_t\\
     &=A_t \Lambda_{t, k} A_t + 2 h_t A_t - 2h_t \id +\eta h_t^2 V_t + h_t^2 \precivar_{t, k}\\
     &= A_t \Lambda_{t, k} A_t -\eta h_t^2 V_t + h_t^2 \precivar_{t, k} \\
     &= A_t \Lambda_{t, k} A_t - h_t^2 \eta V_t \Lambda_{t, k} \precivar_{t, k} \\
\end{align*}


\end{definition}
\begin{definition}\textbf{(Concentration events)}\label{def:events}

The main challenge for the proof of Theorem \ref{theo:main}, is to control the probability of the following events: 
for any $t \in [T]$ we define
\begin{itemize}
\item $\widehat{\Etrue_t} = \left\{\text{for any } a \in \A(x_t):\eqsp \vert\phi(x_t, a)^{\top}\meanpost_t - \phi(x_t, a)^{\top} \thetastar\vert \leq \eqsp g_1(t)\eqsp\eqsp \Vert\phi(x_t, a)\Vert_{\Vinv_t}\right\}$
\item $\Etrue_t = \widehat{\Etrue_t} \bigcap \Big\{\vert \xi_t \vert  <  R\sqrt{1+\log 3t^2}\Big\} \bigcap \Big\{\Vert \stdstdvar_{t, K_t}^{-1/2}(\meanvar_{t, K_t} - \meanmeanvar_{t, K_t})\Vert \leq \sqrt{4d\log3t^3}\Big\}$
\item $\Evar_t = \left\{\text{for any } a \in \A(x_t): \eqsp \vert\phi(x_t, a)^{\top}\thetavar_t - \phi(x_t, a)^{\top} \meanpost_t\vert \leq \eqsp g_2(t)\eqsp \eqsp \Vert\phi(x_t, a)\Vert_{\Vinv_t}\right\}$ ,
\end{itemize}
where $g_1(t) = R \sqrt{d \log(3t^3)} + \sqrt{\lambda}$ and $g_2(t) =10\sqrt{d\log(3t^3)/(\eta \lambda)} $ and $\xi_t$ is the R-sub Gaussian noise of the reward definition defined by the relation
\begin{equation}
r_t = \phi_t^{\top} \thetastar + \xi_t\eqsp.
\end{equation} 

The first event $\widehat{\Etrue_t}$ controls the concentration of $\phi(x_t, a)^{\top}\meanpost_t$ around its mean. Similarly, event $\Evar_t$ controls the concentration of $\phi(x_t, a)^{\top}\thetavar_t$ around its mean. Note that compared to \cite{agrawal2013thompson}, in our case, it is important to include within $\Etrue_t$, the concentration of the distributions $\xi_t$ and $\stdstdvar_{t, K_t}^{-1/2}(\meanvar_{t, K_t} - \meanmeanvar_{t, K_t})$. Consequently, conditionally on $\Etrue_t$ and $\Evar_t$ it holds that: for any  $a \in \A(x_t)$
\begin{align}\label{eq:eventconcentration}
\vert\phi(x_t, a)^{\top}\thetavar_t - \phi(x_t, a)^{\top} \thetastar\vert &\leq \Bigg(R \sqrt{ d \log(3t^3)} + \sqrt{\lambda} + 10\sqrt{d\log(3t^3)/(\eta \lambda)} \Bigg)\Vert\phi(x_t, a)\Vert_{\Vinv_t}\nonumber\\
&\leq 12R\sqrt{d\log(3t^3)/(\eta \lambda)}\eqsp \Vert\phi(x_t, a)\Vert_{\Vinv_t}\nonumber\\
&\stackrel{(a)}{=} \frac{108 d R^2}{\lambda^{3/2}} \sqrt{\log(3t^3) \log(3T^3)} \Vert\phi(x_t, a)\Vert_{\Vinv_t}\nonumber \\
&:= g(t) \Vert\phi(x_t, a)\Vert_{\Vinv_t}\eqsp,
\end{align}
where in $(a)$, we have used that $\eta = 4 \lambda / \Big(81 R^2 d \log(3T^3)\Big)$ and in the last inequality we have used that $g(t) = C R^2 d \sqrt{\log(t) \log(T)} / \lambda^{3/2}$.

\end{definition}

\subsection{Main lemmas}\label{sec:mainlemmas}
\begin{lemma}\textbf{(Concentration lemma for $\meanpost_t$)}\label{lemma:etrue}

Recall the definition of the event $\Etrue_t$ in \eqref{def:events}. Therefore, for any $t \in [T]$,  it holds that
\begin{equation}
    \prob(\Etrue_t) \geq 1 - \frac{1}{t^2}
\end{equation}
This lemma shows that the mean of the posterior distribution $\meanpost_t$ is concentrated around the true parameter $\thetastar$ with high probability.
\end{lemma}
\begin{proof}
Firstly, we apply Lemma \ref{lem:mart}, with $m_t = \phi_t / \sqrt{\lambda} = \phi(x_t, a_t) / \sqrt{\lambda}$ and $\epsilon_t = \big(r_{a_t}(t) - \phi_t^{\top} \thetastar\big) / \sqrt{\lambda}$, where $r_{a_t}(t)$ is sampled from the R-sub-Gaussian reward distribution of mean $\phi_t^{\top} \thetastar$. Let's define the filtration $\mathcal{F}_t' = \{a_{\tau+1}, m_{\tau+1}, \epsilon_{\tau}\}_{\tau \leq t}$. By the definition of $\filt_t'$, $m_t$ is $\filt_{t-1}'$-measurable. Moreover, $\epsilon_t$ is conditionally $R/\sqrt{\lambda}$-sub-Gaussian due to Assumption \ref{as:subgaussian_rew} and is a martingale difference process because $\E[\epsilon_t|\filt_{t-1}'] = 0$. If we denote by
\begin{equation*}
    M_t = \id_d + 1 / \lambda \sum_{\tau=1}^t m_{\tau} m_{\tau}^{\tau} = 1/ \lambda V_{t+1} \eqsp,
\end{equation*}
and
\begin{equation*}
    \zeta_t = \sum_{\tau=1}^t m_{\tau} \epsilon_{\tau}\eqsp,
\end{equation*}
Then, Lemma \ref{lem:mart} shows that $\Vert\zeta_t\Vert_{M_t^{-1}} \leq R / \sqrt{\lambda}\sqrt{d \log(\frac{t+1}{\delta'})}$ with probability at least $1-\delta'$.
Moreover, note that
\begin{align*}
    M_{t-1}^{-1} (\zeta_{t-1} - \thetastar) &= M_t^{-1} (1/\lambda b_t - 1/\lambda \sum_{\tau=1}^{t-1} \phi_{\tau} \phi_{\tau}^{\top} \thetastar - \thetastar)\\
    &= M_{t-1}^{-1} (1/\lambda b_t - M_{t-1} \thetastar)\\
    &= \meanpost_t - \thetastar \eqsp.
\end{align*}

Note that $\vert\vert \thetastar\vert\vert_{M_{t-1}^{-1}} = \normdeux{\thetastar M_{t-1}^{-1/2}} \leq \normdeux{\thetastar} \normdeux{M_{t-1}^{-1/2}}\leq \normdeux{\thetastar}$, where the last inequality is due to Assumption \ref{ass:thetastar}. Then, for any arm $a \in \A(x_t)$ we have
\begin{align*}
    |\phi(x_t, a)^{\top} \meanpost_t - \phi(x_t, a)^{\top} \thetastar| &= |\phi(x_t, a) M_{t-1}^{-1} (\xi_{t-1} - \thetastar)|\\
    &\leq \Vert\phi(x_t, a)\Vert_{M_{t-1}^{-1}} \Vert\xi_{t-1} - \thetastar\Vert_{M_{t-1}^{-1}}\\
    &\leq \Vert\phi(x_t, a)\Vert_{M_{t-1}^{-1}} (\Vert\xi_{t-1}\Vert_{M_{t-1}^{-1}} + \Vert\thetastar\Vert_{M_{t-1}^{-1}})\\
    &\leq \sqrt{\lambda} \left(R/\sqrt{\lambda} \sqrt{d \log(\frac{t}{\delta'})} +1\right) \Vert\phi(x_t, a)\Vert_{\Vinv_t}\\
    &= \sqrt{\lambda}\left(R/\sqrt{\lambda} \sqrt{d \log(3 t^3)} +1\right) \Vert\phi(x_t, a)\Vert_{\Vinv_t} \\
    &=  \left(R\sqrt{d \log(3 t^3)} + \sqrt{\lambda}\right)\Vert\phi(x_t, a)\Vert_{\Vinv_t}\\
    &:= g_1(t) \Vert\phi(x_t, a)\Vert_{\Vinv_t}\eqsp.
\end{align*}
This inequality holds with probability at least $\delta' = 1/(3 t^2)$.

Moreover, recall the definition of the R-subGaussian noise of the reward definition in section \ref{def:events}
\begin{equation*}
    r_t = \phi_t^{\top} \thetastar + \xi_t
\end{equation*}
Then it holds that $\prob(\vert \xi_t \vert > x) \leq \exp(1 - x^2 / R^2)$. It follows that $\prob(\vert \xi_t \vert \leq R \sqrt{1 + \log 3t^2}) \geq 1 - 1 / (3t^2)$, for any $t\leq 1$. Finally, recall the definition of $\stdstdvar_{t, k}$, $\meanvar_{t, k}$ and $\meanmeanvar_{t, k}$ in section \ref{def:paramalgo}. Consequently, the term $\stdstdvar_{t, K_t}^{-1/2}(\meanvar_{t, K_t} - \meanmeanvar_{t, K_t})$ is gaussian with mean $0$ and an identity covariance matrix. Therefore, it holds that
\begin{equation}
    \prob\Big(\Vert \stdstdvar_{t, K_t}^{-1/2}(\meanvar_{t, K_t} - \meanmeanvar_{t, K_t})\Vert \leq \sqrt{4d\log3t^3}\Big) \geq 1 - 1/(3t^2)
\end{equation}

Consequently, we have
\begin{align*}
    \prob\Big(\widehat{\Etrue_{t}} \bigcap \left\{\vert \xi_i \vert  <  R\sqrt{1+\log 3t^2}\right\} \bigcap \left\{\Vert \stdstdvar_{t, K_t}^{-1/2}(\meanvar_{t, K_t} - \meanmeanvar_{t, K_t})\Vert \leq \sqrt{4d\log3t^3}\right\} \Big) \geq 1 - \frac{1}{t^2}\eqsp,
\end{align*}
where $\widehat{\Etrue_{t}}$ is defined is $\ref{def:events}$

\end{proof}

\begin{lemma}\label{lemma:unsaturatedprob}\textbf{Probability of playing an unsaturated arm}

Given $\Etrue_t$ defined in section \eqref{def:events}, the conditional probability of playing an unsaturated arm is strictly positive and is lower bounded as 
\begin{equation}
   \1_{\Etrue_t} \prob_t(a_t \in \mathcal{U}_t) := \prob(a_t \in \mathcal{U}_t|\filt_{t-1}) \geq \1_{\Etrue_t}(p - 1/t^2)\eqsp,
\end{equation}
where $p =1/\sqrt{2\pi \rme}$ and $\mathcal{U}_t$ is defined in \eqref{def:saturated}.
\end{lemma}
\begin{proof}
If we suppose that $\forall a \in \mathcal{S}_t$, $\eqsp\phi(x_t, \as_t)^{\top} \thetavar_t \geq \phi(x_t, a)^{\top} \thetavar_t$, then $a_t \in \mathcal{U}_t$. Indeed, The optimal arm $\as_t$ is obviously in the unsaturated arm set ($\mathcal{U}_t$) and $\phi(x_t, a_t)^{\top} \thetavar_t \geq \phi(x_t, \as_t)^{\top} \thetavar_t$ by construction of the algorithm. Hence we have
    \begin{equation*}
        \prob(a_t \in \mathcal{U}_t|\filt_{t-1}) \geq \prob(\phistar^{\top} \thetavar_t \geq \phi(x_t, a)^{\top} \thetavar_t, \forall a \in \mathcal{S}_t|\filt_{t-1})
    \end{equation*}

Subsequently, given events $\Etrue_t$ and $\Evar_t$ we have
\begin{equation*}
    \left\{\phistar^{\top} \thetavar_t \geq \phi(x_t, a)^{\top} \thetavar_t, \forall a \in \mathcal{S}_t\right\} \supset \left\{\phistar^{\top} \thetavar_t \geq \phistar^{\top} \thetastar\right\}\eqsp.
\end{equation*}
Indeed, for any $a \in \mathcal{S}_t$,
\begin{align*}
\phi(x_t, a)^{\top} \thetavar_t &\stackrel{(a)}{\leq} \phi(x_t, a)^{\top} \thetastar + g(t) \Vert\phi(x_t, a)\Vert_{\stdpost_t}\\
&\stackrel{(b)}{\leq} \phistar^{\top} \thetastar\eqsp,
\end{align*}
where $(a)$ uses that $\Etrue_t$ and $\Evar_t$ hold. And in inequality $(b)$ we have used that $a \in \mathcal{S}_t$, ie, $\phistar_t^{\top} \thetastar - \phi(x_t, a)^{\top} \thetastar :=\Delta_t(a) > g(t) \Vert\phi(x_t, a)\Vert_{\stdpost_t}$. 

Consequently,
\begin{align*}
    \prob(\phistar^{\top} \thetavar_t \geq \phistar^{\top} \thetastar|\filt_{t-1}) &= \prob(\phistar^{\top} \thetavar_t \geq \phistar^{\top} \thetastar|\filt_{t-1}, \Evar_t)\prob(\Evar_t) + \prob(\phistar^{\top} \thetavar_t \geq \phistar^{\top} \thetastar|\filt_{t-1}, \overbar{\Evar_t})\prob(\overbar{\Evar_t})\\
    &\leq \prob(\phistar^{\top} \thetavar_t \geq \phi(x_t, a)^{\top} \thetavar_t, \forall a \in \mathcal{S}_t|\filt_{t-1}) + \prob(\overbar{\Evar_t})
\end{align*}
Therefore,
\begin{align*}
    \prob(a_t \in \mathcal{U}_t|\filt_{t-1}) &\geq \prob(\phistar_t^{\top} \thetavar_t \geq \phistar_t^{\star} \thetastar|\filt_{t-1}) - \prob(\overbar{\Evar_t})\\
    &\geq p - \frac{1}{t^2}\eqsp,
\end{align*}
where the last inequality is due to Lemma \ref{lemma:anticoncentration} and Lemma \ref{lemma:concentrationvar} with $p=1/(2\sqrt{2\pi e})$.
\end{proof}

\begin{lemma}\textbf{(Upper bound of $\sum_{t=1}^T \Vert\phi_t\Vert_{\stdpost_t}$)} \label{lemma:upperboundsum}
The following lemma we will be useful in the derivation of the regret bound later in the proof.
\begin{align*}
     \sum_{t=1}^T \Vert\phi_t\Vert_{\Vinv_{t}} \leq \sqrt{2 d T \log\Big(1 + \frac{T}{\lambda d}\Big)}
\end{align*}
\end{lemma}

\begin{proof}
Recall the relation between the 1-norm and 2-norm for a d-dimensional vector, ie, $\Vert \cdot \Vert_1 \leq \sqrt{d} \Vert \cdot \Vert_2 $. Hence, it follows that
    \begin{align*}
        \sum_{t=1}^T \Vert\phi_t\Vert_{\Vinv_t} &\leq \sqrt{T \sum_{t=1}^T \Vert \phi_t \Vert_{\Vinv_t}^2}
    \end{align*}
Firs, recall the definition of $V_t = \lambda \id \sum_{s=1}^{t-1} \phi_s \phi_s^{\top}$ in \eqref{def:Vt}.
Therefore, we apply Lemma 11 and Lemma 10 of \cite{abbasi:2011}, then we have
\begin{align*}
    \sum_{t=1}^T \Vert \phi_t \Vert_{\Vinv_t}^2 &\leq 2 \log \frac{\det V_t}{\det \lambda \id}\\
    &\leq 2 \log \frac{(\lambda + T/d)^d}{\lambda^d}\\
    &= 2 d \log\big( 1 + \frac{T}{\lambda d}\big) \eqsp.
\end{align*}
Consequently,
\begin{equation*}
    \sum_{t=1}^T \Vert\phi_t\Vert_{\Vinv_{t}} \leq \sqrt{2 d T \log\Big(1 + \frac{T}{\lambda d}\Big)}
\end{equation*}

\end{proof}
\subsection{Technical Lemmas}
\subsubsection{Upper bound of variational mean concentration term}\label{sec:meanmean}
In this section the objective is to bound the mean variational concentration term, ie, $\vert \phi^{\top}(\meanmeanvar_{t, k} - \meanpost) \vert$.

\begin{lemma}\label{lemma:meanmean}
Given $\Etrue_t$ defined in section \eqref{def:events}, the expected mean of the variational posterior at time step t after $K_t$ steps of gradient descent $\meanmeanvar_{t, K_t}$, defined in section \eqref{def:paramalgo}, is equal to:
\begin{equation}
    \tilde{m}_{t, K_t} = \sum_{j=1}^{t-1}\prod_{i=j}^{t-1} A_{i}^{K_i - 1}(\meanpost_j - \meanpost_{j+1}) + \meanpost_{t}
\end{equation}
where $A_i = \id - \eta h_i V_i$.
\end{lemma}
\begin{proof}
Recall the definitions of $\meanvar_{t, k}$ and $\meanmeanvar_{t, k}$ in section \ref{def:paramalgo}. Moreover, this section also presents the  sequence $\{\meanvar_{t, k }\}_{k=1}^{K_t}$ defined recursively in Algorithm \ref{alg:varts} by:
\begin{equation*}
    \meanvar_{t, k +1} =\meanvar_{t, k} - h_t \eta V_t(\thetavar_{t, k} - \meanpost_t)\eqsp.
\end{equation*}
Note that $\{\meanvar_{t, k}\}_{k=1}^{K_t}$ is a sequence of Gaussian samples with mean and covariance matrix $\meanmeanvar_{t, k}$ and $\stdstdvar_{t, k+1}$ respectively (see \eqref{def:paramalgo}). Then, we have,
\begin{align*}
\meanmeanvar_{t, k+1} &= \E[\meanvar_{t, k+1}]\\
    &= \meanmeanvar_{t, k} - \eta  h_t V_t(\meanmeanvar_{t, k} - \meanpost_t)\\
    &= (\id - h_t \eta V_t) \meanmeanvar_{t, k} + \eta h_t V_t \meanpost_t
\end{align*}
Now, we recognise an arithmetico-geometric sequence, therefore the solution is:
\begin{equation*}
    \meanmeanvar_{t, k} = (\id -  h_t \eta V_t)^{k-1}(\meanmeanvar_{t, 1} - \meanpost_t) + \meanpost_t
\end{equation*}
Moreover, in the algorithm we use that $\meanvar_{t, 1} = \meanvar_{t-1, K_{t-1}}$, which implies that $\meanmeanvar_{t, 1} = \meanmeanvar_{t-1, k_{t-1}}$ and $W_{t, 1} = W_{t-1, K_{t-1}}$. Hence, we have
\begin{align}
    \meanmeanvar_{t, K_t} &= \prod_{i=1}^{t}(\id - \eta h_i V_i)^{K_i - 1}(\meanmeanvar_{1, 1} - \meanpost_1) + \sum_{j=1}^{t-1}\prod_{i=j+1}^{t} (\id -  \eta h_i V_i)^{K_i -1}(\meanpost_j - \meanpost_{j+1}) + \meanpost_{t} \label{eq:meanmeanvar}
\end{align}
Moreover, the mean of the variational posterior is initialized at $\meanvar_{1, 1} = 0_d$, then the expected mean of the variational posterior $\meanmeanvar_1 = \meanpost_1 = 0_d$. Therefore the first term of \eqref{eq:meanmeanvar} is null. 
\end{proof}

\begin{lemma}\label{lemma:meanmeaninequality}
Given  $\Etrue_t$, for any $\phi \in \R^d$, it holds that
    \begin{equation*}
    \vert \phi (\meanmeanvar_{t, K_t} - \meanpost_t) \vert \leq  \sum_{j=1}^{t-1} \prod_{i=j}^{t-1} (1 - \eta h \lmin(V_i))^{K_i - 1} \Vert\phi\Vert_{\Vinv_j} \Vert\phi_j\Vert_{\Vinv_j} \left(g_1(t)/\sqrt{\lambda} + R\sqrt{1 + \log(3t^2)}\right)
\end{equation*}
where $\meanmeanvar_{t, K_t}$ is the expected mean of the variational posterior at time step t after $K_t$ steps of gradient descent, ie, $\meanmeanvar_{t, K_t} = \E[\meanvar_{t, K_t}]$, (see section \ref{def:paramalgo}). Recall that $g_1(t) = R \sqrt{ d \log(3t^3)} + \sqrt{\lambda}$ (see section: \ref{def:events}). 
\end{lemma}

\begin{proof}
Lemma \ref{lemma:meanmean} gives us that $\meanmeanvar_{t, K_t} = \sum_{j=1}^{t-1} \prod_{i=j}^{t-1} A_i^{K_i - 1} (\meanpost_j - \meanpost_{j+1}) + \meanpost_{t}$ where $A_i = \id -  \eta h_i V_i$. Then, for any $\phi \in \R^{d}$, the term we want to upper bound is:

\begin{align}
    \vert \phi^\top( \meanmeanvar_{t, K_t} - \meanpost_{t}) \vert &\leq  \sum_{j=1}^{t-1}   \vert \phi^\top \prod_{i=j}^{t-1} A_{i}^{K_i - 1}(\meanpost_j - \meanpost_{j+1})\vert\eqsp,\label{eq:phi_upp}
\end{align}

We can notice that the previous term only depends on the difference between the mean posterior at time j and the one at time j+1, which can be upper bounded. Recall the different relations between $V_j$, $b_j$, $r_j$, $\phi_j$ and $\stdpost_j$ in the linear bandit setting (see equation \eqref{def:Vt}):  $V_{j+1} = V_j +  \phi_j \phi_j^{\top}$, $b_{j+1} = b_j +  r_j \phi_j$ and $\meanpost_j = \Vinv_j b_j$, then by Sherman–Morrison formula we have:
\begin{equation}\label{ref:sm}
    \Vinv_{j+1} = (V_j +  \phi_j \phi_j^{\top})^{-1} = \Vinv_j -  \frac{\Vinv_j \phi_j \phi_j^{\top} \Vinv_j}{1 +  \phi_j^{\top} \Vinv_j \phi_j}
\end{equation}

The difference between the mean posterior at time $j+1$ and the one at time $j$ becomes:
\begin{align}\nonumber
\meanpost_{j+1}-\meanpost_j &= \Vinv_{j+1} b_{j+1}-\Vinv_j b_j \\
& = (\Vinv_j -   \frac{\Vinv_j \phi_j \phi_j^{\top} \Vinv_j}{1 +   \phi_j^{\top} \Vinv_j \phi_j})(b_j +   r_j \phi_j) - \Vinv_j b_j \nonumber\\
&=   r_j \Vinv_j \phi_j -   \frac{\Vinv_j \phi_j \phi_j^{\top} \Vinv_j}{1 +   \phi_j^{\top} \Vinv_j \phi_j}(b_j +   r_j \phi_j) \nonumber\\
&=   \frac{\Vinv_j \phi_j}{1 +   \phi_j^{\top} \Vinv_j \phi_j} \left\{ -\phi_j^{\top} \meanpost_j -   r_j \phi_j^{\top} \Vinv_j \phi_j + r_j  (1 +   \phi_j^{\top} \Vinv_j \phi_j^{\top})\right\} \nonumber\\
&=   \frac{\Vinv_j \phi_j (r_j - \phi_j^{\top} \meanpost_j)}{1 +    \phi_j^{\top} \Vinv_j \phi_j } \nonumber\\
&\stackrel{(a)}{=}   \frac{\Vinv_j \phi_j (\phi_j^{\top} \thetastar + \xi_j - \phi_j^{\top} \meanpost_j)}{1 +   \phi_j^{\top} \Vinv_j \phi_j} \nonumber\\
&\stackrel{(b)}{\leq}   \Vinv_j \phi_j (\phi_j^{\top} \thetastar + \xi_j - \phi_j^{\top} \meanpost_j) \label{eq:diffpost}
\end{align}
where in $(a)$ we have used that $r_j = \phi_j^{\top} \thetastar + \xi_j$ with $\xi_j$ is sampled from a R-Subgaussian distribution. Inequality $(b)$ is due to $\phi_j^{\top} \Vinv_j \phi_i = \Vert\phi_j\Vert_{\Vinv_j}^2 > 0$. 

Subsequently, combining equations \eqref{eq:phi_upp} and \eqref{eq:diffpost}, we obtain the following upper bound
\begin{align*}
    \vert \phi (\meanmeanvar_{t, K_t} - \meanpost_t) \vert &\leq \sum_{j=1}^{t-1} \prod_{i=j}^{t-1} \vert \phi^\top A_{i}^{K_i - 1}(\meanpost_j - \meanpost_{j+1})\vert\\
    &\stackrel{(a)}{\leq} \sum_{j=1}^{t-1} \prod_{i=j}^{t-1}  \left\vert \phi^\top A_{i}^{K_i - 1}  \Vinv_j \phi_j (\phi_j^{\top} \thetastar + \xi_j - \phi_j^{\top} \meanpost_j)\right\vert\\
    &\stackrel{(b)}{\leq}   \sum_{j=1}^{t-1} \prod_{i=j}^{t-1} (1 - \eta h \lmin(V_i) )^{K_i - 1} \phi^{\top} V^{-1/2}_j V^{-1/2}_j \phi_j |(\phi_j^{\top} \thetastar + \xi_j - \phi_j^{\top} \meanpost_j)| 
    \\
    &\stackrel{(c)}{\leq}   \sum_{j=1}^{t-1} \prod_{i=j}^{t-1} (1 -   \eta h \lmin(V_i))^{K_i - 1} \Vert\phi\Vert_{\Vinv_j} \Vert\phi_j\Vert_{\Vinv_j} |(\phi_j^{\top} \thetastar + \xi_j - \phi_j^{\top} \meanpost_j)|
    \\
    &\stackrel{(d)}{\leq}   \sum_{j=1}^{t-1} \prod_{i=j}^{t-1} (1 - \eta  h \lmin(V_i))^{K_i - 1} \Vert\phi\Vert_{\Vinv_j} \Vert\phi_j\Vert_{\Vinv_j} \left(g_1(t)/\sqrt{\lambda} + R\sqrt{1 + \log(3t^2)}\right)
\end{align*}
In the inequality $(a)$ we have used equation \eqref{eq:diffpost}, 
in $(b)$ the relation $A_i^{K_i-1} = (\id - \eta h_i V_i)^{K_i - 1} \preceq (1 - \eta h_i \lmin(V_i))^{K_i - 1}\eqsp \id_d$, 
in $(c)$ the definition of $\Vert\phi\Vert_{\Vinv_t} = \sqrt{\phi^{\top} \Vinv_t \phi} = \sqrt{\phi^{\top} V_t^{-1/2}} \sqrt{V_t^{-1/2} \phi} = \phi^{\top} V_t^{-1/2}$,
and finally $(d)$ is due to $\vert \xi_i \vert  <  R\sqrt{1+\log 3t^2}$ as $\Etrue_t$ holds and  $\vert \phi_i^{\top}  (\thetastar - \meanpost_t) \vert \leq g_1(t) \norm{\phi_i^\top}_{\Vinv_t}  \leq g_1(t) / \sqrt{\lambda}$
\end{proof}

\begin{lemma}
\label{lemma:mean_post_concentration}
Given $\Etrue_t$, for any $\phi \in \R^d$, if the number of gradient descent of Algorithm \ref{alg:update} is such that for $t\geq 2$
\begin{equation*}
    K_t \geq  1 + 2(1+2\kappa_t^2) \log\Big(4R \sqrt{dT\log (3T^3)}\Big)  \eqsp,
\end{equation*}
then it holds that
\begin{equation*}
    |\phi(\meanmeanvar_{t, K_t} - \meanpost_t)| \leq \frac{ 2 \Vert \phi \Vert_{\Vinv_t} }{\lambda}
\end{equation*}
This lemma provides the upper bound for variational mean concentration term.
\end{lemma}

\begin{proof}

Firstly, we can apply Lemma \ref{lemma:meanmeaninequality}, it gives us
\begin{align*}
 \vert \phi (\meanmeanvar_{t, K_t} - \meanpost_t) \vert
    &\leq  \sum_{j=1}^{t-1} \prod_{i=j}^{t-1} (1 -  \eta h_t \lmin(V_t))^{K_i - 1} \Vert\phi\Vert_{\Vinv_j} \Vert\phi_j\Vert_{\Vinv_j} \left(g_1(t)/\sqrt{\lambda} + R\sqrt{1 + \log(3t^2)}\right)
\end{align*}
where $g_1(t) = R \sqrt{ d \log(3t^3)} + \sqrt{\lambda}$. Moreover, for $t\geq 2,$ 
\begin{align}
R \sqrt{1+\log 3t^2} + g_1(t) / \sqrt{\lambda} &\leq R \sqrt{\log 3t^2}+R \sqrt{d \log \left(3t^3\right) / \lambda}+R+1 \\ &\leq 4R \sqrt{d\log 3t^2/\lambda}
\label{eq:R}
\end{align}
where we have used that $R\geq 1$ and $\lambda\leq 1$. Moreover, for any $j \in [1, t]$ we have 
\begin{align}\label{eq:norm}
    \Vert\phi \Vert_{\Vinv_j} &\leq \Vert\phi\Vert_2 / \sqrt{\lambda}\\
    &\leq   \lmax(V_t)^{1/2}\Vert\phi\Vert_{\Vinv_t} / \sqrt{\lambda} \\
    &= \lmax(V_t)^{1/2}\Vert\phi\Vert_{\Vinv_t}/\lambda^{1/2}
\end{align}

Let's define  $\epsilon = \left( 4R \sqrt{dt\log (3t^2)} \right)^{-1} \leq 1/2$ and let's take $K_i$ such that $(1 - h_t \lmin(V_t))^{K_i-1} \leq \epsilon$, this condition will be explained later in the proof. It follows that

\begin{align*}
    \vert \phi (\meanmeanvar_{t, K_t} - \meanpost_t) \vert &\leq  \sum_{j=1}^{t-1} \prod_{i=j}^{t-1} (1 - \eta h_t \lmin(V_t))^{K_i - 1} \Vert\phi\Vert_{\Vinv_j} \Vert\phi_j\Vert_{\Vinv_j} \epsilon^{-1}
    \\
    &\stackrel{(a)}{\leq} \frac{  \Vert \phi \Vert_{\Vinv_t} \lmax(V_t)^{1/2} }{\lambda}  \sum_{j=1}^{t-1} \prod_{i=j}^{t-1} (1 -  h_t \lmin(V_T))^{K_i-1} \epsilon^{-1}
    \\
    &\stackrel{(b)}{\leq} \frac{  \Vert \phi \Vert_{\Vinv_t} }{\lambda} \sum_{j=1}^{t-1} \epsilon^{t-j-1}
    \\
    &\stackrel{(c)}{\leq} \frac{  \Vert \phi \Vert_{\Vinv_t} }{\lambda}\times \frac{1}{1- \epsilon}
    \\
    &\stackrel{(d)}{\leq} \frac{ 2 \Vert \phi \Vert_{\Vinv_t} }{\lambda} \eqsp,
\end{align*}
where $(a)$ comes from equations \eqref{eq:norm} and \eqref{eq:R}. The point (b) comes that $\lmax(V_t)\leq  \sqrt{t}$ because $ \lambda\leq 1$ and definition of $\epsilon$,  then $(c)$ from the geometric series formula. Finally, in $(d)$, we have used $\epsilon \leq 1/2$.

Now, let's focus on that condition on $K_i$ presented previously. For any $i \in[t]$, recall the definition of the step size $h_t$ in \ref{def:paramalgo}.
\begin{equation*}
    h_i = \frac{ \lmin(V_i)}{2 \eta (\lmin(V_i)^2 + 2\lmax(V_i)^2)}  \eqsp, 
\end{equation*}
and define $\kappa_i = \lmax(V_i) / \lmin(V_i)$. Therefore, it holds that
\begin{equation*}
    \left(1 - \eta h_i\lmin(V_i)\right)^{K_i-1} = \left(1 - \frac{1 }{2 (1 + 2\kappa_i^2)}\right)^{K_i-1}
\end{equation*}

For any $\epsilon >0$, we want that $\left(1 - h_i\lmin(V_i)\right)^{K_i-1} \leq\epsilon$. Hence we deduce that

\begin{equation*}
    K_i \geq 1 + \frac{\log(1/\epsilon)}{\log(1-1/(2 (1+2\kappa_i^2)))} \eqsp.
\end{equation*}
Moreover, if $0 < x < 1$ then we have $-x > \log(1 - x)$, if follows that
\begin{equation*}
    K_i \geq 1 + 2 (1 + 2\kappa_i^2) \log(1/\epsilon) \eqsp.
\end{equation*}
We note that,
\begin{align*}
\log(1/\epsilon) &= \log\Big(4R \sqrt{dt\log 3t^3}\Big)
\\
&\leq  \log\Big(4R \sqrt{dT\log 3T^3}\Big)  \eqsp.
\end{align*}

Finally, taking $K_i \geq 1 + 2 (1+2\kappa_i^2) \log\Big(4R \sqrt{dT\log (3T^3)}\Big)$, we obtain the condition
\begin{equation*}
    (1 -  \eta h_i \lmin(V_t))^{K_i - 1} \leq \epsilon\eqsp,
\end{equation*}
which concludes the proof.
\end{proof}

\subsubsection{Control the variational covariance matrix}\label{sec:sigma}
The objective of this section is to control the following term: $\boldmath{\vert\phi^{\top}\left(\thetavar_{t, k}-\meanvar_{t, k}\right)\vert }$. As $\thetavar_{t, k}$ is a sample from a Gaussian distribution with mean $\meanvar_{t, k}$, the previous term will be controlled using Gaussian concentration and an upper bound of the norm of the variational covariance matrix $\stdvar_{t, k}$. Recall the definitions of parameters $\stdpost_t$, $B_{t, k}$, $\thetavar_{t, k}$ and $\meanvar_{t, k}$ in section \ref{def:paramalgo}.

\begin{lemma}\label{lemm:decroissance}
For any $t \in [T]$ and $k \in [K_t]$, the following relation holds:
\begin{equation}
   \Hyp:\eqsp\stdvar_{t, k} \succeq \Vinv_t / (2\eta)
\end{equation}

\end{lemma}

\begin{proof}
The sequence $\{\stdvar_{t, n}\}_{t \in [T], \eqsp n \in [k_t]}$ is initialized by $\stdvar_{1, 1} = \id /(\lambda\eta) = \Vinv_1 /\eta \succeq \Vinv_t / (2\eta)$. Hence, $\Hyp$ holds for the pair $t=1$ and $k=1$. 
Therefore, to conclude the proof, we have to show that the following transitions are true:

\begin{enumerate}[label=$\bullet$,wide, labelwidth=!, labelindent=0pt]
    \item for any $t \in [T]$, if $\Hyp$ holds at step $(t, K_t)$ then it stays true at step $(t+1, 1)$  (\texttt{recursion in }$\mathtt{t}$),\label{enum:1}
    \item for any $k \in [K_t]$, if $\Hyp$ holds at step $(t, k)$ then it stays true at step $(t, k+1)$ (\texttt{recursion in }$\mathtt{k}$).
\end{enumerate}

Firstly, let's focus on the first implication and suppose that $\Hyp$ holds at step $(t, K_t)$. Therefore we have
\begin{align*}
    \stdvar_{t+1, 1} \stackrel{(a)}{=} \stdvar_{t, K_t}
    \stackrel{(b)}{\succeq}  \Vinv_t/(2 \eta)
    \stackrel{(c)}{\succeq}  \Vinv_{t+1} / (2 \eta)
\end{align*}
where $(a)$ comes from the initialization of the sequence $\{\stdvar_{t, k}\}_{k \in [k_t]}$, $(b)$ from the hypothesis $\Hyp$ at step $(t, K_t)$. And finally, $(c)$ is due to $V_{t+1} = V_t + \phi_t \phi_t^{\top} \succeq V_t$. Then we can conclude that $\Hyp$ holds at step $(t+1, 1)$. 

Now we focus on the second implication and we suppose that $\Hyp$ holds at step $(t, k)$. For ease of notation we denote by $Z_{t, k} := \stdvar_{t, k} - \Vinv_t / (2 \eta)$. Therefore using the recursive definition of $\stdvar_{t, k}$ given in section \eqref{def:paramalgo}, we have
\begin{align*}
    Z_{t, k+1} &= A_t \stdvar_{t, k} A_t + 2 h_t A_t + h_t^2 \stdvar_{t, k}^{-1} - \Vinv_t/(2 \eta)\\
    &= A_t Z_{t, k} A_t + 2 h_t A_t + h_t^2 \stdvar_{t, k}^{-1} - h_t \id + \eta h_t^2 V_t / 2\\
&= A_t Z_{t, k} A_t + h_t \id - 3 h_t^2 \eta V_t / 4 + h_t^2 \stdvar_{t, k}^{-1}
\end{align*}
where in the last inequalities we have used that $A_t = (\id - \eta h_t^2 V_t)$. Moreover, all terms in the previous inequality are positive semi-definite. Indeed, as $\Hyp$ holds at step $(t, k)$, we know that $Z_{t, k} \succeq 0$ and then that $A_t Z_{t, k} A_t \succeq 0$. Moreover, $\stdvar_{t, k} \succeq \Vinv_t / (2 \eta) \succeq 0$, so $\stdvar_{t, k}^{-1} \succeq 0$. Finally, recall the definition of $h_t$ in section \eqref{def:paramalgo}
\begin{align*}
h_t &\leq \frac{\lmin(V_t)}{2 \eta (\lmin(V_t)^2 + 2\lmax(V_t)^2)} \\
&= \frac{1/\kappa_t}{2 \eta \lmax(V_t) ((1/\kappa_t)^2 + 1)}\\
&= \frac{4 }{3 \eta \lmax( V_t)} \times \frac{3/\kappa_t}{8 (1 + (1/\kappa_t^2))}\\
&\leq \frac{4 }{3 \eta \lmax( V_t)}\eqsp,
\end{align*}
where $\kappa_t = \lmax(V_t) / \lmin(V_t) \geq 1$. Consequently, the matrix $\id - 3 \eta h_t V_t / 4$ is also positive semi-definite. Subsequently, we have
\begin{equation*}
    Z_{t, k+1} \succeq 0\eqsp.
\end{equation*}
\end{proof}
\begin{lemma}
\label{lemma:Bbound1}
    For any $\phi \in \R^{d}$,  let $B_{t, k}$ the square root of the covariance matrix defined in Algorithm \eqref{alg:update}. It holds that
\begin{align*}
    &\Vert B_{t, K_t} \phi\Vert_2 \leq 1/\sqrt{\eta} \Big(1 + \sqrt{\normdeux{V_t} C_t}\Big)\Vert\phi\Vert_{\Vinv_t} \\
    & \Vert B_{t, K_t} \phi\Vert_2 \geq 1/\sqrt{\eta} \Big(1 - \sqrt{\normdeux{V_t} C_t}\Big)\Vert\phi\Vert_{\Vinv_t}
\end{align*}
where   $C_t=1/\lambda \sum_{j=1}^{t-1} \prod_{i=j+1}^{t} \left(1 - \frac{3h_t\eta }{2} \lmin(V_i)\right)^{K_i - 1}$ .
\end{lemma}
\begin{proof}
Recall the recursive relation of $\Lambda_{t, k}$ defined in Section \eqref{def:paramalgo}.
\begin{align*}
    \Lambda_{t, k+1} &= A_t \Lambda_{t, k} A_t - \eta h_t^2 V_t \Lambda_{t, k} \precivar_{t, k}\eqsp,
\end{align*}
Hence, we have the following relation on the norm of $\Lambda_{t, k+1}$:
\begin{align*}  
\normdeux{\Lambda_{t, k+1}} &\leq \normdeux{A_t}\normdeux{\Lambda_{t, k}} \normdeux{A_t} + \eta h_t^2 \normdeux{V_t}\normdeux{\Lambda_{t, k}} \normdeux{\precivar_{t, k}}\\
&= \left(\lmax(A_t)^2  + \eta h_t^2 \lmax(V_t) \lmax(\precivar_{t, k}) \right) \normdeux{\Lambda_{t, k}}\\
&\stackrel{(a)}{=} \left(1 - 2 \eta h_t \lmin(V_t) + \eta^2 h_t^2 \lmin(V_t)^2 + \eta h_t^2 \lmax(V_t) \lmax(\precivar_{t, k})\right) \normdeux{\Lambda_{t, k}}\\
&= \left(1 - \frac{3h_t \eta}{2} \lmin(V_t) + \eta h_t \{ h_t(\eta \lmin(V_t)^2 + \lmax(V_t) \lmax(\precivar_{t, k})) - \lmin(V_t) / 2\} \right) \normdeux{\Lambda_{t, k}}\\
&\stackrel{(b)}{\leq} \left(1 - \frac{3h_t\eta}{2} \lmin(V_t)\right) \normdeux{\Lambda_{t, k}}\eqsp,
\end{align*}

where $(a)$ uses that $\lmax(A_t) = 1 - \eta h\lmin(V_t)$. Finally, inequality $(b)$ is due to: $\stdvar_{t, k} \succeq \Vinv_t/(2 \eta)$ (Lemma \ref{lemm:decroissance}). Indeed it implies that
\begin{align*}
    h_t(\eta \lmin(V_t)^2 + \lmax(V_t) \lmax(\precivar_{t, k})) &\leq h_t(\eta \lmin(V_t)^2 + 2 \eta \lmax(V_t)^2)\\
    &\leq  \lmin(V_t) /2\eqsp,
\end{align*}
 where the inequality comes from the definition of the step size:  $h_t \leq \lmin(V_t)/\Big(2 \eta (\lmin(V_t)^2 + 2\lmax(V_t)^2)\Big)$. Subsequently, 
\begin{align}
    \normdeux{\Lambda_{t, K_t}} &\leq \left(1 - \frac{3h_t\eta }{2} \lmin(V_t)\right) \normdeux{\Lambda_{t, K_t - 1}}\nonumber\\
    &\leq \left(1 - \frac{3h_t\eta }{2} \lmin(V_t)\right)^{K_t - 1} \normdeux{\stdvar_{t, 1} - 1/\eta \Vinv_t}\nonumber
    \\
    &= \left(1 - \frac{3h_t\eta }{2} \lmin(V_t)\right)^{K_t - 1} \normdeux{\stdvar_{t-1, k_{t-1}} - 1/\eta \Vinv_t}\nonumber
    \\
    & \leq \prod_{i=1}^{t-1} \left(1 - \frac{3h_t\eta}{2} \lmin(V_i)\right)^{K_i - 1} \normdeux{\stdvar_{1, 1} - 1/\eta \Vinv_1} \label{eq:lambdabound}\\
    &+ \sum_{j=1}^{t-1} \prod_{i=j+1}^{t} \left(1 - \frac{3h_t\eta}{2} \lmin(V_i)\right)^{K_i - 1} \normdeux{1/\eta \Vinv_{j} - 1/\eta \Vinv_{j+1}}\nonumber
    \\
    & \stackrel{(a)}{\leq} \frac{1}{\eta} \sum_{j=1}^{t-1} \prod_{i=j+1}^{t} \left(1 - \frac{3h_t\eta}{2} \lmin(V_i)\right)^{K_i - 1} \normdeux{\Vinv_{j} - \Vinv_{j+1}}\nonumber
    \\
    &\stackrel{(b)}{\leq}  \frac{1}{\lambda \eta}  \sum_{j=1}^{t-1} \prod_{i=j+1}^{t} \left(1 - \frac{3h_t\eta}{2} \lmin(V_i)\right)^{K_i - 1}\eqsp,  \nonumber\\
    &:= C_t / \eta \nonumber
\end{align}
where in $(a)$ we have used that $\stdvar_{1, 1} = \frac{1}{\lambda\eta} \id = 1/\eta \Vinv_1$. Moreover $\normdeux{\Vinv_{j} - \Vinv_{j+1}} = \normdeux{\big(\Vinv_{j} \phi_{j} \phi_{j}^{\top} \Vinv_{j}\big) / \big(1 - \phi_{j}^{\top} \Vinv_{j} \phi_{j}\big)}$ see result \eqref{ref:sm}. It implies that  $\normdeux{\Vinv_{j} - \Vinv_{j+1}} \leq \normdeux{\Vinv_{j}}^2 \leq \normdeux{\Vinv_{1}}^2 = 1/\lambda$.

Finally, for any $\phi \in \R^d$, 
\begin{align*}
    \normdeux{B_{t, K_t} \phi} &= \sqrt{\phi^{\top} B_{t, K_t}^{\top} B_{t, K_t} \phi}\\
    &= \sqrt{\phi^{\top} \stdvar_{t, K_t} \phi}\\
    &= \sqrt{\phi^{\top} (\stdvar_{t, K_t} - 1/\eta \Vinv_t) \phi + 1/\eta \phi^{\top}\Vinv_t \phi}\\
    &\leq \normdeux{\phi} \sqrt{\normdeux{\stdvar_{t, K_t} - 1/\eta \Vinv_t}} + 1/\sqrt{\eta} \Vert\phi\Vert_{\Vinv_t}
\end{align*}
where the last inequality comes from the fact that for $a,b>0, \sqrt{a+b}<\sqrt{a}+ \sqrt{b}$, Moreover, 
\begin{align*}
    \normdeux{\phi} &= \normdeux{\phi V^{-1/2}_t V^{1/2}_t}\\
    &\leq \Vert\phi\Vert_{\Vinv_t} \normdeux{V_t^{1/2}}
\end{align*} 
Consequently, we have

\begin{align*}
    &\Vert B_{t, K_t} \phi\Vert_2 \leq 1/\sqrt{\eta} \Big(1 + \sqrt{\normdeux{V_t} C_t}\Big)\Vert\phi\Vert_{\Vinv_t}\\
\end{align*}
The lower bound of this lemma 

\begin{align*}
    \Vert B_{t, K_t} \phi\Vert_2 \geq 1/\sqrt{\eta} \Big(1 - \sqrt{\normdeux{V_t} C_t}\Big)\Vert\phi\Vert_{\Vinv_t}
\end{align*}
is obtained  because 

\begin{align*}
    \normdeux{B_{t, K_t} \phi} &= \sqrt{\phi^{\top} B_{t, K_t}^{\top} B_{t, K_t} \phi}\\
    &= \sqrt{\phi^{\top} \stdvar_{t, K_t} \phi}\\
    &= \sqrt{\phi^{\top} (\stdvar_{t, K_t} - 1/\eta \Vinv_t) \phi + 1/\eta \phi^{\top}\Vinv_t \phi}\\
    &\geq- \normdeux{\phi} \sqrt{\normdeux{\stdvar_{t, K_t} - 1/\eta \Vinv_t}} + 1/\sqrt{\eta} \Vert\phi\Vert_{\Vinv_t}
    &\leq  \Vert B_{t, K_t} \phi\Vert_2 \geq 1/\sqrt{\eta} \Big(1 - \sqrt{\normdeux{V_t} C_t}\Big)\Vert\phi\Vert_{\Vinv_t},
\end{align*}

where the first inequality comes from remarkable identity
$ \sqrt{a}-\sqrt{b}<\sqrt{a+b}$ for $a,b>0$.

\end{proof}

\begin{lemma}\label{lemma:Bbound}
For any $t \in [T]$ and $ a \in \A(x_t)$, if the number of gradient descent steps of Algorithm \ref{alg:update} is  $K_t \geq 1 + 4 (1 + 2\kappa_t^2) \log(2T) /(3\eta)$, therefore it holds that 
\begin{align*}
    \normdeux{\phi(x_t, a)^{\top}B_{t, K_t}} &\leq 1/\sqrt{\eta}\Big(1 + 1/\sqrt{\lambda} \Big)\Vert\phi\Vert_{\Vinv_t}\\
    \normdeux{\phi(x_t, a)^{\top}B_{t, K_t}} &\geq 1/\sqrt{\eta} \Big(1 - 1/\sqrt{\lambda}\Big)\Vert\phi\Vert_{\Vinv_t}\eqsp.
\end{align*}
 
\end{lemma}
\begin{proof}
Firstly, Lemma \ref{lemma:Bbound1}, gives us
\begin{align*}
    &\Vert B_{t, K_t} \phi\Vert_2 \leq 1/\sqrt{\eta} \Big(1 + \sqrt{\normdeux{V_t} C_t}\Big)\Vert\phi\Vert_{\Vinv_t}\\
    &\Vert B_{t, K_t} \phi\Vert_2 \geq 1/\sqrt{\eta} \Big(1 - \sqrt{\normdeux{V_t} C_t}\Big)\Vert\phi\Vert_{\Vinv_t}
\end{align*}
with $C_t = 1/\lambda \sum_{j=1}^{t-1} \prod_{i=j+1}^{t} (1 - \frac{3 h_t\eta}{2} \lmin(V_t))^{K_i-1}$. Furthermore, for any $t \in[T]$, recall that
\begin{equation*}
    h_t = \frac{\lmin(V_t)}{2 \eta ( \lmin(V_t)^2 + 2\lmax(V_t)^2)}\eqsp, 
\end{equation*}
and define $\kappa_t = \lmax(V_t) / \lmin(V_t)$. Therefore, it holds that

\begin{equation*}
    \left(1 - \frac{3 \eta h_t}{2} \lmin(V_t)\right)^{K_t-1} = \left(1 - \frac{3  }{4  (1 + 2\kappa_t^2)}\right)^{K_t-1}
\end{equation*}

For any $\epsilon >0$, we want that $\left(1 - \frac{3 h_t\eta }{2} \lmin(V_t)\right)^{K_t-1} \leq\epsilon$. Hence we deduce the following relation for $K_t$:

\begin{equation*}
    K_t \geq 1 + \frac{\log(\epsilon)}{\log(1-3\eta /(4 (1+2\kappa_t^2)))} \eqsp.
\end{equation*}

Moreover, if $0 < x < 1$ then we have $-x > \log(1 - x)$, then we have
\begin{equation*}
    K_t \geq 1 + 4 (1 + 2\kappa_t^2) \log(1/\epsilon) /3\eqsp.
\end{equation*}

Subsequently, let's apply the last result to $\epsilon = 1/(2 t)$. Then for $K_t \geq 1 + 4 (1 + 2\kappa_t^2) \log(2T) /3$, we have
\begin{align*}
   \normdeux{V_t} /\lambda \sum_{j=1}^{t-1} \prod_{i=j+1}^{t} \left(1 - \frac{3 \eta h_t}{2} \lmin(V_t)\right)^{K_i-1} &\leq \normdeux{V_t} \epsilon /\lambda\sum_{j=1}^{t-1}\epsilon^{t-j-1}
    \\
&\leq \normdeux{V_t} \epsilon /\lambda \sum_{j=0}^{+\infty}\epsilon^{j}
\\
& \stackrel{(a)}{\leq}\frac{\normdeux{V_t} }{2\lambda t (1 - \epsilon)}
\\
&\stackrel{(b)}{\leq} 1/\lambda \eqsp,
\end{align*}
where $(a)$ and $(b)$ come from the geometric serie because $\epsilon \leq 1/ 2$ and we have used that $\normdeux{V_t} = \normdeux{\lambda \id + \sum_{s=1}^{t-1} \phi \phi^{\top}} \leq \lambda + t-1 \leq t$, as $\lambda \leq 1$.
Consequently, we have
\begin{align*}
    &\Vert B_{t, K_t} \phi\Vert_2 \leq 1/\sqrt{\eta} \Big(1 + 1/\sqrt{\lambda} \Big)\Vert\phi\Vert_{\Vinv_t}\\
    &\Vert B_{t, K_t} \phi\Vert_2 \geq 1/\sqrt{\eta} \Big(1 - 1/\sqrt{\lambda}\Big)\Vert\phi\Vert_{\Vinv_t}
\end{align*}
\end{proof}

\begin{lemma}
\label{lemma:concentrationsigma}
For any $t \in [T]$ and $ a \in \A(x_t)$, if the number of gradient descent steps of Algorithm \ref{alg:update} is  $K_t \geq 1 + 4 (1 + 2\kappa_t^2) \log(2T) /3$ , then with probability at least $1 - 1/t^2$, we have

\begin{equation*}
    \vert \phi(x_t, a)^{\top}\left(\thetavar_{t, K_t}-\meanvar_{t, K_t}\right) \vert \leq \sqrt{4d\log(t^3)/\eta}\Big(1 + 1/\sqrt{\lambda} \Big)\Vert\phi(x_t, a)\Vert_{\Vinv_t}
\end{equation*}
\end{lemma}
\begin{proof}
For any $a \in \A(x_t)$, if $K_t \geq 1 + 4 (1 + 2\kappa_t^2) \log(2T) /(3\eta)$, Lemma \ref{lemma:Bbound} gives us that
\begin{align*}
    \vert \phi(x_t, a)^{\top}\left(\thetavar_{t, K_t}-\meanvar_{t, K_t}
\right) \vert &\leq\normdeux{B_{t, K_t}^{-1}( \thetavar_{t, K_t}-\meanvar_{t, K_t})}  \normdeux{\phi(x_t, a)^{\top}B_{t, K_t}}
\\
&\leq  \normdeux{B_{t, K_t}^{-1} (\thetavar_{t, K_t}-\meanvar_{t, K_t})}(1/\sqrt{\eta}) \Big(1 + 1/\sqrt{\lambda}\Big)\Vert\phi\Vert_{\Vinv_t}\eqsp.
\end{align*}
where first inequality comes from classical matrix norm inequality and the second one is previous lemma \ref{lemma:Bbound}, recall that $\thetavar_{t, K_t} \sim \N(\meanvar_{t, K_t}, B_{t, K_t} B_{t, K_t}^{\top})$, hence $B_{t, K_t}^{-1}(\thetavar_{t, K_t} - \meanvar_{t, K_t}) \sim \N(0, \id_d)$. Therefore, with probability $1 - 1/t^2$ we have
\begin{equation*}
    B_{t, K_t}^{-1}(\thetavar_{t, K_t} - \meanvar_{t, K_t}) \leq \sqrt{4 d \log(t^3)}\eqsp.
\end{equation*} 
Finally, we conclude that with probability $1 - 1/t^2$, it holds that
\begin{equation*}
    \vert\phi(x_t, a)^{\top} \left(\thetavar_{t, K_t} - \meanvar_{t, K_t}\right)\vert \leq \sqrt{4d\log(t^3)/\eta} \left(1 + 1/\sqrt{\lambda}\right) \Vert\phi(x_t, a)\Vert_{\Vinv_t}\eqsp.
\end{equation*}
\end{proof}

\subsubsection{Concentration of the mean of the Variational posterior around its mean \label{sec:W concentration} }

In this section, the objective is the show to concentration of $\meanvar_{t, k}$ around its mean $\meanmeanvar_{t, k}$ More precisely, we want an upper bound of $\boldmath{\vert \phi^{\top}\left(\meanvar_{t, k}-\meanmeanvar_{t, k}\right)\vert}$.

\begin{lemma}\label{lemma:wform}
For any $t \in [T]$ and $k\in [K_t]$, we have the following relation 
    \begin{align*}
    \stdstdvar_{t,k+1} &= (\id - \eta h_t V_t)\stdstdvar_{t,k}(\id - \eta h_t V_t)^T + \eta^2 h_t^2 V_t\E[\stdvar_{t,k}]V_t 
\end{align*}
where the sequence $\{\stdstdvar_{t, k}\}_{k=1}^{K_t}$ is introduced in section \ref{def:paramalgo}. (Recall : $\meanvar_{t, k} \sim \N(\meanmeanvar_{t,k}, \stdstdvar_{t, k})$ )
\end{lemma}
\begin{proof}
 In this section, we focus on the covariance matrix $\stdstdvar_{t, k}$ (see definition \ref{def:paramalgo}), by definition we have
\begin{align*}
    \stdstdvar_{t,k+1} &= \E[(\meanvar_{t,k+1} - \meanmeanvar_{t,k+1})(\meanvar_{t,k+1} - \meanmeanvar_{t,k+1})^{\top}]\\
    &=\E[a_{t, k+1}a_{t, k+1}^{\top}]\eqsp,
\end{align*}
where $a_{t, k}$ is the difference between $\meanvar_{t, k}$ and its mean. For ease of notation, let's define $\Omega_{t, k} := \thetavar_{t, k} - \meanmeanvar_{t,k}$, then we have
\begin{align*}
    a_{t, k+1} &= \meanvar_{t,k} - \meanmeanvar_{t,k} - \eta h_t V_t(\thetavar_{t, k} - \meanmeanvar_{t,k}) = \meanvar_k - \meanmeanvar_{t,k} - \eta h_t V_t\Omega_{t,k}.\\
\end{align*}
Consequently,
\begin{align*}
    a_{t, k+1} a_{t, k+1}^{\top} &= (\meanvar_{t,k} - \meanmeanvar_{t,k})(\meanvar_{t,k} - \meanmeanvar_{t,k})^{\top} - \eta h_t V_t\Omega_{t,k}(\meanvar_{t,k} - \meanmeanvar_{t,k})^{\top}\\
    &\quad- \eta h_t(\meanvar_{t,k}- \meanmeanvar_{t,k})\Omega_{t,k}^{\top}V_t+ \eta^2 h_t^2 V_t\Omega_{t,k} \Omega_{t,k}^{\top} V_t
\end{align*}
Moreover, note that  $\thetavar_{t,k} = \meanvar_{t,k} + B_{t, k} \epsilon_{t,k}$ where $\epsilon_{t,k} \sim \N(0, \id)$. Subsequently we have $\Omega_{t,k} = \meanvar_{t,k} - \meanmeanvar_{t,k} +  \stdvar_{t,k}^{1/2} \epsilon_{t,k}$. Then we have $\E[\Omega_{t,k} \Omega_{t,k}^{\top}] = \stdstdvar_{t, k} + \E[B_{t, k} B_{t, k}^{\top}]$, $\E[\Omega_{t,k} (\meanvar_{t,k} - \tilde{m}_{t,k})^{\top}] = W_{t,k}$, and $\E[(\meanvar_{t,k} - \tilde{m}_{t,k})\Lambda_{t,k}^{\top}] = \stdstdvar_{t,k}$.
Finally we obtain that
\begin{align*}
    \stdstdvar_{t,k+1} &=\E[a_{t, k+1}a_{t, k+1}^{\top}]\\
    &= \stdstdvar_{t,k} - \eta h_t V_t \stdstdvar_{t,k} - \eta h_t \stdstdvar_{t,k}V_t + \eta^2 h_t^2V_t \stdstdvar_{t,k} V_t + \eta^2 h_t^2 V_t\E[B_{t, k}B_{t, k}^{\top}]V_t\\
    &= (\id - \eta h_t V_t)\stdstdvar_{t,k}(\id - \eta h_t V_t)^{\top} + \eta^2 h_t^2 V_t\E[B_{t, k}B_{t, k}^{\top}]V_t \eqsp.
\end{align*}
\end{proof}

\begin{lemma}\label{lemma:wbound}
Recall that $\meanvar_{t, K_t}$, the mean of the variational posterior after $K_t$ steps of gradient descent, is a sample from the Gaussian with mean $\meanmeanvar_{t, K_t}$ and covariance matrix $\stdstdvar_{t, K_t}$, ie, $\meanvar_{t, K_t} \sim \N(\meanmeanvar_{t, K_t}, \stdstdvar_{t, K_t})$. 
Recall the definition of $\Lambda_{t, k} = \stdvar_{t, k} - 1/\eta \Vinv_t$, and let denote by $\Gamma_{t, k} = \stdstdvar_{t, k} - J_t \Vinv_t$, where $J_t = h_t (2 \id - \eta h_t V_t)^{-1} V_t$.

This Lemma shows that the 2-norm of $\Gamma_{t, K_t}$  is controlled by
\begin{equation*}
    \normdeux{\Gamma_{t, K_t}} \leq\sum_{j=1}^{t} \frac{6 \kappa_j h_j \normdeux{V_j}}{\lambda} \sum_{r=1}^{j} \prod_{i=r}^{t} D_i^{K_i - 1}\eqsp,
\end{equation*}
where $\kappa_j = \lmax(V_j) / \lmin(V_j)$ and $D_i = \id - 3 \eta h_i \lmin(V_i) / 2$.
\end{lemma}
\begin{proof}
Lemma \eqref{lemma:wform} gives us that
 \begin{align*}
    \stdstdvar_{t,k+1} &= A_t \stdstdvar_{t,k} A_t  + \eta^2 h_t^2 V_t\stdvar_{t,k}V_t\eqsp,
\end{align*}
where $A_t = \id - \eta h_t V_t$.

Note that $J_t$ and $V_t$ commute, therefore we have
\begin{align*}
    A_t J_t \Vinv_t A_t = J_t \Vinv_t - 2 h_t \eta J_t + \eta^2 h_t^2 J_t V_t\eqsp.
\end{align*}
Consequently, by combining the two previous equations we obtain 
\begin{align*}
    \Gamma_{t, k+1} &= A_t \Gamma_{t, k} A_t - 2 h_t \eta J_t + \eta^2 h_t^2 J_t V_t + \eta h_t^2 V_t + \eta^2 h_t^2 V_t \Lambda_{t, k} V_t\\
    &= A_t \Gamma_{t, k} A_t + \eta^2 h_t^2 V_t \Lambda_{t, k} V_t - h_t \eta J_t (2 \id - \eta h_t V_t) + \eta h_t^2 V_t\\
    &= A_t \Gamma_{t, k} A_t + \eta^2 h_t^2 V_t \Lambda_{t, k} V_t\eqsp.
\end{align*}
It follows that
\begin{align*}
    \normdeux{\Gamma_{t, k+1}} \leq \normdeux{A_t}^2 \normdeux{\Gamma_{t, k}} + \eta^2 h_t^2 \normdeux{V_t}^2 \normdeux{\Lambda_{t, k}}
\end{align*}

Therefore, iterating over $k$ gives us
\begin{align*}
    \normdeux{\Gamma_{t, k+1}} &\leq \normdeux{A_t}^{2k} \normdeux{\Gamma_{t, k}} + \eta^2 h_t^2 \sum_{j=0}^{k-1} \normdeux{A_t}^{2j} \normdeux{V_t}^2 \normdeux{\Lambda_{t, k-j}}\eqsp.
\end{align*}

Moreover, Equation \eqref{eq:lambdabound} is used to controls the following quantity
\begin{equation*}
    \normdeux{\Lambda_{t, k}} \leq \left(1 - \frac{3 \eta h_t}{2} \lmin(V_t)\right)^{k - 1} \normdeux{\Lambda_{t, 1}}\eqsp.
\end{equation*}
Let's denote by $D_t = 1 - 3 \eta h_t\lmin(V_t) / 2$, Subsequently
\begin{align*}
    \normdeux{\Gamma_{t, k+1}} \leq \normdeux{A_t}^{2k} \normdeux{\Gamma_{t, k}} + \eta^2 h_t^2 \sum_{j=0}^{k-1} \normdeux{A_t}^{2j} \normdeux{V_t}^2 D_t^{k-j-1} \normdeux{\Lambda_{t, 1}}\eqsp.
\end{align*}
However, $\normdeux{A_t}^{2} = (1 - \eta h_t \lmin(V_t))^2 < (1-\frac{3 \eta h_t}{2} \lmin(V_t))$, because $\eta h_t \leq 1 / (4  \lmin(V_t))$. Consequently, the geometric sum has a common ratio strictly lower than 1, then it is upper bounded by:
\begin{align}
    \sum_{j=0}^{k-1} \left(\frac{\normdeux{A_t}^{2}}{(1-\frac{3\eta h_t}{2}\lmin(V_t))}\right)^j & \leq \sum_{j=0}^{+\infty} \left(\frac{\normdeux{A_t}^{2}}{(1-\frac{3\eta h_t}{2}\lmin(V_t))}\right)^j\nonumber\\
    &= \frac{1-\frac{3\eta h_t}{2}\lmin(V_t)}{1-\frac{3\eta h_t}{2}\lmin(V_t) - \normdeux{A_t}^{2}}\nonumber\\
    &\leq \frac{1-\frac{3\eta h_t}{2}\lmin(V_t)}{1/2\eqsp \eta h_t \lmin(V_t) - \eta^2 h_t^2 \lmin(V_t)^2}\nonumber\\
    &\leq \frac{6}{\eta h_t \lmin(V_t)}\label{eq:geosum}\eqsp,
\end{align}
where in the first inequality we have used that the ratio of the previous sum is positive. In the last inequality we have used that $\eta h_t \leq 1 / (6  \lmin(V_t))$ in the denominator and we can remove the negative part of the numerator. Therefore, it holds that
\begin{align*}
    \normdeux{\Gamma_{t, k+1}} \leq \normdeux{A_t}^{2k} \normdeux{\Gamma_{t, k}} + 6\eta \kappa_t h_t D_t^{k-1} \normdeux{V_t} \normdeux{\Lambda_{t,1}} \eqsp,
\end{align*}
where the last inequality comes from \eqref{eq:geosum} and the definition of $\kappa_t = \lmax(V_t)/\lmin(V_t)$. Finally, iterating over $t$ yields to:
\begin{align*}
    \normdeux{\Gamma_{t, k+1}} &\leq \normdeux{A_t}^{2k} \prod_{j=1}^{t-1} \normdeux{A_j}^{2(K_j-1)} \normdeux{\Gamma_{1, 1}} + \sum_{j=1}^{t-1} \normdeux{A_t}^{2k}\prod_{i=j+1}^{t-1} \normdeux{A_i}^{2(K_i-1)}  \Big(6 \eta \kappa_j h_j D_j^{K_j-1} \normdeux{V_j} \normdeux{\Lambda_{t, 1}}\Big)\\
    & \quad+ 6 \eta \kappa_t h_t D_t^{k-1} \normdeux{V_t} \normdeux{\Lambda_{t,1}}\\
    &\leq \sum_{j=1}^{t-1} \normdeux{A_t}^{2k}\prod_{i=j+1}^{t-1} \normdeux{A_i}^{2(K_i-1)}  \Big(6 \eta \kappa_j h_j D_j^{K_j-1} \normdeux{V_j} \normdeux{\Lambda_{t, 1}}\Big) + 6 \eta \kappa_t h_t D_t^{k-1} \normdeux{V_t} \normdeux{\Lambda_{t,1}}\eqsp,
\end{align*}
where in the last inequalities we have used that $W_{1, 1}$ is initialized such that $W_{1, 1} = 1/(11\eta \lambda) \id$ and that $J_1 V_1 = h_1 (2 \id - \eta h_1 \lambda \id)^{-1}  = 1/(11\eta \lambda) \id$ because $h_1 = 1 /(6 \eta \lambda)$.
Finally, we can conclude
\begin{align*}
    \normdeux{\Gamma_{t, K_t}} &\leq \sum_{j=1}^{t-1} \normdeux{A_t}^{2 (K_t-1)}\prod_{i=j+1}^{t-1} \normdeux{A_i}^{2(K_i-1)}  \Big(6 \eta \kappa_j h_j D_j^{K_j-1} \normdeux{V_j} \normdeux{\Lambda_{t, 1}}\Big) + 6 \eta \kappa_t h_t D_t^{K_t-1} \normdeux{V_t} \normdeux{\Lambda_{t,1}}\\
    &= \sum_{j=1}^{t} \prod_{i=j+1}^{t} \normdeux{A_i}^{2(K_i-1)}  \Big(6 \eta \kappa_j h_j D_j^{K_j-1} \normdeux{V_j} \normdeux{\Lambda_{t, 1}}\Big)\\
    &\leq  \sum_{j=1}^{t} \prod_{i=j}^{t} D_i^{K_i-1}  \Big(6 \eta \kappa_j h_j \normdeux{V_j} \normdeux{\Lambda_{t, 1}}\Big)\eqsp,
\end{align*}
where in the last inequality we have used that $\normdeux{A_t}^2 \leq D_t$. Moreover, equation \eqref{eq:lambdabound} gives us that $\normdeux{\Lambda_{j, 1}} \leq 1/(\eta\lambda) \sum_{r=1}^{j} \prod_{l=r}^{j-1} D_l^{K_l - 1}$. Consequently, it holds that
\begin{align*}
    \normdeux{\Gamma_{t, K_t}} &\leq  \sum_{j=1}^{t} \frac{6 \kappa_j h_j \normdeux{V_j}}{\lambda} \sum_{r=1}^{j} \prod_{l=r}^{j-1} D_l^{K_l - 1} \prod_{i=j}^{t} D_i^{K_i-1}\\
    &= \sum_{j=1}^{t} \frac{6 \kappa_j h_j \normdeux{V_j}}{\lambda} \sum_{r=1}^{j} \prod_{i=r}^{t} D_i^{K_i - 1}\eqsp.
\end{align*}
    
\end{proof}

\begin{lemma}
    \label{lemma:W concentration}
    For any $t\geq 2$, given $\Etrue_t$, if the number of gradient descent steps is $K_t \geq 1 + 4 (1 + 2\kappa_t^2) \log\Big(2\kappa_t d^2 T\log^2( 3T^3)\Big) /3$, therefore it holds that
    \begin{equation*}
        \vert \phi^{\top}\left(\meanvar_{t, K_t}-\meanmeanvar_{t, K_t}\right)\vert\leq \Bigg(\sqrt{\frac{3}{\eta \lambda d \log(3t^3)}} + \sqrt{4 d \log(3t^3) / (11\eta)} \Bigg)\Vert \phi \Vert_{\Vinv_t}\eqsp.
    \end{equation*}
\end{lemma}

\begin{proof}

For any $\phi \in \R^d$, 
\begin{equation}\label{eq:wdiff}
    \vert \phi^{\top}\left(\meanvar_{t, K_t}-\meanmeanvar_{t, K_t}\right)\vert \leq \normdeux{ \phi^\top   \stdstdvar_{t, K_t}^{1/2}  } \normdeux{  \stdstdvar_{t, K_t}^{-1/2} (\meanvar_{t, K_t}-\meanmeanvar_{t, K_t}))} \eqsp,
\end{equation}
where $ \stdstdvar_{t, K_t}^{1/2}$ is the unique symmetric square root of $ \stdstdvar_{t, K_t}$. Firstly, given $\Etrue_t$, the term $\normdeux{\stdstdvar_{t, K_t}^{-1/2} (\meanvar_{t, K_t}-\meanmeanvar_{t, K_t})}< \sqrt{4d\log(3t^3)} $.

Then, we observe that
\begin{align}
   \sqrt{4d\log(3t^3)} \normdeux{\stdstdvar_{t, K_t}^{1/2} \phi}&= \sqrt{4 d \log(3t^3) \phi \stdstdvar_{t, K_t} \phi^{\top}}\nonumber\\
   &\leq \sqrt{4 d \log(3t^3) \phi \Gamma_{t, K_t} \phi^{\top}} + \sqrt{4 d \log(3t^3) \phi J_t\Vinv_t \phi^{\top}}\label{eq:boundforw}\eqsp,
\end{align}
where $J_t = h_t (2 \id - \eta h_t V_t)^{-1} V_t = (2 \Vinv_t / h_t - \eta \id)^{-1}$ and $\Gamma_{t, k} = \stdstdvar_{t, k} - J_t \Vinv_t$.

Moreover, 
\begin{align*}
    \sqrt{\phi J_t \Vinv_t \phi^{\top}} &= \normdeux{(J_t \Vinv_t)^{1/2} \phi}\\
    &\stackrel{(a)}{=} \normdeux{J_t^{1/2} V_t^{-1/2} \phi}\\
    &\leq \normdeux{J_t^{1/2}} \Vert \phi \Vert_{\Vinv_t}\eqsp,
\end{align*}
where in inequality $(a)$ we have used that $J_t$ and $\Vinv_t$ commute.

Recall that $V_t$ is a symmetric matrix, therefore we have $\lmin(V_t) \id \preceq V_t \preceq \lmax(V_t) \id$. It follows that
\begin{equation*}
   \frac{2}{h_t \lmax(V_t)} \id \preceq \frac{2}{h_t} \Vinv_t \preceq \frac{2}{h_t \lmin(V_t)} \id.
\end{equation*}
Recall the definition of $h_t = \lmin(V_t) / \big(2 \eta (\lmin(V_t)^2 + \lmax(V_t)^2)\big)$. Consequently, the previous relation becomes
\begin{equation}\label{eq:J}
    \Big(\frac{4 \eta (1 + 2\kappa_t^2)}{\kappa_t} - \eta\Big)\id \preceq \frac{2}{h_t} \Vinv_t - \eta \id \preceq \Big(3 \eta + 8 \eta \kappa_t^2\Big) \id\eqsp.
\end{equation}
The left hand term is obviously positive, therefore it holds that
\begin{align*}
    \normdeux{J_t^{1/2}} &= \normdeux{(\frac{2}{h_t} \Vinv_t - \eta \id)^{-1/2}}\\
    &\leq \sqrt{\frac{\kappa_t}{\eta (4 + 8\kappa_t^2 - \kappa_t)}}\\
    &\leq \frac{1}{\sqrt{\eta (4 + 7 \kappa_t^2 )}}\\
    &\leq \frac{1}{\sqrt{11\eta}}\eqsp.
\end{align*}
Finally,
\begin{align*}
    \sqrt{4 d \log(3t^3) \phi J_t\Vinv_t \phi^{\top}} \leq \sqrt{4 d \log(3t^3) / (11 \eta)} \Vert \phi \Vert_{\Vinv_t}\eqsp.
\end{align*}

Now, we focus on the first term of equation \ref{eq:boundforw}. Lemma \ref{lemma:wbound} gives us that
\begin{align*}
    \normdeux{\Gamma_{t, K_t}} &\leq\sum_{j=1}^{t} \frac{6 \kappa_j h_j \normdeux{V_j}}{\lambda} \sum_{r=1}^{j} \prod_{i=r}^{t} D_i^{K_i - 1}\\
    &\leq\sum_{j=1}^{t} \frac{\kappa_j}{\eta \lambda} \sum_{r=1}^{j} \prod_{i=r}^{t} D_i^{K_i - 1}\eqsp,
\end{align*}
where in the last inequality we have used that $h_t \normdeux{V_t} = \kappa_t / \big(2 \eta (1 + 2\kappa_t)\big) \leq 1 / (6 \eta)$. 
For any $j \in [2, t]$, let's define $\epsilon_j = 1 / (2 (\kappa_j d^2 t^2 \log^2(3t^3)))$. Additionally, let's fix $K_i$ such that $D_i^{K_i-1} \leq \epsilon_j$ (this condition will be explained later in the Lemma). Subsequently, we have 

\begin{align*}
    4d\log(3t^3)\normdeux{V_t}/(\eta\lambda) \sum_{j=1}^{t} \kappa_j \sum_{r=1}^{j} \prod_{i=r}^{t} D_i^{K_i - 1} &\leq 4d\log(3t^3)\normdeux{V_t}/(\eta\lambda) \sum_{j=1}^{t} \kappa_j \sum_{r=1}^{j} \epsilon_{j}^{t-r+1}
    \\
    &\leq \frac{2\normdeux{V_t}}{t^2 \eta \lambda d \log(3t^3)} \sum_{r=1}^{t} \sum_{j=r}^{t} \epsilon_{j}^{t-r}
    \\
    &\stackrel{(a)}{\leq} \frac{2\normdeux{V_t}}{t^2 \eta \lambda d \log(3t^3)}  \sum_{r=1}^{t} \sum_{j=r}^{t} \Big(\frac{1}{2d^2 t^2 \log^2(3 t^3)}\Big)^{t-r}
    \\
    &\stackrel{}{\leq} \frac{2\normdeux{V_t}}{t\eta \lambda d \log(3t^3)} \sum_{r=1}^{t} \frac{t-r+1}{t} \Big(\frac{1}{2d^2 t^2 \log^2(3 t^3)}\Big)^{t-r}
    \\
    &\stackrel{(b)}{\leq} \frac{2\normdeux{V_t}}{t \eta \lambda d \log(3t^3)} \sum_{r=1}^{t} \Big(\frac{1}{2d^2t^2 \log^2(3t^3)}\Big)^{t-r}
    \\
    &\stackrel{(c)}{\leq} \frac{2\normdeux{V_t}}{\eta t \lambda d \log(3t^3)} \sum_{u=0}^{t-1} \Big(\frac{1}{25}\Big)^{u}\\
    &\leq  \frac{3}{\eta \lambda d \log(3t^3)}\eqsp,
\end{align*}
where in $(a)$ we have used that $\epsilon_{j} \leq 1/(2 (d^2 t^2 \log^2(3t^3))$. Inequality $(b)$ is due to $t-r+1 \leq t$. The inequality $(c)$ is obtained because $1/(2d^2t^2 \log^2 3t^3) \leq 1/(4 \times \log^2(8))\leq 1/25$ and $u = t-r$. For the last inequality we have used the geometric series formula and $\normdeux{V_t} = \normdeux{\lambda \id + \sum_{s=1}^{t-1} \phi \phi^{\top}} \leq \lambda + t - 1 \leq t$, because $\lambda \leq 1$.

Consequently, as $\normdeux{\phi} \leq \normdeux{V_t^{1/2}} \Vert\phi\Vert_{\Vinv_t}$, we obtain
 \begin{align}
   \sqrt{4 d\log(3t^3)\phi \Gamma_{t, K_t} \phi^{\top}}&\leq \sqrt{\frac{3}{\eta \lambda d \log(3t^3)}} \Vert \phi \Vert_{\Vinv_t}\eqsp.
\end{align}

Moreover, the previous inequalities hold if $(1-(3/2)\eta h_i\lmin(V_i))^{K_i-1}\leq\epsilon$, following a similar reasoning than in section \ref{sec:sigma}, it follows that we need 
\begin{equation*}
    K_t \geq 1 + 4 (1 + 2\kappa_t^2) \log\Big(2\kappa_t d^2 T^2\log^2( 3T^3)\Big) / 3
\end{equation*}
\end{proof}
\section{Concentration and anti-concentration}\label{sec:conanti}

\begin{lemma}\textbf{(Concentration lemma for $\thetavar_t$)}\label{lemma:concentrationvar}

For any $t \in [T]$, given $\Etrue_t$, the following event is controlled
\begin{equation*}
    \prob(\Evar_t|\filt_{t-1}) \geq 1 - \frac{1}{t^2}
\end{equation*}
\end{lemma}
\begin{proof}
Firstly, if $t=1$, the condition is obvious because $\prob(\Evar_t|\filt_{t-1})\geq 0 $. For the rest of the proof, we assume that $t\geq 2$. Recall the definition of the event $\Evar_t$:
\begin{equation*}
    \Evar_t = \left\{\text{for any } a \in \A(x_t), \vert\phi(x_t, a)^{\top}\thetavar_t - \phi(x_t, a)^{\top} \meanpost_t\vert  \leq g_2(t) \Vert\phi(x_t, a)\Vert_{\Vinv_t}\right\}\eqsp.
\end{equation*}
with $g_2(t)= 10\sqrt{d\log(3t^3)/(\eta \lambda)} $.

Let $a \in \A(x_t)$, it holds that
\begin{align*}
    \vert \phi(x_t, a)^{\top} (\thetavar_t - \meanpost_t)\vert  &\leq \vert \phi(x_t, a)^{\top}(\thetavar_{t, K_t}-\meanvar_{t, K_t})\vert  +\vert \phi(x_t, a)^{\top}(\meanvar_{t, K_t}-\meanmeanvar_{t, K_t})\vert  + \vert \phi(x_t, a)^{\top}(\meanmeanvar_{t, K_t} -\meanpost)\vert  \eqsp,
\end{align*}
where $\thetavar_t = \thetavar_{t, K_t}$ is a sample from the variational posterior distribution trained after $K_t$ steps of Algorithm \ref{alg:update}. $\meanvar_{t, K_t}$ and $\stdvar_{t, K_t}$ are, respectively, the mean and covariance matrix of the variational posterior. Moreover, $\meanvar_{t, K_t}$ is gaussian with mean $\meanmeanvar_{t, K_t}$ and covariance matrix $\stdstdvar_{t, K_t}$ (see section \ref{def:paramalgo}). If the number of gradient descent steps is  $K^{(1)}_t \geq 1 + 4 (1 + 2\kappa_t^2) \log(2T) /3$, then Lemma \ref{lemma:concentrationsigma} shows that with probability at least $1 - 1/t^2$, we have
\begin{align*}
    \vert \phi(x_t, a)^{\top}\left(\thetavar_{t, K_t}-\meanvar_{t, K_t}\right) \vert &\leq \sqrt{4d\log(t^3)/\eta}\Big(1 + 1/\sqrt{\lambda} \Big)\Vert\phi(x_t, a)\Vert_{\Vinv_t}\\
    &\leq 4 \sqrt{d\log(t^3)/(\eta \lambda)} \Vert\phi(x_t, a)\Vert_{\Vinv_t},
\end{align*}
where the last inequality is due to $\lambda \leq 1$.

Similarly, Lemma \ref{lemma:W concentration} shows that for any $t \geq 2$, given $\Etrue_t$, if $  K^{2}_t \geq 1 + 4 (1 + 2\kappa_t^2) \log\Big(2\kappa_t d^2 T^2\log^2( 3T^3)\Big) / 3$, therefore we have
\begin{align*}
    \vert \phi(x_t, a)^{\top}\left(\meanvar_{t, K_t}-\meanmeanvar_{t, K_t}\right)\vert &\leq \Bigg(\sqrt{\frac{3}{\eta \lambda d \log(3t^3)}} + \sqrt{4 d \log(3t^3) / (11 \eta)} \Bigg) \Vert \phi(x_t, a) \Vert_{\Vinv_t}
\end{align*}
where in the last simplification we have used $\lambda \leq 1$.

Finally, Given $\Etrue_t$, let's apply Lemma \ref{lemma:mean_post_concentration} with a number of gradient descent steps such $K^{(3)}_t \geq  1 + 2(1+2\kappa_t^2) \log\left(4 R\sqrt{d T\log(3T^3)}\right)$, we obtain that
\begin{align*}
    \vert\phi(\meanmeanvar_{t, K_t} - \meanpost_t)\vert &\leq 2/\lambda \Vert\phi(x_t, a) \Vert_{\Vinv_t}.
\end{align*}

Note that $ K_t = 1 + 2 (1 + 2\kappa_t^2) \log\Big(2R\kappa_t d^2 T^2\log^2( 3T^3)\Big)  \geq \max\{K^{(1)}_t, K^{(2)}_t, K^{(3)}_t\}$ (see Equation \eqref{def:kt}), then with probability at least $1 - 1/t^2$ we have
\begin{align*}
    \vert\phi(x_t, a) ^{\top}\left(\thetavar_{t, k}-\meanpost_t\right)\vert &\leq \vert\phi(x_t, a)^{\top}\left(\thetavar_{t, k}-\meanvar_{t, k}
\right)\vert+\vert\phi(x_t, a)^{\top}\left(\meanvar_{t, k}-\meanmeanvar_{t, k}\right)\vert + \vert\phi(x_t, a)^{\top}\left(\meanmeanvar_{t, k} -\meanpost\right)\vert
 \\ 
&\leq \Big(4 \sqrt{d\log(t^3)/(\eta \lambda)} + \sqrt{\frac{3}{\eta \lambda d \log(3t^3)}} + \sqrt{4 d \log(3t^3) / (11 \eta)} + 2/\lambda\Big) \Vert\phi(x_t, a)\Vert_{\Vinv_t}\\
&\leq 10\sqrt{d\log(3t^3)/(\eta \lambda)}  \Vert\phi(x_t, a)\Vert_{\Vinv_t}\\
&\leq g_2(t) \Vert\phi(x_t, a)\Vert_{\Vinv_t}\eqsp.
\end{align*}
where the last inequality holds because $t\geq2$, $\lambda \leq 1$ and $\eta \leq 1$.
\end{proof}

\begin{lemma}(Anti-concentration lemma)
\label{lemma:anticoncentration}
Given $\Etrue_t$, if the number of gradient steps is $ K_t = 1 + 2 (1 + \kappa_t^2) \log\Big(2R\kappa_t d^2 T^2\log^2( 3T^3)\Big) 
$ 
Therefore, it holds that
\begin{equation*}
\prob\left(\phistar_t^{\top} \thetavar_{t, k}>\phistar_t^{\top} \thetastar\right)  \leq p\eqsp,
\end{equation*}
where $p = 1 / (2 \sqrt{2 \pi e})$

\end{lemma}

\begin{proof}
Firstly, note that
\begin{align*}
\prob\left(\phistar_t^{\top} \thetavar_{t, K_t}>\phistar_t^{\top} \thetastar\right) & = \prob\left(\frac{\phistar_t^{\top} \thetavar_{t, K_t}-\phistar_t^{\top} \meanmeanvar_{t, K_t}}{\sqrt{\phistar_t^{\top} \stdvar_{t, K_t} \phistar_t + \phistar_t^{\top} \stdstdvar_{t, K_t} \phistar_t}}>\frac{\phistar_t^{\top} \thetastar-\phistar_t^{\top} \meanmeanvar_{t, K_t}}{\sqrt{\phistar_t^{\top} \stdvar_{t} \phistar_t + \phistar_t^{\top} \stdstdvar_{t, K_t} \phistar_t}}\right)\eqsp.
\end{align*}
Recall that 
\begin{equation*}
\phistar_t^{\top} \meanvar_t \sim \N\big(\phistar_t^{\top}\meanmeanvar_t, \eqsp\phistar_t\stdstdvar_{t, k}\phistar_t^{\top}\big) \eqsp \text{and} \quad\phistar_t^{\top} \thetavar_{t, K_t}\sim \N\big(\phistar_t^{\top} \meanvar_{t, k}, \eqsp\phistar_t^{\top} \stdvar_{t} \phistar_{t}\big)\eqsp.
\end{equation*}

Therefore, using the conditional property of Gaussian vectors, we have 
\begin{equation*}
\phistar_t^{\top} \thetavar_t \sim \N\big(\phistar_t^{\top}\meanmeanvar_t, \eqsp\phistar_t\stdvar_t\phistar_t^{\top}+\phistar_t\stdstdvar_t\phistar_t^{\top}\big)\eqsp.  
\end{equation*}

Consequently, we have to control the term 
\begin{equation*}
Y_t := \big(\phistar_t^{\top} \thetastar - \phistar_t^{\top} \meanmeanvar_{t, K_t}\big)/\big(\sqrt{\phistar_t^{\top} \stdvar_{t, K_t} \phistar_{t, K_t} +\phistar_t^{\top} \stdstdvar_{t, K_t} \phistar_{t, K_t}}\big)
\end{equation*}
 and use the Gaussian anti-concentration lemma (Lemma \ref{abramowitzanti}). First, in this lemma, we suppose that $\Etrue_t$ holds, therefore we have
\begin{align*}
    \vert\phistar_t^{\top} (\meanpost_t - \thetastar)\vert &\leq g_1(t) \Vert \phistar_t\Vert_{\Vinv_t}\\
    &=\left(R\sqrt{d \log(3 t^3)} + \sqrt{\lambda}\right) \Vert \phistar_t\Vert_{\Vinv_t}\eqsp.
\end{align*}

Moreover, as the number of gradient descent, defined in section \ref{def:hyp} is upper than $K_t^{(1)} = 1 + 2 (1+2\kappa_i^2) \log\Big(4R \sqrt{dT\log (3T^3)}\Big)$, then
Lemma \ref{lemma:mean_post_concentration} gives us that
\begin{align*}
    \vert \phistar_t^{\top} (\meanmeanvar_{t, K_t} - \meanpost_t) \vert &\leq \frac{ 2 \Vert   \phistar_t\Vert_{\Vinv_t}}{\lambda}.
\end{align*}

Consequently, the numerator of $Y_t$ is upper bounded by
\begin{align*}
    \vert \phi_t^{* \top} (\thetastar- \meanmeanvar_{t, K_t}) \vert &\stackrel{a}{\leq}  \vert \phistar_t^{\top} (\thetastar- \meanpost_{t, K_t}) \vert + \vert \phistar_t^{\top} ( \meanpost_{t, K_t} -\meanmeanvar_{t, K_t}) \vert
    \\
    \stackrel{(b)}{\leq}& \Bigg(R\sqrt{d \log(3 t^3)} + \sqrt{\lambda} + \frac{2}{\lambda}\Bigg) \Vert \phistar_t^{\top}\Vert_{\Vinv_t} \\
\end{align*}

Regarding the denominator of $Y_t$, we need a lower bound for  $\Vert B_{t, k} \phistar_t\Vert_2$. Lemma \ref{lemma:Bbound1} gives us that 
\begin{equation*}
   \Vert B_{t, K_t} \phi\Vert_2 \geq 1/\sqrt{\eta} \Big(1 - \sqrt{\normdeux{V_t} C_t}\Big)\Vert\phistar_t\Vert_{\Vinv_t}
\end{equation*}
with
\begin{align}
\label{last_ineq}
    -C_t^{1/2} = -\Big(\frac{1}{\lambda} \sum_{j=1}^{t-1} \prod_{i=j+1}^{t} (1 - \frac{3 h_t\eta}{2} \lmin(V_t))^{K_i-1}\Big)^{1/2}.
\end{align}
Finally, we find a lower bound to this quantity.
\begin{align*}
     \Vert B_{t, K_t} \phi\Vert_2 &\geq 1/\sqrt{\eta} \Big(1 - \sqrt{\normdeux{V_t} C_t}\Big)\Vert\phistar_t\Vert_{\Vinv_t}
    \\
    \stackrel{(a)}{=}& \frac{\Vert\phistar_t\Vert_{\Vinv_t}}{\sqrt{\eta}} \Big(1 - \sqrt{\normdeux{V_t}}\Big(\frac{1}{\lambda} \sum_{j=1}^{t-1} \prod_{i=j+1}^{t} (1 - \frac{3 h_t\eta}{2} \lmin(V_t))^{K_i-1}\Big)^{1/2}\Big) 
    \\
    & \stackrel{(b)}{=}\frac{\Vert\phistar_t\Vert_{\Vinv_t}}{\sqrt{\eta}} \Bigg(1- \Big( \sum_{j=1}^{t-1} \epsilon^{t-j-1}\Big)^{1/2} \Bigg) 
    \\
     & \stackrel{(c)}{=}\frac{\Vert\phistar_t\Vert_{\Vinv_t}}{\sqrt{\eta}} \Bigg(1- \Big( \sum_{j=0}^{t-2} \epsilon^{j}\Big)^{1/2} \Bigg) 
     \\
     &\ \stackrel{(d)}{\geq}\frac{\Vert\phistar_t\Vert_{\Vinv_t}}{\sqrt{\eta}} \Big(1- \frac{1}{9t^{1/4}}\Big)
     \\
     &\stackrel{(e)}{\geq}\frac{\Vert\phistar_t\Vert_{\Vinv_t}}{\sqrt{\eta}} \Big(1- \frac{1}{9}\Big)\\
     &=\frac{8\Vert\phistar_t\Vert_{\Vinv_t}}{9\sqrt{\eta}}
\end{align*}

with (a)  is \ref{last_ineq}.
Where (b) we use $\normdeux{V_t}\leq t$ and setting $\epsilon=( 4t )^{-1} $, point (c) comes from a change of variable, (d) comes from the fact that for any $t\geq1$, $\sum_{j=0}^{t-2} \epsilon^{j}<1/(81\sqrt{t})$. Finally, (e) comes from that  $1/t$ by can be upper bounded by 1 for any $t$.

Finally, regrouping the nominator and the denominator, we have the following expression for $Y_t$: 
\begin{align*}
    Y_t&\leq \frac{\phistar_t^{\top} \thetastar - \phistar_t^{\top} \meanmeanvar_{t, K_t}}{\sqrt{\phistar_t^{\top} \stdvar_{t, K_t} \phistar_{t, K_t} + \phistar_t^{\top} \stdstdvar_{t, K_t} \phistar_{t, K_t}}}\\
    &\leq \frac{\phistar_t^{\top} \thetastar - \phistar_t^{\top} \meanmeanvar_{t, K_t}}{\norm{\phistar_t}_{\stdvar_{t, K_t} }}\\
    &\leq \frac{R\sqrt{d \log(3 t^3)} + \sqrt{\lambda} + \frac{2}{\lambda}}{8/(9\sqrt{\eta})}\\
    &\leq \frac{9 R\sqrt{d \log(3 t^3)} \sqrt{\eta}}{2\lambda }\\
\end{align*}

Recall the definition of $\eta$ in Section \ref{def:hyp}
\begin{equation*}
    \eta = \frac{4 \lambda^2}{81 R^2 d \log(3T^3)}
\end{equation*}
Consequently, it yields that $\vert  Y_t \vert \leq 1$. 

Finally, Lemma \ref{abramowitzanti} gives us that
\begin{equation*}
    \prob\Big(\phistar_t^{\top} \thetavar_{t, K_t} > \phistar_t^{\top} \thetastar\Big) \geq \frac{1}{2\sqrt{2 \pi e}}
\end{equation*}

\end{proof}

\subsection{Auxiliary Lemmas}
\begin{lemma}\textbf{(Azuma-Hoeffding inequality)} \label{lemma:azuma}
We define  $\{X_s\}_{s\in[T]}$ a super-martingale associated to the filtration $\filt_t$. If it holds that for any $s \geq 1$, $|X_{s+1} - X_s| \leq c_{s+1}$. Then for any $\epsilon >0$, we have
\begin{equation*}
    \prob(X_T - X_0 \geq \epsilon) \leq \exp\left(-\frac{\epsilon^2}{2 \sum_{s=1}^T c_s^2}\right)\eqsp. 
\end{equation*}
\end{lemma}

\begin{lemma}\textbf{(Martingale Lemma \cite{abbasi:2011})}\label{lem:mart}
 Let $(\mathcal{F}_t)_{t\geq 0}$ be a filtration, $(m_t)_{t\geq1}$ be an $\R^d$-valued stochastic process such that $m_t$ is $(\mathcal{F}_{t-1}')$-measurable, $(\epsilon_t)_{t\geq 1}$ be a real-valued martingale difference process such that $\epsilon_t$ is $(\mathcal{F}_t')$-measurable. For $t\geq 0$, define $\zeta_t = \sum_{\tau =1}^t m_{\tau}\epsilon_{\tau}$ and $M_t = \id_d + \sum_{\tau=1}^t m_{\tau} m_{\tau}^{\top}$, where $\id_d$ is the d-dimensional identity matrix. Assume $\epsilon_t$ is conditionally R-sub-Gaussian. Then, for any $\delta'>0$, $t\geq 0$, with probability at least $1 - \delta'$,
\begin{equation*}
    \Vert\zeta_t\Vert_{M_t^{-1}} \leq R \sqrt{d \log(\frac{t+1}{\delta'})}
\end{equation*}
where $\Vert\zeta_t\Vert_{M_t^{-1}} = \sqrt{\zeta_t^{\top} M_t^{-1} \zeta_t}$
\end{lemma}

\begin{lemma}\textbf{(Gaussian concentration \citep{abramowitz1964handbook})}
\label{abramowitzanti}
 Suppose $Z$ is a Gaussian random variable $Z \sim \mathcal{N}\left(\mu, \sigma^2\right)$, where $\sigma>0$. For $0 \leq z \leq 1$, we have
\begin{equation}
   \mathbb{P}(Z>\mu+z \sigma) \geq \frac{1}{\sqrt{8 \pi}} e^{-\frac{z^2}{2}}, \quad \mathbb{P}(Z<\mu-z \sigma) \geq \frac{1}{\sqrt{8 \pi}} e^{-\frac{z^2}{2}} 
\end{equation}

And for $z \geq 1$, we have
\begin{equation*}
    \frac{e^{-z^2 / 2}}{2 z \sqrt{\pi}} \leq \mathbb{P}(|Z-\mu|>z \sigma) \leq \frac{e^{-\frac{z^2}{2}}}{z \sqrt{\pi}}
\end{equation*}

\end{lemma}

\section{Approximation of our algorithm and complexity}

\label{DLsequence}
In this section, the objective is to approximate the inversion of the matrix $B_{t, k}$ of Algorithm \ref{alg:update}. Indeed, Algorithm \ref{alg:update}, requires to compute the inversion of a $d\times d$ matrix at each step $t$ and $k$, which represents a complexity of $\mathcal{O}(d^3)$. In the approximated version of Algorithm \ref{alg:update}, we consider both the sequence of square root covariance matrix $\{B_{t, k}\}_{k=1}^{K_t}$ and the sequence of their approximations $\{C_{t, k}\}_{k=1}^{K_t}$ such that: for any $t \in [T]$ and $k \in [K_t]$
\begin{equation*}
    C_{t, k} \approx B_{t, k}^{-1}\eqsp.
\end{equation*}

Recall the recursive definition of $B_{t, k}$,
\begin{align}
    B_{t, k+1} &= \left\{\id - h_t A_{t, k})\right\} B_{t, k} + h_t(B_{t, k}^{\top})^{-1\nonumber}\\
    &\approx \left\{\id - h_t A_{t, k})\right\} B_{t, k} + h_tC_{t, k}^{\top}\label{seq:B}\eqsp,
\end{align}
where $A_{t, k} = B_{t, k}^2(\thetavar_{t, k} - \meanvar_{t, k})(\nabla U_t(\thetavar_{t, k})^{\top}$ if the hessian free algorithm is used or $A_{t, k} = \nabla^2 U(\thetavar_{t, k})$ otherwise. Recall that $\thetavar_{t, k} \sim \N(\meanvar_{t, k}, B_{t, k} B_{t, k}^{\top})$. Furthermore, we can now focus on the definition of the sequence $\{C_{t, k}\}_{k=1}^{K_t}$. Firstly, we recall that
\begin{align*}
     B_{t, k+1} &= \left\{\id -  h_t A_{t, k} \right\} B_{t, k} + h_t (B_{t, k}^{\top})^{-1} \\
    &= \{\id + h_t ((B_{t, k}^{\top})^{-1} (B_{t, k})^{-1} - A_{t, k})\}  B_{t, k}.
\end{align*}

Then, let's use a first order Taylor expansion of the previous equation in $h_t$, we obtain the approximated inverse square root covariance matrix:
\begin{align}
     C_{t, k+1} =  C_{t, k}^{}  \{(\id - h_t ( C_{t, k}^{\top} C_{t, k} - A_{t, k})\}\label{seq:C}.
\end{align}
Note that the lower is $h_t$, the better is the approximation and in our case the step size $h_t$ is decreasing with $t$. The approximated recursive definition of the square root covariance matrix defined in equation \eqref{seq:B} and its approximated inverse defined in equation \eqref{seq:C} are used to defined our the approximated version of $\VITS$ called $\VITSII$ and is presented in Algorithm \ref{alg:update_approx}. Moreover,  note that the updating step of Algorithm \ref{alg:update_approx} uses only matrix multiplication and sampling from independent Gaussian distribution $\N(0, \id)$. Therefore the global complexity of the overall algorithm is $\mathcal{O}(d^2)$.

\section{Discussion on the difference between the algorithm of \cite{urteaga2018variational}  and our algorithm $\mathrm{VITS}$.}
\label{VTS}

The main difference between our setting and the one of \cite{urteaga2018variational} is the bandit modelisation. Indeed, given a context $x$ and an action $a$, in our setting, the agent receives a reward $r \sim \mathrm{R}(\cdot|x, a)$. Consequently, a parametric model $\mathrm{R_{\theta}}$ is used to approximate the reward distribution and it yields to a posterior distribution $\post$. In the setting of \cite{urteaga2018variational}, the agent receives a reward $r \sim \mathrm{R}_a(\cdot|x)$. Then, it considers a set of parametric models $\{\mathrm{R_{\theta_a}}\}_{a=1}^{K}$ and a set of posterior distributions: $\{\post_a\}_{a=1}^{K}$. The setting we have used in this paper is richer as it consider the correlation between the arms distributions compared to \cite{urteaga2018variational} which consider that the arm distributions are independents. For example, if we consider the case of the Linear bandit. In this setting, the posterior distribution is Gaussian. With the modelisation of \cite{urteaga2018variational}, we have for any $a\in [K]$,  $\post_a := \N(\mu_a, \Sigma_a)$, where $\mu_a \in \R^{d}$ and $\Sigma_a \in \sdp^{d}$. However, with our modelisation, $\post := \N(\mu, \Sigma)$, where $\mu \in \R^{d \times K}$ and $\Sigma \in \sdp^{d\times K}$. We can see that the covariance matrix $\Sigma$ encodes the correlations between the different arms, which is not the case of $\{\Sigma_a\}_{a=1}^{K}$. 
In addition, in our setting, we can consider any model for the mean of the reward distribution. For example we can choose $g(\theta, x, a)$ as a Neural Networks. This kind of model is unusable in the formulation of \cite{urteaga2018variational}.

Moreover, the approximate families used in both papers are different. Indeed, we consider the family of non-degenerate Gaussian distributions, and \cite{urteaga2018variational} is focused on the family of mixture of mean-field Gaussian distribution. The mixture of Gaussian distribution is richer than the classic Gaussian distribution. However, the non mean-field hypothesis allow to keep the correlation between arms distributions. 

Furthermore, $\mathtt{VTS}$ from \cite{urteaga2018variational} scales very poorly with the size of the problem. The variational parameters are very large: $\alpha \in \R^{K \times M}$, $\beta \in \R^{K \times M}$, $\gamma \in \R^{K \times M}$, $u \in \R^{K \times M \times d}$, $V \in \R^{K \times M \times d \times d}$ where $K$ is the number of arms, $M$ is the number of mixtures and $d$ the parameter dimension. In addition, the parameter updating step is also very costly in term of memory and speed. We have re-implemented an efficient version of their algorithm in JAX in order to scale as much as possible but many memory problems occur.

Finally, our algorithm comes with theoretical guarantees in the Linear Bandit case and outperforms empirically the others approximate TS methods. $\mathtt{VTS}$ performs poorly in practice and has no theoretical guarantee, even in the Linear case.

\section{Hyper-parameters tuning} \label{sec:hypergrid}
This section summarizes the different grid-search used to compute all the plots in this paper for the algorithms: LinTS, LMC-TS,, $\VITSI$, $\VITSII$ and $\VITSIIHF$.  

\begin{table}[H]
    \centering
    \begin{tabular}{c||c}
       Parameter  & Value \\
       \hline \hline
       inverse temperature $\eta$  & $10, 100, 500, 1000$\\
       \hline
       regularization $\lambda$  & $0.1, 1, 10$\\
    \end{tabular}
    \caption{LinTS hyperparameter grid-search}
    \label{tab:my_label}
\end{table}

\begin{table}[H]
    \centering
    \begin{tabular}{c||c}
       Parameter  & Value \\
       \hline \hline
       inverse temperature $\eta$  & $10, 100, 500, 1000$ \\
       \hline
       regularization $\lambda$  & $0.1, 1, 10$ \\
       \hline
       Nb gradient steps $K_t$ & 10, 50\\
       \hline
       learning rate $h$ & $0.001, 0.01, 0.1$
    \end{tabular}
    \caption{LMC-TS hyperparameter grid-search}
    
\end{table}

\begin{table}[H]
    \centering
    \begin{tabular}{c||c}
       Parameter  & Value \\
       \hline \hline
       inverse temperature $\eta$  & $10, 100, 500, 1000$\\
       \hline
       regularization $\lambda$  & $0.1, 1, 10$\\
       \hline
       Nb gradient steps $K_t$ & 10\\
       \hline
       learning rate $h$ & $0.001/\eta, 0.01/\eta, 0.1/\eta$ \\
       \hline
       Monte Carlo samples & $1$ (\texttt{Hessian}) and 20 (\texttt{Hessian-free})
    \end{tabular}
    \caption{$\VITS$ hyperparameter grid-search}
   
\end{table}

\section{Experimental comparison between Langevin Monte Carlo and VI} \label{sec:compar}

In this section, we conduct an experimental comparison between Langevin Monte Carlo (LMC) and two variants of Variational Inference (VI), denoted as VI-I and VI-II, in approximating a specific target distribution. Our target distribution is a straightforward Gaussian distribution, represented as $\mathrm p_{\star} = \N(\mu_{\star}, \Sigma_{\star})$. We perform LMC, VI-I, and VI-II for a designated number of iterations. In each iteration, we calculate the Kullback-Leibler distance between the approximated distribution and the target distribution. In this context, all distributions generated by LMC, VI-I, and VI-II take the form of Gaussians. To compute the mean and covariance matrix for LMC, we perform parameter averaging over the results obtained after 1000 burn-in steps (which are excluded from the plotted data). Then, the training is stopped when 
\begin{equation}\label{eq:kleps}
\KL(q_k, \mathrm p_{\star}) \leq \epsilon \eqsp,
\end{equation}
or if the number of steps exceeds $50000$ steps. 

\begin{figure}[H]
    \centering
    \includegraphics[width=0.32\textwidth]{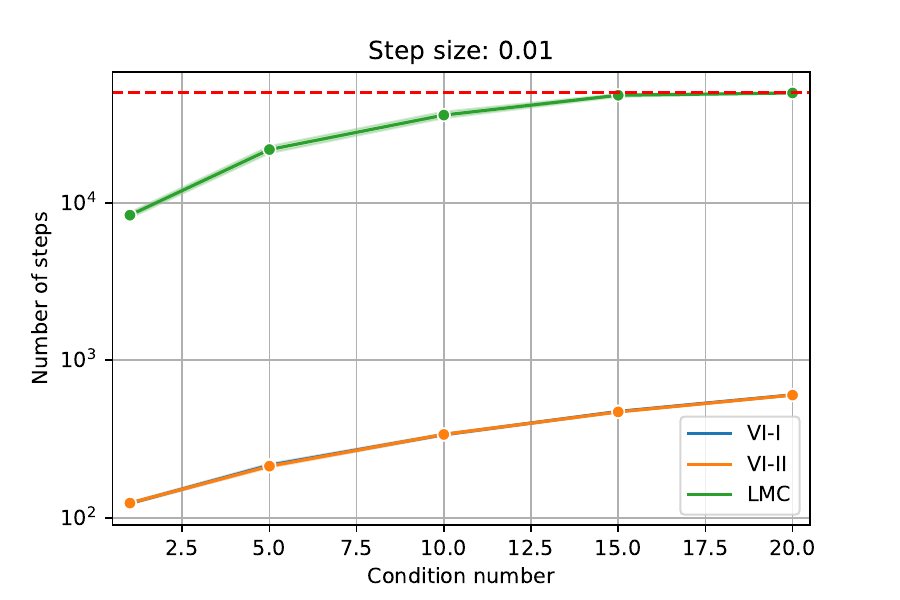}
    \includegraphics[width=0.32\textwidth]{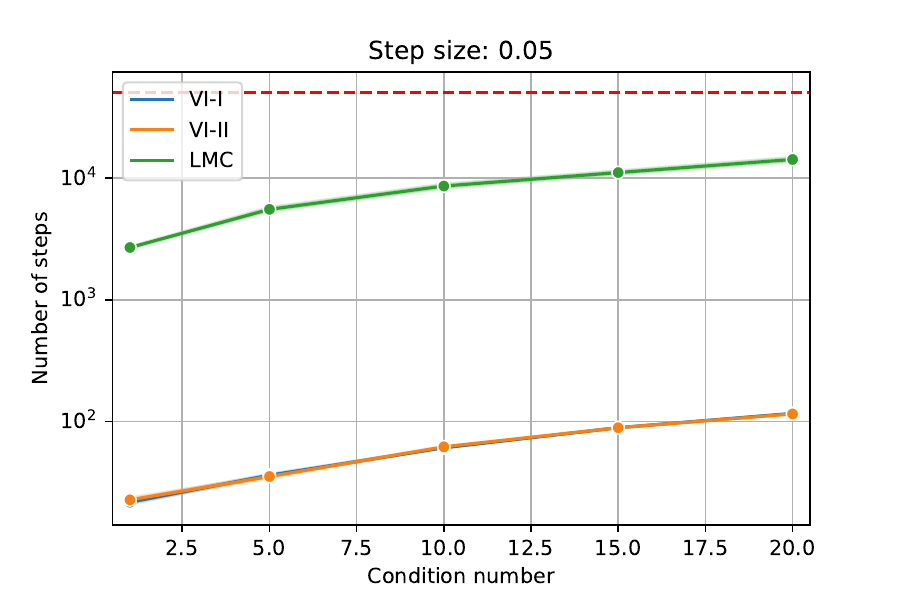}
    \includegraphics[width=0.32\textwidth]{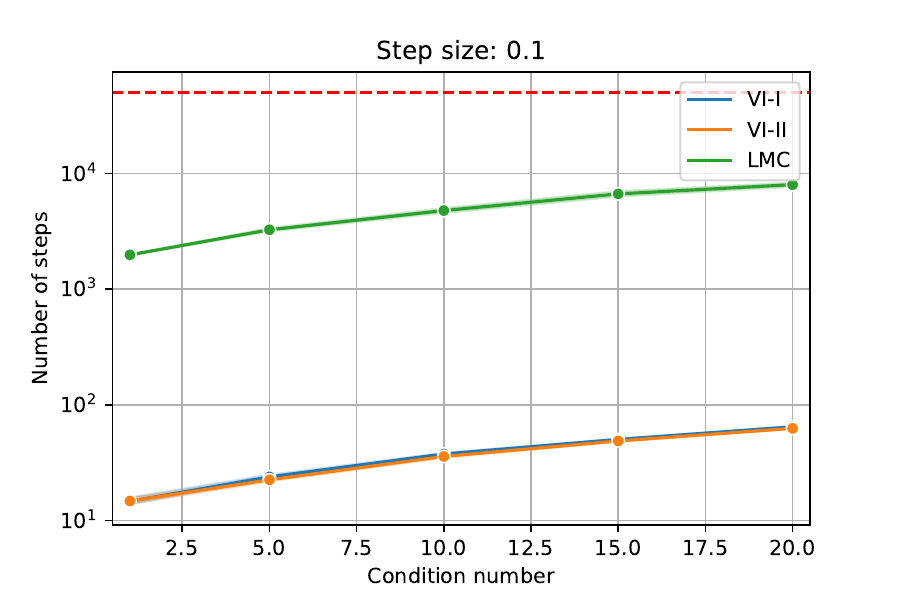}
    \caption{Comparison Langevin Monte Carlo and Variational inference}
    \label{fig:lgv_vi}
\end{figure}

Figure \ref{fig:lgv_vi} illustrates the relationship between the condition number of $\Sigma_{\star}$ and the number of steps needed to achieve \eqref{eq:kleps}. We conducted these experiments with three different step sizes and repeated them across 100 different seeds. The red dashed line in the figure represents the maximum allowable number of iterations.

The first observation drawn from these figures is that VI-I and VI-II exhibit identical behavior, even when using a relatively large step size of 0.1. The second finding suggests that both LMC and VI exhibit a linear dependency on the condition number. However, we cannot definitively conclude that one algorithm is more robust in the face of varying condition numbers. Lastly, the third conclusion highlights that VI consistently requires fewer iterations to achieve \eqref{eq:kleps}. 
\label{sec:lmcts_vs_vits}

\section{Additional Results on non-contextual bandits}
\label{sec:add_simu}
\subsection{Linear and logistic bandit on synthetic data (non contectual)}
In this subsection, we consider a contextual bandit setting with a parameter dimension $d = 10$ and a number of arms $K = 10$. The bandit environment is simulated by a random vector $\theta^{\star} \in \R^d$ sampled from a normal distribution $\N(0, \id_d)$ and subsequently scaled to unit norm. To create a complex environment that necessitates exploration, we define the set of contextual vectors as $\X := \{\thetastar, \thetastar_{\epsilon}, x_2, \dots, x_{K}\}$. Here, $\thetastar_{\epsilon}$ is defined as $(\thetastar + \epsilon) / \Vert (\thetastar + \epsilon)\Vert_2$, where $\epsilon$ is sampled from a normal distribution with mean $0$ and standard deviation $0.1$. This contextual vector corresponds to a small modification of $\thetastar$. The other contextual vectors are sampled from a normal distribution $\N(0, 1)$ and then scaled to unit norm.

\textbf{Linear bandit scenario.} Here, the true reward $\mathcal{R}(\cdot |x_a,a)$ associated to an action $a \in \{1, \ldots, K\}$ and an arm $x_a \in \rset^d$ corresponds to the distribution of $r_a = x_a^{\top} \thetastar + \xi$, where the noise $\xi$ is sampled from $\N(0, \id)$. In this complex setting, we can calculate the expected reward for each arm as follows: $\mu_0 = \E[r_0] = 1$, $\mu_1 \approx 1 < \mu_0$, and for any $i > 1$, $\mu_i < \mu_1$. Intuitively, the first and second arms offered high rewards, while the remaining arms offered low rewards. On the other hand, finding the optimal arm is challenging and needs a significant amount of exploration. 

\textbf{Logistic bandit framework.} We consider the same contextual set $\X$, but the true reward $\mathcal{R}(\cdot |x_a,a)$ associated to an action $a \in \{1, \ldots, K\}$ and an arm $x_a$  now corresponds to $r_a \sim \Ber(\sigma(\langle x_a, \thetastar\rangle))$, where $\Ber$ is the Bernoulli distribution, and $\sigma(x) = 1 / (1 + e^{-x})$ is the logistic function. Similarly to the linear bandit, the logistic framework introduces a complex environment where a significant amount of exploration is required to accurately distinguish between the first and second arm.

\begin{figure}[H]
    \centering
\includegraphics[width=0.45\textwidth]{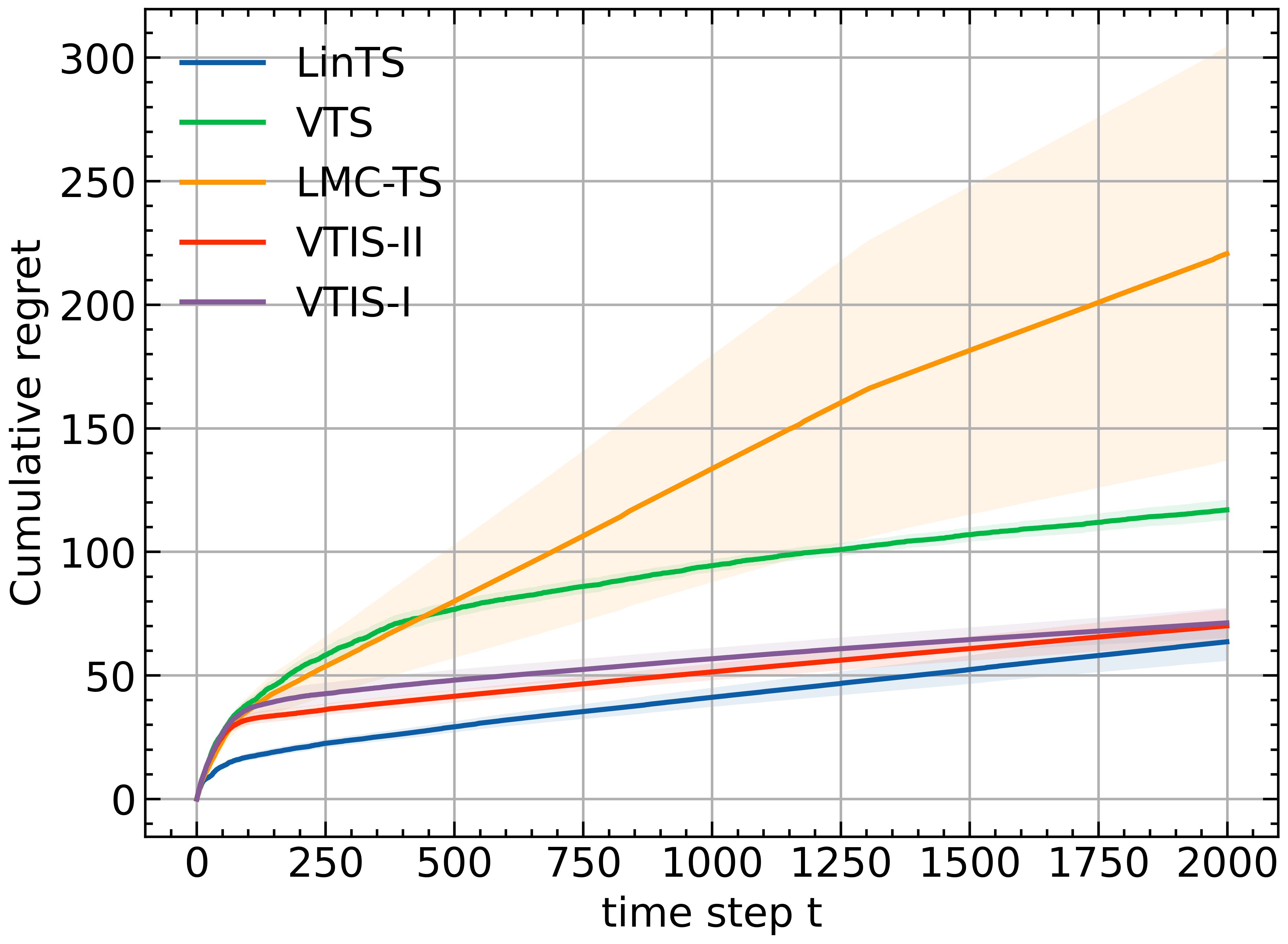} 
    \quad \quad 
\includegraphics[width=0.45\textwidth]{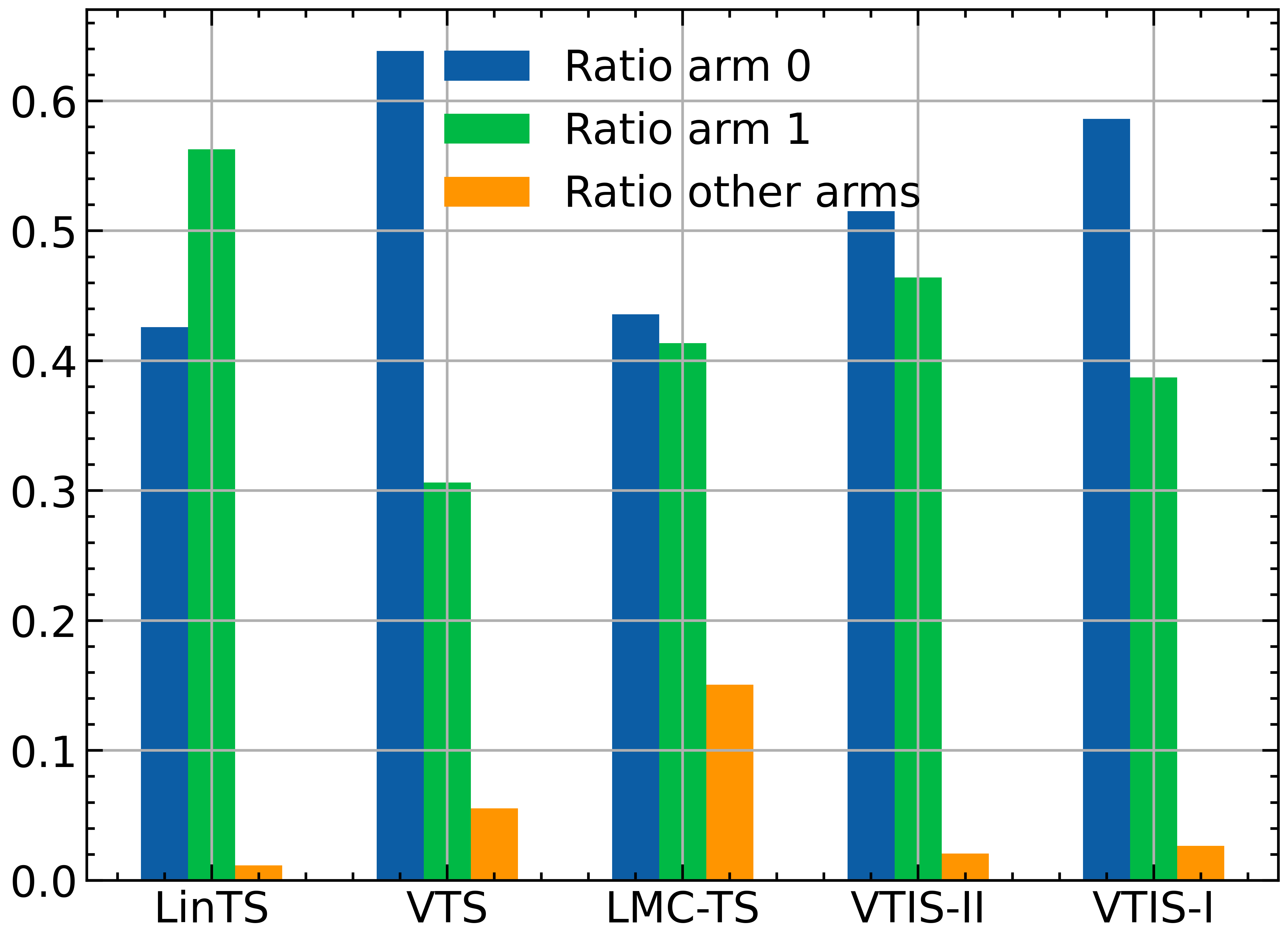} 
    \caption{Linear Bandits}%
    \label{fig:linear}%
\end{figure}

\begin{figure}{}{}
\centering
 
    \includegraphics[width=0.45\textwidth]{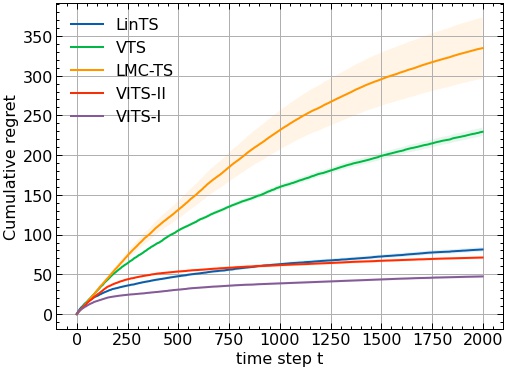}
    \caption{Logistic Bandits}%
    \label{fig:logistic}%
  
\end{figure}

 Figures \ref{fig:linear} and \ref{fig:logistic}
display the cumulative regret \eqref{eq:regret} obtained by various TS
algorithms, namely Linear TS (LinTS), Langevin Monte Carlo TS
(LMC-TS), Variational TS (VTS), \VITSI~and \VITSII~in  the linear
and logistic bandit settings.
The figure shows the mean and
standard error of the cumulative regret over 20 samples. As depicted in Figure \ref{fig:linear}, 
\VITSI~outperforms the other approximate TS
algorithms in the linear bandit scenario. Note that the cumulative regret of
\VITSI~and \VITSII~ is comparable to that
of Lin-TS, which uses the true posterior distribution. This observation highlights the efficiency of the variational TS algorithms
in approximating the true posterior distribution and achieving similar
performance to the Lin-TS algorithm. Figure \ref{fig:logistic} shows
that $\VITS$ outperforms all other TS
algorithms in the logistic setting too. This highlights the importance of employing 
approximation techniques in scenarios where the true posterior
distribution cannot be sampled exactly. Moreover, both figures
illustrate that \VITSII~ achieves a comparable regret to
\VITSI~while significantly reducing the computational
complexity of the algorithm.
Finally, as emphasized earlier, the settings we have chosen  require a good tradeoff between exploration and exploitation that LMC-TS cannot achieve, as illustrated by the histogram in Figure \ref{fig:linear}.

\section{Details about experiences in synthetic contextual bandits
\label{sec:simu}
with synthetic data}

In this subsection, we provide more details about the toy example derive in this paper. Firstly, we consider a fixed pool of arms denoted as $P = [\tilde{x}_1, \ldots, \tilde{x}_{n}]$ with $n = 50$, where each arm $\tilde{x}_i$ follows a normal distribution $\N(0_d, \id_d)$. Then, at each step $t \in [T]$, for every arm, we randomly sample a vector $\tilde{x}_i$ from the pool $P$, and the contextual vector associated with this arm is defined as $x = \tilde{x}_i + \zeta \epsilon$, where $\epsilon \sim \mathcal{N}(0_d, \id_d)$.
The bandit environment is simulated using a random vector $\theta^{\star}$ sampled from a normal distribution $\N(0_d, \sigma^{\star} \id_d)$. We opted for $\sigma^{\star} = 1/d$ to ensure that the variance of the scalar product $x^{\top} \theta^{\star}$ remains independent of the dimension $d$. Indeed, both linear and quadratic settings, the reward only depends on the scalar product between the context and the true parameter. If we denote by $x[i]$ and $\thetastar[i]$ the $i^{\text{th}}$ coordinate of the vector $x$ and $\thetastar$ respectively, then the scalar product is defined by
\begin{equation*}
    x^{\top} \thetastar = \sum_{i=1}^{d} x[i] \thetastar[i],
\end{equation*}
and its variance is defined by
\begin{align*}
    \V[ x^{\top} \thetastar] &= \V[\sum_{i=1}^{d} x[i] \thetastar[i]]\\
    &= \sum_{i=1}^{d} \V[x[i]] \V[\thetastar[i]]\\
    &= d  \sigma^{\star}\eqsp \V[x[i]].
\end{align*}
In the previous equations we have used that all coordinates are independents identically distributed and centered. Therefore, taking $\sigma^{\star} = 1/d$ ensure that the variance of the scalar product remains independent of d. In the linear bandit setting, the reward depends linearly on the contextual vector $x$, more precisely,

\begin{equation*}
    r = x^{\top} \thetastar + \alpha \epsilon \eqsp,
\end{equation*}
where $\epsilon \sim \N(0_d, \id_d)$. However, to maintain problem complexity independent of $\zeta$, we have set the signal-to-noise ratio to a fixed value of 1. This signal-to-noise ratio is the ratio between $\E[(x^{\top} \theta^{\star})^2]$ and $\E[(\alpha \epsilon)^2]$.
Firstly,
\begin{align*}
    \E[(x^{\top} \theta^{\star})^2] &= \V[x^{\top} \theta^{\star}]\\
    &= \V[x[i]]\\
    &= 1 + \zeta^2 \eqsp,
\end{align*}
where in the last equation we have used that $x = \tilde{x}_i + \zeta \epsilon$ and $\V[x[i]] = 1 + \zeta^2$. Moreover, the  denominator of the signal-to-noise ratio is $\E[(\alpha \epsilon)^2] = \alpha^2$. Consequently, a signal-to-noise ratio equals to $1$ implies that  $\sqrt{1 + \zeta^2} = \alpha$.

In the quadratic bandit setting, the reward depends quadratically on the contextual vector $x$, more precisely,
\begin{equation*}
    r = (x^{\top} \thetastar)^2 + \alpha \epsilon\eqsp,
\end{equation*}
where $\epsilon \sim \N(0, \id)$. In this setting, the reward also depends only on the scalar product between $x$ and $\thetastar$, thus, we also choose $\sigma^{\star} = 1/d$. We also ensure a signal-to-noise equal to 1, it implies a more sophisticated condition on the noise: $\alpha= (\zeta^2 +1) \sqrt{3 +6/d}$. More precisely, in the quadratic setting, the signal-to-noise ratio is defined as follow

\begin{equation*}
    \frac{\E[(x^{\top} \theta^{\star})^4]}{\E[(\alpha \epsilon)^2]} = 1.
\end{equation*}

Firstly,
\begin{align*}
    \E(x^{\top} \theta^{\star})^4] &= \E[(\sum_{i=1}^{d} x[i] \thetastar[i])^4]\\
    &= \E\Bigg[\sum_{i=1}^{d} (x[i] \thetastar[i])^4 + 4 \sum_{i=1}^{d} \sum_{j \neq i} (x[i] \thetastar[i])^3 x[j] \thetastar[j] + 6 \sum_{i=1}^{d} \sum_{j<i} (x[i] \thetastar[i])^2 (x[j] \thetastar[j])^2 \\
    &+ 12 \sum_{i=1}^{d} \sum_{j \neq i} \sum_{k \neq i, k <j} (x[i] \thetastar[i])^2 x[j] \thetastar[j] x[k] \thetastar[k] + 24 \sum_{i=1}^{d} \sum_{j < i} \sum_{k < j} \sum_{l < k} x[i] \thetastar[i] x[j] \thetastar[j] x[k] \thetastar[k] x[l] \thetastar[l] \Bigg]\\
    &= \sum_{i=1}^{d} \E[x[i]^4] \E[\thetastar[i]^4] + 6 \sum_{i=1}^{d} \sum_{j<i} \E[x_i^2] \E[x_j^2] \E[\thetastar[i]^2] \E[\thetastar[j]^2]\\
    &= \frac{9 (\zeta^2 + 1)^2}{d} + 6 \binom{d}{2} \frac{(\zeta^2 + 1)^2}{d^2}\\
    &= (\zeta^2 + 1)^2 (\frac{9}{d} + \frac{3 (d-1)}{d})\\
    &= (\zeta^2 + 1)^2 (\frac{6}{d} + 3)
\end{align*}

which gives that $\alpha= (\zeta^2 +1) \sqrt{3 +6/d}$

\section{Computation complexity}\label{sec:compt}
We conduct an experimental comparison between Langevin Monte Carlo
(LMC) and three variants of Variational Inference, denoted as $\VITSI$, $\VITSII$ and $\VITSIIHF$, in approximating a specific target distribution. Our target distribution is a straightforward Gaussian distribution, represented as $p^{\star} = \N(\mu^{\star},\Sigma^{\star})$. At each iteration, we calculate the Kullback-Leibler distance between the approximated distribution and the target distribution. In this context, all distributions generated by LMCTS, $\VITSI$, $\VITSII$ and $\VITSIIHF$ take the form of Gaussians. To compute the mean and
covariance matrix for LMC, we perform parameter averaging. As both the posterior
and its approximation are Gaussians, the Kullback-Leibler divergence is
easily tractable. Then, the training is stopped when
\begin{equation*}\label{eq:klbound}
    \KL(q_k, p^{\star}) \leq \epsilon 
\end{equation*}

or if the number of steps exceeds 10000 steps.

The following Figure illustrates the relationship between the obtained Kullback-
Leibler divergence and the computational time needed to achieve \ref{eq:klbound}. The computational
time is the total time (in second) required to run all updating steps of
the algorithm. This experiment is repeated across 1000 different seeds to compute the confidence interval. We decide not to compare with LinTS or LinUCB
algorithms as they do not allow to approximate complex posteriors compared
to LMCTS and VITS algorithms.

\begin{figure}[H]
    \centering
    \includegraphics[width=0.5\textwidth]{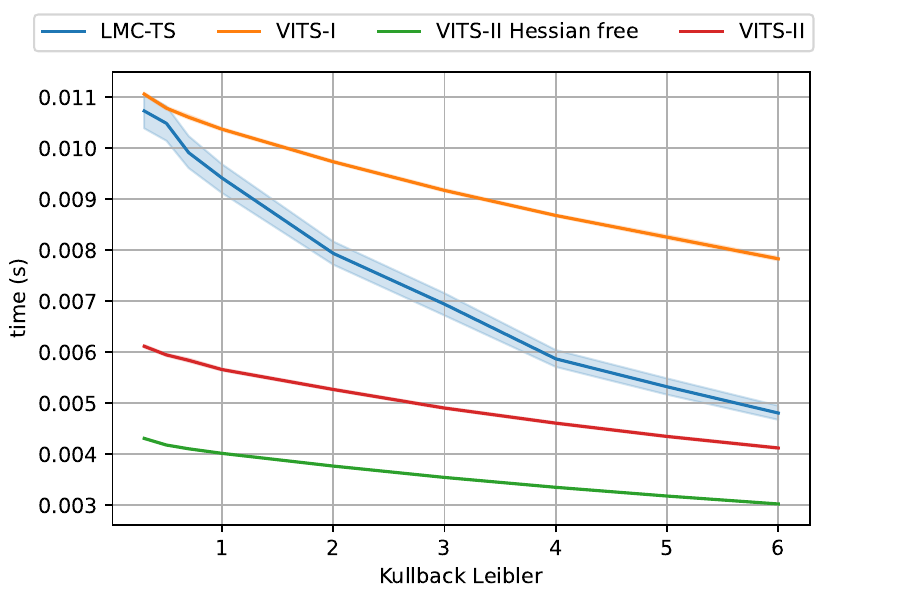}
\end{figure}

This figure shows that $\VITSII$ and $\VITSIIHF$ are faster (in term of computational time) than LMCTS to obtain a certain Kullback-Leibler
divergence. Note that $\VITSI$ is the slowest algorithm, this is due to the costly
inverse matrix calculation.

\section{Computational Power}

In this work, we use GPUs v100-16g or v100-32g for running our code with GPU Nvidia Tesla V100 SXM2 16 Go and CPUs with 192 Go per node.

\end{document}